\documentclass[acmtog,authorversion]{acmart}
\acmSubmissionID{837}
\usepackage{booktabs} % For formal tables
\usepackage[ruled]{algorithm2e} % For algorithms
\usepackage{multicol}
\usepackage{multirow}
\usepackage{cleveref}
\usepackage{array}
\usepackage{subcaption}
\usepackage{wrapfig}
\usepackage{amsmath}
\usepackage{makecell}
\usepackage{diagbox}
\usepackage{tikz}
\usepackage{xcolor}
\usepackage{fontawesome5}
\usepackage{pifont}
\newcolumntype{?}{!{\vrule width 1pt}}

\SetAlFnt{\small}
\SetAlCapFnt{\small}
\SetAlCapNameFnt{\small}
\SetAlCapHSkip{0pt}
\newcommand{\eg}[0]{\textit{e.g. }}

\newcommand{\etc}[0]{\textit{etc}}
\DeclareRobustCommand{\colordot}[1]{%
  \tikz[baseline=-0.6ex]\draw[black,fill=#1,radius=0.7ex] (0,0) circle ;%
}
\DeclareRobustCommand{\cmark}{\textcolor{green!70!black}{\ding{51}}} % ✓
\DeclareRobustCommand{\xmark}{\textcolor{red!70!black}{\ding{55}}} % ✗

\definecolor{junglegreen}{rgb}{0.113, 0.639, 0.5}
\newcommand{\mhe}[1]{{\color{black}{#1}}}

\newcommand{\methodname}[0]{\textit{EgoRelight}}
\begin{document}
%
%%%%%%%%%%%%%%%%%%%%%%%%
% TITLE
%%%%%%%%%%%%%%%%%%%%%%%%
%
\title{EgoRelight: Egocentric Human Capture and Illumination Recovery for Relightable and Photoreal Avatar Rendering} 
%%%%%%%%%%%%%%%%%%%%%%%%
% AUTHORS
%%%%%%%%%%%%%%%%%%%%%%%%
%

\author{Jianchun Chen}
\affiliation{%
  \institution{MPI for Informatic, SIC \& VIA Research Center}
  \city{Saarbr{\"u}cken}
  \country{Germany}}
\email{jchen@mpi-inf.mpg.de}

\author{Yinda Zhang}
\affiliation{%
 \institution{Google}
 \city{Mountain View}
 \country{USA}}
 \email{yindaz@google.com}

\author{Rohit Pandey}
\affiliation{%
 \institution{Google}
 \city{Mountain View}
 \country{USA}}
 \email{rohitpandey@google.com}

\author{Thabo Beeler}
\affiliation{%
 \institution{Google}
 \city{Z{\"u}rich}
 \country{Switzerland}}
 \email{tbeeler@google.com}

\author{Marc Habermann}
\affiliation{%
  \institution{MPI for Informatic \& VIA Research Center}
  \city{Saarbr{\"u}cken}
  \country{Germany}}
\email{mhaberma@mpi-inf.mpg.de}

\author{Christian Theobalt}
\affiliation{%
  \institution{MPI for Informatic, SIC \& VIA Research Center}
  \city{Saarbr{\"u}cken}
  \country{Germany}}
\email{theobalt@mpi-inf.mpg.de}

%%%%%%%%%%%%%%%%%%%%%%%%
% SHORT AUTHOR
%%%%%%%%%%%%%%%%%%%%%%%%
%
%
%%%%%%%%%%%%%%%%%%%%%%%%
% TEASER IMAGE
%%%%%%%%%%%%%%%%%%%%%%%%
%
\begin{teaserfigure}
\includegraphics[trim={0cm 0cm 0cm 0cm},clip,width=\textwidth]{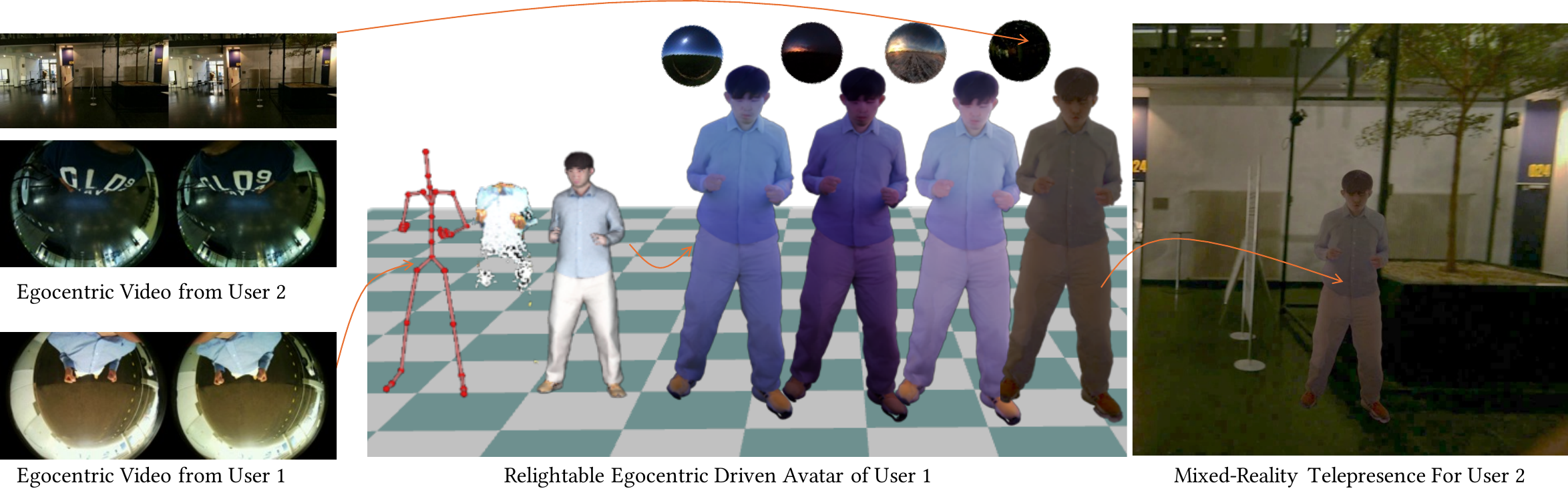}
%
% \vspace{-24pt}
\caption{We present \methodname, a new method for photoreal and relightable avatars solely driven from egocentric stereo views. Our method allows relighting the avatar into arbitrary environments while we also support lighting acquisition just from a head-mounted display. In the right column, we demonstrate a telepresence application where we seamlessly insert User 1 into User 2's real environment.
}
\label{fig:teaser}
\end{teaserfigure}
%
%%%%%%%%%%%%%%%%%%%%%%%%
% ABSTRACT
%%%%%%%%%%%%%%%%%%%%%%%%
%
\begin{abstract}
%
%%%%%%%%%%%%%%%%%%%%%%%%%%%%%%%%%%%%%%%%%%%%%%
%
Mixed Reality (MR) headsets promise a future of immersive telepresence where virtual humans blend indistinguishably into their real or virtual surroundings.
Achieving this vision requires a method capable of capturing a user's motion, estimating their appearance under novel lighting, and understanding the surrounding environment -- all from the constrained viewpoint of a head-mounted display (HMD).
Existing approaches treat these as isolated problems: 
they either focus on driving avatars with baked-in lighting or rely on complex studio setups for relighting.
In this paper, we present \methodname, a holistic framework for egocentric telepresence that simultaneously captures full-body human performance, synthesizes photorealistic and relightable appearance, and estimates high dynamic range (HDR) environment maps from a single HMD.
First, to ensure accurate motion and surface reconstruction, we propose an egocentric perception module that leverages stereo down-facing cameras to extract dense depth maps, which serve as robust geometric control signals to drive a mesh-based avatar.
Second, we introduce a novel neural appearance model that learns to synthesize view-dependent specular and view-independent diffuse shading separately. 
By employing a specialized ray-sampling strategy, our model generalizes to unseen illumination without relying on restrictive analytical BRDF priors.
Third, we enable seamless avatar integration into the physical world via a test-time inverse rendering process, which recovers an HDR environment map by matching the pre-trained avatar’s appearance to the live egocentric camera observations.
We demonstrate the efficiency of our system through a social telepresence application, where remote users are coherently relit according to their local physical environment. 
Extensive experiments show that our individual components and the integrated system significantly outperform the best combinations of state-of-the-art baselines in both geometric accuracy and rendering as well as relighting fidelity.
Further details can be found on our project page\footnote{https://vcai.mpi-inf.mpg.de/projects/EgoRelight/}.
%%%%%%%%%%%%%%%%%%%%%%%%%%%%%%%%%%%%%%%%%%%%%%
%
\end{abstract}
%
%%%%%%%%%%%%%%%%%%%%%%%%
% CCS
%    <concept>
%        <concept_id>10010147.10010371.10010372.10010376</concept_id>
%        <concept_desc>Computing methodologies~Reflectance modeling</concept_desc>
%        <concept_significance>500</concept_significance>
%        </concept>
%  </ccs2012>
% \ccsdesc[500]{Computing methodologies~Reflectance modeling}
%
%%%%%%%%%%%%%%%%%%%%%%%%
%
\begin{CCSXML}
<ccs2012>
   <concept>
       <concept_id>10010147.10010371.10010372.10010376</concept_id>
       <concept_desc>Computing methodologies~Reflectance modeling</concept_desc>
       <concept_significance>500</concept_significance>
       </concept>
 </ccs2012>
\end{CCSXML}
\ccsdesc[500]{Computing methodologies~Reflectance modeling}
%
%%%%%%%%%%%%%%%%%%%%%%%%
% KEYWORDS
%%%%%%%%%%%%%%%%%%%%%%%%
%
\keywords{Human rendering, relighting, neural rendering, performance capture, egocentric vision.}
%
%%%%%%%%%%%%%%%%%%%%%%%%
% MAKE TITLE
%%%%%%%%%%%%%%%%%%%%%%%%
%
\maketitle
%
%%%%%%%%%%%%%%%%%%%%%%%%
% BODY
%%%%%%%%%%%%%%%%%%%%%%%%
%
%
%%%%%%%%%%%%%%%%%%%%%%%%
% SECTIONS
%%%%%%%%%%%%%%%%%%%%%%%%
%

%
%%%%%%%%%%%%%%%%%%%%%%%%%%%%%%%%%%%%%%%%%%%%%%
%
\section{Introduction} \label{sec:intro}
Driven by the burgeoning AR/VR industry, virtual telepresence is rapidly emerging as a compelling application.
To deliver truly immersive user experiences, the latest generation of VR headsets increasingly emphasize Mixed Reality (MR) capabilities, seamlessly blending real and virtual environments \cite{milgram1994taxonomy}. 
This evolution introduces a significant challenge for existing egocentric telepresence systems \cite{chen2024egoavatar}: 
how to faithfully capture a real person from body-mounted sensing devices and then seamlessly integrate the digital twin into virtual or real environments.
To achieve this, we identify two core technical requirements: 
1) an egocentric driven, accurate, and efficient motion capture and character animation system; 
and 2) a photorealistic and \textit{relightable} avatar model, integrated with an on-device illumination measurement.
%
%%%%%%%%%
%
\par 
In the domain of digital twin creation, remarkable successes~\cite{liu2021neural,bagautdinov2021driving,zheng2023avatarrex, peng2021neural} have been achieved with the help of morphable body models~\cite{loper2023smpl} and neural rendering~\cite{mildenhall2020nerf}.
These approaches learned from monocular/multi-view videos can animate virtual humans given input motion sequences, and synthesize novel view human performance in real time.
Subsequent works introduce person- and garment-specific~\cite{xiang2021modeling,xiang2022dressing,habermann2021real} templates, more advanced neural rendering algorithms~\cite{Pang_2024_CVPR}, and additional sparse-view sensory inputs~\cite{remelli2022drivable, shetty2024holoported} to enhance the fidelity of the rendered virtual double.
Conversely, egocentric vision~\cite{grauman2022ego4d,xu2019mo,rhodin2016egocap} primarily focuses on extracting human pose and understanding human-scene interactions from head-mounted devices (HMD).
While a limited amount of work~\cite{hu2021egorenderer,elgharib2020egocentric} attempts to bridge the gap between egocentric motion capture and rendering of dynamic digital avatar, these approaches lack photorealism and temporal consistency.
Most recently, \citet{chen2024egoavatar} proposed EgoAvatar, a personalized full-body 3DGS~\cite{kerbl20233d} avatar animated from egocentric videos, enabling photorealistic egocentric telepresence for the first time.
Nevertheless, existing high-fidelity avatars typically exhibit baked-in lighting, limiting their ability to synthesize realistic shading effects when integrated into new environments.
%
%%%%%%%%%
%
\par 
\mhe{
On the other hand, faithfully relighting a full-body avatar to blend into arbitrary real or virtual environments remains highly challenging.
Under static illumination, prior work~\cite{chen2022relighting4d,relightneuralactor2024eccv,wang2024intrinsicavatar} combines inverse rendering~\cite{zhang2021nerfactor} with animatable avatars by jointly optimizing BRDF parameters~\cite{burley2012physically} and fixed lighting.
However, due to the entanglement of geometry, material, and illumination~\cite{ramamoorthi2001signal} and imperfect human tracking, these approaches struggle to produce physically accurate shading under novel lighting.
To reduce lighting ambiguity, the lightstage~\cite{debevec2002lighting} enables precise illumination control, but OLAT-based relighting cannot be directly applied to dynamically deforming full-body humans.
As a result, relightable human models initially focused on faces or hands~\cite{saito2024relightable,bi2021deep,iwase2023relightablehands,li2023megane}, sequence replay~\cite{he2024diffrelight,meka2020deep}, and more recently full-body avatars relying on complex surface registration~\cite{wang2025relightable}.
Nevertheless, these methods require controlled capture domes and synthetic illumination, limiting real-world applicability, while HDR environment map estimation from portable devices remains largely unexplored.
In summary, no existing method jointly captures full-body human performance and surrounding lighting using egocentric cameras while enabling relighting under novel environments.
}
%
%%%%%%%%%
%
\par 
We address the aforementioned challenges by proposing \methodname, a novel egocentric telepresence system designed to simultaneously capture human performance, estimate environmental lighting, and relight virtual characters online solely from a compact HMD.
By leveraging the estimated human performance and HDR environment maps, we seamlessly blend the user's relightable digital twin into the physical world, thereby providing an immersive first-person see-through experience, as illustrated in Fig.~\ref{fig:teaser}.
%
%%%%%%%%%
%
\par 
\methodname{} is a person-specific approach, which at test time drives a full-body and relightable avatar from egocentric stereo views while for training the tracking and relighting modules we assume multi-view recordings within a lightstage are provided.
\textit{First}, the dynamic human character is driven by our egocentric perception module.
In detail, we extract both sparse 3D keypoint and depth maps as geometry control signals from stereo egocentric down-facing cameras.
These outputs collectively animate the character by our feed-forward neural networks, ensuring a precise recovery of frontal body geometry as observed in the stereo input images.
\textit{Second}, we create a person-specific relightable appearance model from lightstage capture.
The relightable avatar appearance is learned in an data-driven manner, where our neural networks fit the surface reflectance function of the subject character given input light direction and intensity of a target illumination condition.
For higher efficiency, we propose to separately model the specular and diffuse shading, and sample the prominent rays to account for specularities.
Our end-to-end learning-based approach, which alleviates the inductive bias introduced by traditional rendering procedures (\eg principled BRDF model, spherical harmonics, \etc), therefore achieves better photorealism.
\mhe{\textit{Third}, we scan the scene at test-time using front-facing egocentric cameras, and color calibrate the acquired LDR panorama with our HDR environment maps that were used for training the relighting modules.
Importantly, we propose to leverage the character itself as a color calibration target by inverse rendering the avatar onto the egocentric down-facing views to optimize the color correction parameters.
Consequently, we obtain a color-correct HDR map that ensures the coherent and realistic integration of relightable avatars within the digitized environment.}
\par 
Our contributions are summarized as follows:
%
%%%%%%%%%
\begin{itemize}
    \item We present \methodname, an egocentric-driven, photoreal, relightable, and full-body avatar, which can be seamlessly inserted into surrounding real or virtual environments.
    \item We propose a novel egocentric mesh tracking method that leverages stereo depth maps as conditioning signal (Sec.~\ref{sec:ego-preception}).
    \item We further introduce a novel motion-driven full-body relightable avatar, which end-to-end learns the diffuse and specular shading from lightstage captures (Sec.~\ref{sec:relighting}).  
    \item We introduce an affordable in-the-wild HDR environment map capture technique from LDR egocentric cameras, which seamlessly blends the relightable virtual characters into real surroundings (Sec.~\ref{sec:env-map-capture}).
\end{itemize}
To the best of our knowledge, \methodname{} is the first in literature that \textit{holistically} supports precise egocentric full-body motion and geometry capture, photorealistic virtual human relighting,  and environment lighting recovery solely requiring a single HMD. 
%
%
%%%%%%%%%%%%%%%%%%%%%%%%%%%%%%%%%%%%%%%%%%%%%%%%%%%
%%%%%%%%%%%%%%%%%%%%%%%%%%%%%%%%%%%%%%%%%%%%%%%%%%%
%
\section{Related Work} \label{sec:related_work}
%
%%%%%%%%%%%%%%%%
%%%%%%%%%%%%%%%%
%
\subsection{Human Avatar Relighting and Inverse Rendering}
Human relighting allows users to seamlessly insert digital human avatars into novel environments with coherent shading, which directly benefits a wide range of applications in film making, games, and VR teleconferencing.
%
%%%%%%%%%%%%%%%%
%
\par \noindent \textbf{In-the-wild Human Relighting.}
Under a natural environment with uncalibrated static illumination, researchers build relightable human avatars by jointly recovering human geometry, spatially varying material, and lighting.
Specifically, the material is represented by principled BRDF models, e.g., Disney~\cite{burley2012physically}, Microfacet~\cite{cook1982reflectance}, and Lambertian~\cite{nicodemus1977geometrical} BRDF, to generate physics-based shading.
With differentiable renderers, e.g., Mitsuba~\cite{nimier2019mitsuba}, NeRF~\cite{mildenhall2020nerf}, one can optimize unknown geometry, material, and lighting parameters from images like monocular video captures~\cite{chen2022relighting4d,wang2024intrinsicavatar}, sparse video captures~\cite{xu2024relightable} or static multi-view imagery~\cite{guo2025pgc, li2024animatablegaussians, zhan2025interactive}.
However, due to the large geometry-material-lighting ambiguity, inverse rendering-based methods still suffer from intrinsic decomposition failure, which generates noticeable artifacts in novel lighting conditions in contrast to our high-quality avatars.
%
%%%%%%%%%%%%%%%%
%
\par \noindent \textbf{Human Relighting from Calibrated Lights.} 
To reduce the lighting ambiguity, \citet{debevec2002lighting} invented a Lightstage, a performance capture dome with controllable illumination composed of individual LED lights.
Following established material acquisition pipelines, researchers typically capture one-light-at-a-time (OLAT) sequences for \textit{static} characters. 
This approach leverages the linearity of light transport, enabling a screen-space synthesis of relit imagery via linear combination of OLAT captures.
However, it is nontrivial to extend the OLAT capture strategy to animatable and relightable avatars, due to the challenge of conducting geometry registration for \textit{dynamically moving} humans and clothing deformations.
Significant advancements were achieved in body parts with higher rigidity, e.g., face~\cite{bi2021deep, saito2024relightable, li2024uravatar}, hands~\cite{iwase2023relightablehands} and eyeglasses~\cite{li2023megane}, or \textit{replay} of full-body avatars~\cite{meka2020deep}.
Most recently, \citet{wang2025relightable} introduce the first photorealistic relightable avatar from lightstage data by learning radiance transfer functions based on a highly accurate surface tracking from OLAT captures.
Meanwhile, \citet{singh2025relightable} propose to effectively capture HDR-lit images sampled from a HDR environment map dataset~\cite{gardner2017learning} and to effectively learn relightable appearance of human avatars with a transformer-based neural network.
However, unlike our method, none of these works drives the acquired avatars from egocentric driving signals and estimates the lighting from HMDs directly.
%
%%%%%%%%%%%%%%%%
%
\par \noindent \textbf{Generative Human Relighting.}
Inspired by the trend of generative AI, researchers~\cite{pandey2021total,kim2024switchlight,zhang2025scaling} proposed universal priors for image-based portrait relighting.
Fine-tuned on lightstage captures, diffusion-based generative models support a large amount of image editing applications, including harmonization~\cite{ren2024relightful} and shadow removal~\cite{yoon2024generative}.
However, in contrast to our work, these approaches are limited in creating relightable and photorealistic 3D avatars, as they inevitably lack 1) physically accurate shading, 2) 3D consistency across views (as pointed out by \citet{poirier2024diffusion}), and 3) the generalization ability to out-of-distribution illuminations.
% 
%%%%%%%%%%%%%%%%
%%%%%%%%%%%%%%%%
%
\subsection{Motion- and View-driven Avatars}
Animatable avatars~\cite{bagautdinov2021driving,habermann2021real,liu2021neural,peng2021neural} aim at learning a mapping from skeletal motion to deforming geometry and photorealistic appearance that can be rendered into novel views.
Early works~\cite{liu2021neural,peng2021neural,weng2022humannerf,zheng2023avatarrex} leverage the SMPL~\cite{loper2023smpl} model to track the coarse geometry of the human and apply neural radiance fields~\cite{mildenhall2020nerf} to fit the appearance of the avatar under neutral illumination.
However, limited geometry tracking quality poses substantial challenges for accurate texture modeling, consequently preventing these approaches from producing sharp visual details.
With the pursuit of a higher-fidelity temporal geometry reconstruction, recent researchers have explored person-specific mesh template and character animation algorithms.
In particular, \citet{xiang2022dressing,xiang2021modeling} represent the cloth as an individual mesh layer and introduce a physics-based simulator to enhance the realism of the cloth animation.
\citet{habermann2021real} captures detailed clothing movement with a coarse-to-fine mesh deformation and a deep dynamic texture.
Despite plausible rendering results in novel view and novel pose, these methods are unable to faithfully re-produce the stochastic clothing dynamics.
\par
To tackle this problem, some researchers exploit stationary capture devices~\cite{lawrence2024project,shao2023tensor4d} and import video streams from sparse camera views as additional RGB~\cite{shetty2024holoported,remelli2022drivable} or RGBD~\cite{xiang2023drivable} inputs.
Direct observation of the human body ensures the pixel-perfect re-rendering of dynamic visual details, whereas the camera rig limits the portability and the general use case of these systems.
In practical scenarios given solely an egocentric image from VR headset as driving signal, EgoRenderer~\cite{hu2021egorenderer} learns the latent texture of the naked SMPL mesh through adversarial training, which lacks realism.
Recently, EgoAvatar~\cite{chen2024egoavatar} sequentially reconstructs motion, mesh, and Gaussian splats from egocentric images and achieves photorealistic free-viewpoint rendering.
Although the proposed test-time silhouette refinement excels at modeling fine body shapes from egocentric image signals, its prohibitive computational cost presents a bottleneck for real-world applicability. 
Thus, we are motivated to seek optimization-free solutions and extract additional depth conditions from binocular egocentric camera to faithfully recover the visible surface.
Further, none of the above works supports relighting whereas our method can seamlessly blend the avatar into novel lighting conditions.
%
%%%%%%%%%%%%%%%%
%%%%%%%%%%%%%%%%
%
\subsection{Egocentric Human Perception}
Driven by the rapid growth of the AR/VR market, researchers are increasingly investigating a broad spectrum of egocentric perception problems, primarily focusing on understanding human behavior and human-environment interaction from a first-person perspective.
In general, egocentric cameras can be categorized into front-facing or down-facing ones.
Front-facing cameras~\cite{grauman2022ego4d} have a better perception of the environment.
In this setup, some researchers \cite{yuan2019ego, luo2021dynamics, li2023ego, yi2025estimating} investigate recovering head pose from SLAM and solving a physically plausible motion sequence that matches the head trajectory.
However, due to the limited observation of human body, it is infeasible to faithfully capture world-space body motion and clothing dynamics from front-facing cameras.
With a down-facing camera setup, researchers~\cite{xu2019mo, tome2019xr} precisely recover joint locations, while challenges still exist due to severe self-occlusion and fast camera movements, as indicated by \citet{martinez2017simple}.
To address this difficulty, \citet{wang2021estimating,wang2022estimating,wang2023scene} introduce 3D scene constraints and generate hand joints in a denoising process, while \citet{akada2022unrealego} learn a strong egocentric motion prior by creating a large synthetic dataset.
Most recently, \citet{camiletto2025frame} and \citet{lee2025rewind} proposed to integrate the VR head trajectory with stereo egocentric cameras for human pose estimation, achieving state-of-the-art accuracy.
Our HMD prototype and algorithm support both types of egocentric perception, with front-facing cameras used in perceiving the environmental illumination and down-facing cameras for human capture.
Regarding egocentric pose estimation, we continue the strategy of personalized finetuning from EgoAvatar~\cite{chen2024egoavatar} to produce high-fidelity full-body and hand joints localization.
%
%%%%%%%%%%%%%%%%%%%%%%%%%%%%%%%%%%%%%%%%%%%%%%%%%%%
%%%%%%%%%%%%%%%%%%%%%%%%%%%%%%%%%%%%%%%%%%%%%%%%%%%
%
%
%%%%%%%%%%%%%%%%%%%%%%%%%%%%%%%%%%%%%%%%%%%%%%%%%%%
%%%%%%%%%%%%%%%%%%%%%%%%%%%%%%%%%%%%%%%%%%%%%%%%%%%
%
\section{Data Acquisition} \label{sec:dataset}
\begin{table}[t!]
    \centering
    \caption{\textbf{Data Capture.} For each subject, we capture four separate sequences for training and evaluation of our proposed method. The FPS column reports the frame rate of the multi-view studio cameras, while the egocentric video streams are always captured at 30FPS and synchronized with studio cameras. Note that in-the-wild demo sequences are not included in this table. \colordot{white} stands for flat-lit frames with LED projecting white light and icons in other color \colordot{red} \colordot{yellow} \colordot{orange} \colordot{blue} refer to the HDR-lit frames.}
    \label{tab:data-scheme}
    \footnotesize
    \begin{tabular}{|c|c|c|c|c|}
    
        \hline
        \textbf{Purpose} & \textbf{Length} & \textbf{FPS} & \textbf{HMD Recording} & \textbf{Light Pattern} \\
        \hline
        \makecell{Hand-eye \\ Calibration} & \textasciitilde 10s & 30 & \cmark & \colordot{white} $\cdots$ \colordot{white}\\
        \hline
        \makecell{Egocentric \\ Perception \\ Training (Sec.\ref{sec:ego-preception})} & 290s& 30& \cmark & \colordot{white} $\cdots$ \colordot{white}\\
        \hline
        \makecell{Animatable \& \\ Relightable \\Avatar Training\\ (Sec.~\ref{sec:relighting})}& 290s& 60& \xmark & \colordot{white} \colordot{red}  \colordot{white} $\cdots$ \colordot{white} \colordot{yellow} \colordot{white} \\
        \hline
        \makecell{Evaluation \\ (Sec.~\ref{sec:exp})} & 145s& 30 & \cmark & \colordot{orange} $\cdots$ \colordot{orange} \colordot{blue} \\
        \hline
    \end{tabular}
\end{table}
\begin{figure*}[t]
    \centering
    \includegraphics[width=0.95\textwidth]{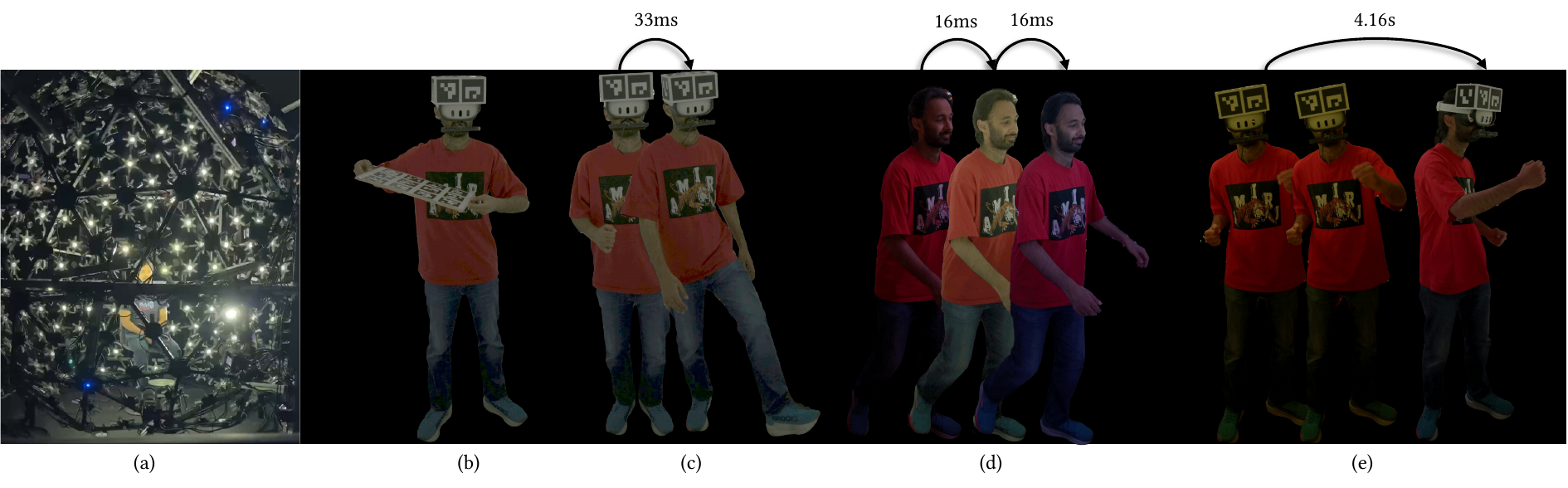}
    \vspace{-12pt}
    \caption{\textbf{Visualization of the Data Capture.} From left to right, we show our data capture setup, i.e., lightstage (a) with four separate sequences including a hand-eye calibration sequence (b), a flat-lit training sequence (c), a re-lit training sequence (d), and a testing sequence (e).}
    \label{fig:data-capture}
\end{figure*}
To create an egocentric-driven and relightable avatar, we collect person-specific recordings in a lightstage following the procedure outlined in Tab.~\ref{tab:data-scheme} and Fig.~\ref{fig:data-capture}. 
Notably, it is the first dataset that captures full body humans from an egocentric view under diverse lighting conditions.
\par
Our data capture studio, i.e., lightstage, contains calibrated multiview HDR cameras with multiple individually programmable RGB light sources covering diverse light directions, where the subject performs diverse motions in the center of the dome.
The HMD is built based on a commercial VR headset, i.e., Meta Quest 3\footnote{\url{https://www.meta.com/quest/quest-3}}, on which we mount a pair of egocentric down-facing cameras and a pair of stereo front-facing cameras.
Inside the capture dome, we additionally attach Aruco Markers to the headset, which ensures an accurate head tracking quality and alignment between the egocentric and studio camera coordinate system.
The synchronization among egocentric and multi-view cameras are conducted at the start and end of each recording session using light signals from the lightstage.
\par 
In particular, for each subject we capture a hand-eye calibration sequence and a training sequence where the person performs a variety of motions captured from egocentric cameras and external multi-view cameras.
Here, all lights are turned on with a neutral white light illumination.
This bright lighting allows us to obtain precise motion tracking and geometry reconstruction to collect data for our egocentric perception modules.
Further, we record an additional training sequence with repeated motion and interleaved diverse natural lighting conditions to capture human's appearance under real illumination.
The data acquisition scheme for learning our relightable avatar follows \citet{singh2025relightable} with synchronized cameras and LEDs capturing at 60FPS.
The light patterns switch between flat-lit frame and follow-up HDR-lit frame with HDR map sampled from the Laval dataset~\cite{gardner2017learning}.
The HMD is taken off in this sequence to capture the facial details.
Lastly, we record a testing sequence under unseen lighting condition to evaluate the robustness of the egocentric mesh capture and the relighting quality.
Here, the subject again wears the HMD. 
We discuss the data collection details in the supplemental document.
%
%%%%%%%%%%%%%%%%%%%%%%%%%%%%%%%%%%%%%%%%%%%%%%%%%%%
%%%%%%%%%%%%%%%%%%%%%%%%%%%%%%%%%%%%%%%%%%%%%%%%%%%
%
%
%%%%%%%%%%%%%%%%%%%%%%%%%%%%%%%%%%%%%%%%%%%%%%%%%%%
%%%%%%%%%%%%%%%%%%%%%%%%%%%%%%%%%%%%%%%%%%%%%%%%%%%
%
\section{Human Character Representation} \label{sec:preliminary}
Before presenting our method, we first describe the underlying human representation.
Concretely, we leverage a personalized 3D parametric model to represent the explicit geometry of the 3D human avatar animated by egocentric driving signals.
Then, we add a layer of 3D Gaussian primitives \cite{kerbl20233d} parameterized in the mesh's UV space on top of the mesh to model the photorealistic appearance.
We introduce our two-layer human representation as follows.
%
%%%%%%%%%%%%%%%%%%%%%%%%%%%%%%%%%%%%%%%%%%%%%%%%%%%
%
\subsection{Mesh-based Geometry Model}
Our geometry model is a person-specific parametric character model that reliably tracks the non-rigid full-body human and clothing dynamics.
The character model \cite{habermann2021real} contains a template mesh $\mathbf{V}_0 \in \mathbb{R}^{4890\times 3}$, a skeleton $\mathbf{B} \in \mathbb{R}^{173\times 3}$, and skinning weights $\mathbf{W} \in \mathbb{R}^{4890\times 69}$. 
With a group of predicted control parameters, i.e., skeleton motion $\boldsymbol{\theta} \in \mathbb{R}^{107}$, embedded graph rotation and translation $\boldsymbol{\alpha}, \boldsymbol{\beta} \in \mathbb{R}^{489\times 3}$, and per-vertex offset $\mathbf{o}\in\mathbb{R}^{4890\times 3}$, we deform the template mesh $\mathbf{\bar{V}}$ in canonical space using embedded graph deformation $\mathcal{E}(\cdot)$ \cite{sumner2007embedded} and transform it to the posed space 
%
%%%%
\begin{equation}
    \mathbf{V}= \mathcal{W}(\mathcal{E}(\mathbf{\bar{V}}, \boldsymbol{\alpha},\boldsymbol{\beta},\mathbf{o}), \mathcal{K}(\boldsymbol{\theta},\mathbf{B}), \mathbf{W})
    \label{eq:dqs}
\end{equation}
%%%%
%
using dual quaternion skinning (DQS) $\mathcal{W}(\cdot)$ \cite{kavan2008geometric} where $\mathcal{K}(\cdot)$ is a forward kinematics function.
This model allows an effective driving of the mesh-based avatar from skeletal motion signals while preserving higher-order deforming geometry details like clothing wrinkles.
Please refer to the supplemental document for details.
%
%%%%%%%%%%%%%%%%%%%%%%%%%%%%%%%%%%%%%%%%%%%%%%%%%%%
%
\subsection{Gaussian-based Appearance Model}
The human model detailed in the prior section addresses dynamic mesh animation, but still remains insufficient for high-fidelity photorealistic rendering.
Therefore, we introduce a layer of Gaussian texture on top of the deformed mesh, as 3DGS~\cite{kerbl20233d} has shown great potential in rendering photorealistic human avatars~\cite{Pang_2024_CVPR}, particularly thin structures and view-dependent appearance details. 
\par
We define 3D Gaussian ellipsoids in the UV space of the mesh $\mathbf{V}$, which facilitates the decoding of surface-aligned primitives using standard 2D CNNs.
For each texel, a corresponding Gaussian is parameterized with 3D offset $\Delta \mathbf{p}$, color $\mathbf{c}$, rotation $\boldsymbol{\phi}$, scaling $\mathbf{s}$, and opacity $\boldsymbol{\sigma}$, which can be transformed to its 3D location as $\mathbf{p} =\mathbf{p}'+\Delta \mathbf{p}$.
Here, $\mathbf{p}'$ is the original 3D position of the texel on the mesh surface obtained from barycentric interpolation across the vertices of the underlying face. 
\par 
A 3D Gaussian $G$ with mean $\mathbf{p}$, rotation matrix $\mathbf{R}$, and scaling matrix $\mathbf{S}$ is represented in 2D as $G(\mathbf{x},\boldsymbol{\Sigma}) = \exp ^{-\frac{1}{2}\mathbf{x}^T \boldsymbol{\Sigma}^{-1} \mathbf{x}}$.
Here, we compute its covariance matrix 
%
%%%%
\begin{equation}
    \boldsymbol{\Sigma} = \mathbf{JWRSS}^T\mathbf{R}^T\mathbf{W}^T\mathbf{J}^T
    \label{eq:gaussian_cov}
\end{equation}
%%%%
%
following \citet{zwicker2002ewa}, where $\mathbf{W}$ is the viewing transformation, and $\mathbf{J}$ is the Jacobian of its affine approximation.
By rasterizing and blending all Gaussians in the image plane, the color $\mathbf{C_q}$ of pixel $\mathbf{q}$ is rendered as 
%
%%%%
\begin{equation}
    \mathbf{C_q} = \sum_{i\in \mathcal{N}} \mathbf{c}_i' \boldsymbol{\sigma}_i' \prod_{j=1}^{i-1} (1-\boldsymbol{\sigma}'_k)
    \label{eq:gaussian_render}
\end{equation}
%%%%
%
where $\mathcal{N}$ is the set of Gaussians that are rasterized into pixel $\mathbf{q}$. 
The opacity and color of $i$-th Gaussian in pixel are considered as $\mathbf{c}_i'=\mathbf{c}_iG(\mathbf{q},\boldsymbol{\Sigma}'_i), \boldsymbol{\sigma}_i'=\boldsymbol{\sigma}_iG(\mathbf{q},\boldsymbol{\Sigma}'_i)$.
Notably, to better characterize the high-frequency shading effects in relighting, we opt for a direct prediction of view-dependent color rather than using spherical harmonics in 3DGS.
%
%%%%%%%%%%%%%%%%%%%%%%%%%%%%%%%%%%%%%%%%%%%%%%%%%%%
%%%%%%%%%%%%%%%%%%%%%%%%%%%%%%%%%%%%%%%%%%%%%%%%%%%
%
\section{Methodology Overview}
% 
%%%%%%%%%
%
%%%%%%%%%%%%%%%%%%%%%%%%%%%%%%%%%%%%%%%%%%%%%%%%%%%
%%%%%%%%%%%%%%%%%%%%%%%%%%%%%%%%%%%%%%%%%%%%%%%%%%%
%
\begin{figure*}[t]
    \centering
    \includegraphics[width=\textwidth]{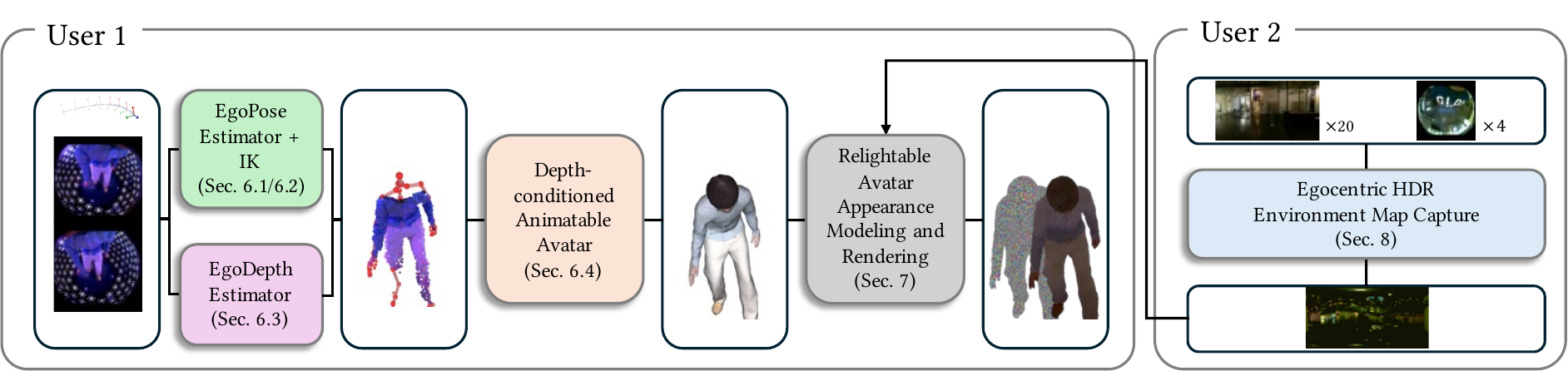}
    \caption{\textbf{Overview of \methodname.} 
    \mhe{At test time, \textit{EgoRelight} enables driving a photoreal and relightable full-body avatar solely from egocentric stereo views. 
    Moreover, it supports affordable environment illumination estimation just using the front-facing cameras of an HMD. 
    In detail, we first estimate egocentric pose (green box) and depth maps (purple box), which then drive a mesh-based human character (red box).
    The recovered surface that encodes the pose of the human then serves as input for our relightable avatar (grey box), which also takes a target illumination for relighting (blue box).}
    }
    \label{fig:full-pipeline}
\end{figure*}
%
%%%%%%%%%%%%%%%%%%%%%%%%%%%%%%%%%%%%%%%%%%%%%%%%%%%
%%%%%%%%%%%%%%%%%%%%%%%%%%%%%%%%%%%%%%%%%%%%%%%%%%%
%
%%%%%%%%%
%
\methodname{} comprises three core components. 
First, we introduce an egocentric perception module (Sec.~\ref{sec:ego-preception}) that recovers skeleton pose and depth map from stereo down-facing cameras, which we then use to drive the geometry of our avatar. 
Second, we present a relightable avatar framework (Sec.~\ref{sec:relighting}) capable of rendering photorealistic appearance under novel illumination. 
Finally, we demonstrate an inference-time environment map capture scheme (Sec.~\ref{sec:env-map-capture}), enabling a mixed-reality telepresence application that seamlessly inserts virtual users into the physical world.
An overview of our method is also shown in Fig.~\ref{fig:full-pipeline}.
%
%%%%%%%%%%%%%%%%%%%%%%%%%%%%%%%%%%%%%%%%%%%%%%%%%%%
%%%%%%%%%%%%%%%%%%%%%%%%%%%%%%%%%%%%%%%%%%%%%%%%%%%
%
\section{Egocentric Perception} \label{sec:ego-preception}
This section converts the input stereo egocentric video streams and head trajectory into avatar driving signals.
In particular, we extract sparse 3D joint keypoints $\boldsymbol{\zeta}$ in Sec.~\ref{sec:egopose} and recover skeleton motion $\boldsymbol{\theta}$ in Sec.~\ref{sec:ik}.
Meanwhile, from egocentric images, we measure depth maps $\mathbf{d}$ and unproject them as 3D point clouds $\mathbf{z}$ in Sec.~\ref{sec:egodepth}, which captures dense 3D information from egocentric observations.
Those are finally used to recover the control signals for driving the geometry of mesh-based avatar (Sec.~\ref{sec:depth-driven-avatar}).
%
%%%%%%%%%%%%%%%%%%%%%%%%%%%%%%%%%%%%%%%%%%%%%%%%%%%
%
\subsection{Egocentric Pose Estimator} \label{sec:egopose}
Among all emerging egocentric pose detectors~\cite{wang2021estimating, wang2022estimating, wang2023scene, luo2024real}, few recent methods~\cite{camiletto2025frame,lee2025rewind} have been developed under a similar egocentric setup as us with stereo down-facing camera and VR head tracking from the commercial VR headset, which is compatible with our input assumptions.
In particular, we adapt FRAME~\cite{camiletto2025frame} as our backbone pose estimator and fine-tune per identity using ground truth keypoint tracking from markerless MoCap system~\cite{thecaptury2020captury}.
To support hand gesture tracking, we replace the last layer of the FRAME network with additional channels that end-to-end regress hand joints together with body joints while re-weighting their importance in the training loss.
Formally, our egocentric pose estimator $\mathcal{F}_\mathrm{Pose}$ takes as input stereo egocentric images $\{\mathbf{I}_\mathrm{left},\mathbf{I}_\mathrm{right}\}$ and head pose $\mathbf{H}$, and predicts 3D keypoints
%
%%%%
\begin{equation}
    \boldsymbol{\zeta} = \mathcal{F}_\mathrm{Pose}(\mathbf{I}_\mathrm{left},\mathbf{I}_\mathrm{right}, \mathbf{H}) \in \mathbb{R}^{57\times 3}.
    \label{eq:pose_regressor}
\end{equation}
%%%%
%
Despite training data captured under uniform lighting, we employ severe color jittering in brightness, contrast, and color hue to augment input egocentric images, effectively compensating for illumination variations in test and real-world sequences.
Unlike the original FRAME model, which primarily addresses pose estimation under natural illumination, our egocentric system demonstrates universal applicability across diverse and challenging lighting conditions. 
%
%%%%%%%%%%%%%%%%%%%%%%%%%%%%%%%%%%%%%%%%%%%%%%%%%%%
%
\subsection{Inverse Kinematics} \label{sec:ik}
Now, given the 3D keypoint $\boldsymbol{\zeta}$, our Inverse Kinematics (IK) solves for a temporally plausible motion sequence $\{\boldsymbol{\theta}\}_1^T$ that best aligns the skeleton joints $\mathcal{K}(\{\boldsymbol{\theta}\}_1^T)$ with detected 3D joints $\{\boldsymbol{\zeta}\}_1^T$.
Following EgoAvatar~\cite{chen2024egoavatar}, we hierarchically optimize global 6D pose $\boldsymbol{\theta}_\mathrm{Rt}$, body motion $\boldsymbol{\theta}_\mathrm{body}$, and hand motion $\boldsymbol{\theta}_\mathrm{hand}$ by minimizing the energy function
%
%%%%
\begin{equation}
    {\arg \min}_{\{\boldsymbol{\theta}\}_1^T} E_\mathrm{Data}+E_\mathrm{Temporal}+E_\mathrm{DoFLimit}+E_\mathrm{Reg}.
    \label{eq:ik}
\end{equation}
%%%%
% 
However, accurately localizing 3D hand joints remains a challenge for our egocentric pose detector $\mathcal{F}_\mathrm{Pose}$ without incorporating temporal information, due to the hands' fine-grained motions and frequent self-occlusions.
Inspired by \citet{pavlakos2019expressive}, we leverage a PCA model for each hand to reduce the degree of freedom per hand and thereby constrain the hand motion $\boldsymbol{\theta}_\mathrm{hand} \in \mathbb{R}^{14}$ to a PCA subspace $\boldsymbol{\eta} \in \mathbb{R}^6$.
Moreover, by employing a temporal energy term 
%
%%%%
\begin{equation}
    E_\mathrm{Temporal} = \sum_\tau \sum_{h\in\{\mathrm{left,right}\}}||2\boldsymbol{\eta}_h^\tau - (\boldsymbol{\eta}_h^{\tau-1}+\boldsymbol{{\eta}}_h^{\tau+1})||_2
    \label{eq:ik-pca-temporal}
\end{equation}
%%%%
% 
to the PCA space, our method effectively re-weights the importance of different DoFs, resulting in temporally consistent hand gestures.
Please refer to the supplemental document for more details about other energy terms.
%
%%%%%%%%%%%%%%%%%%%%%%%%%%%%%%%%%%%%%%%%%%%%%%%%%%%
%
\subsection{Egocentric Depth Estimator} \label{sec:egodepth}
Skeleton motion reflects the coarse body movement, while lacking the fine clothing dynamics.
Given the recent advances in monocular depth estimation~\cite{khirodkar2025sapiens,yang2024depth}, we focus on extracting dense geometry signals from input stereo egocentric images.
Specifically, the depth map $\mathbf{d}$ is a pixel-aligned representation that captures 3D geometry of the visible surface.
Intuitively, depth can be directly measured from a calibrated camera system with a stereo matching algorithm.
However, stereo depth estimation suffers from the lack of detected cross-view correspondence, since 1) in egocentric perspective the body region co-observed by two cameras is limited especially under extreme head poses; and 2) finding correspondence of plain texture garment is challenging.
Instead, the generalizable depth estimator, i.e., DepthAnythingV2 \cite{yang2024depth}, is a reasonable and off-the-shelf relative depth prior.
We propose to extend it to an egocentric metric depth predictor.
Here, we first conduct multi-view and implicit stereo reconstruction, i.e., NeuS2~\cite{wang2023neus2}, to recover 3D surface of the human, which we then render back to the egocentric view.
The depth estimator backbone
%
%%%%
\begin{equation}
    \mathbf{d}_h = \mathcal{F}_\mathrm{Depth}(\mathbf{I}_h) \quad \mathrm{for} \enspace h\in \{\mathrm{left, right}\}
    \label{eq:ego-depth}
\end{equation}
%%%%
% 
is then person-specifically fine-tuned with the ground-truth metric depth map as supervision.
At inference time, we combine two unprojected depth maps into a point cloud $\mathbf{z}\in \mathbb{R}^{20000\times 3}$ and reconstruct the surface normal $\mathbf{n}_{\mathbf{z}}$ using KD tree search~\cite{hoppe1992surface}.
The depth map unprojection 
%
%%%%
\begin{equation}
    \mathbf{z} = (\mathbf{K}\boldsymbol{\Pi}\mathbf{H})^{-1} \mathbf{d}
    \label{eq:depth-unproj}
\end{equation}
%%%%
%
projects estimated point clouds from depth maps into global space, where $\mathbf{K}$ is the intrinsics of the egocentric camera and $\boldsymbol{\Pi}$ is the hand-eye calibration matrix.
To achieve generalization ability for unseen illuminations, we perform the same data augmentation as described in Sec.~\ref{sec:egopose}.
%
%%%%%%%%%%%%%%%%%%%%%%%%%%%%%%%%%%%%%%%%%%%%%%%%%%%
%
\subsection{Depth-conditioned Animatable Avatar} \label{sec:depth-driven-avatar}
%
%%%%
%
%%%%%%%%%%%%%%%%%%%%%%%%%%%%%%%%%%%%%%%%%%%%%%
%
\begin{figure}[t]
    \centering
    \includegraphics[width=0.48\textwidth]{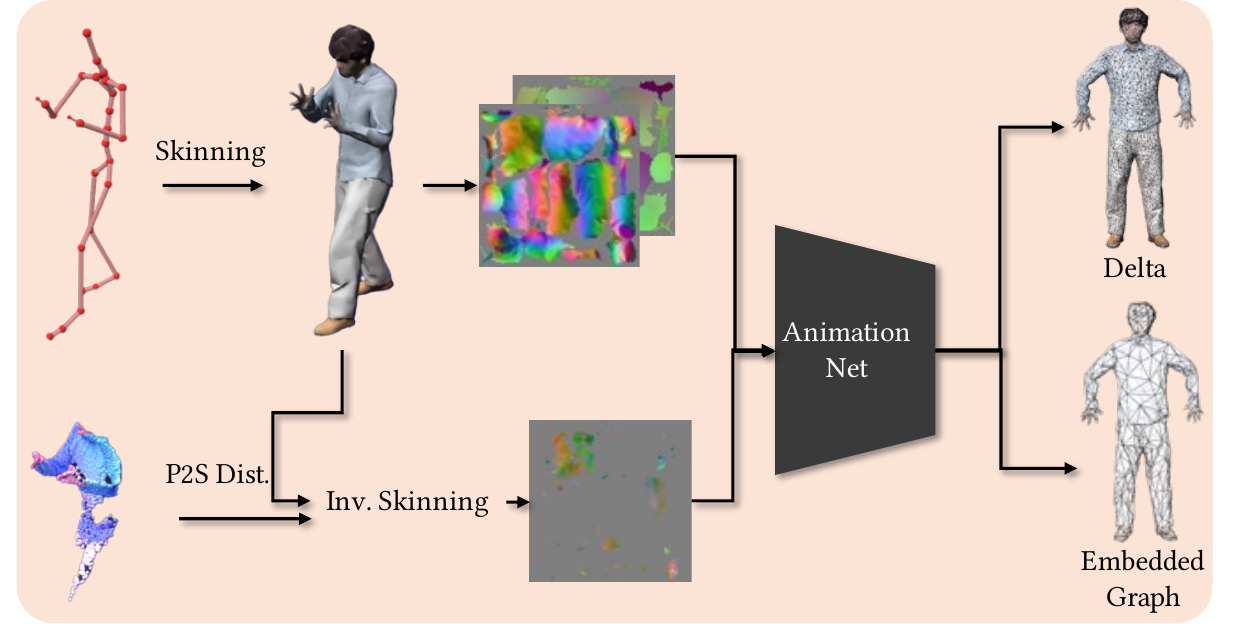}
    \caption{
    \textbf{Illustration of Depth-conditioned Animatable Avatar.} 
    Given paired egocentric skeleton motion and point clouds unprojected from the regressed egocentric depth map, we leverage the point to surface distance from the depth point cloud to the LBS animated template mesh as condition signal with canonical template normal and position map, and hierarchically predict embedded graph parameters and delta vertex displacements using our AnimationNet.
    }
    \label{fig:depth-cond-avatar}
\end{figure}
%
%%%%%%%%%%%%%%%%%%%%%%%%%%%%%%%%%%%%%%%%%%%%%%
%
%%%%
%
The egocentric perception modules extract geometry representation, i.e., skeleton motion and depth, of the 3D human as driving signals, whereas this section focuses on reconstructing the explicit full-body mesh of the character.
Deep Dynamic Characters~\cite{habermann2021real} has proven the success of a coarse-to-fine procedure in modeling motion-dependent body and cloth dynamics.
Given the recovered egocentric depth maps, one intuitive solution is to introduce an optimization procedure for registering the mesh surface of the animated character to the input depth.
However, such test-time optimization suffers from two main limitations: 1) it presumes the point cloud $\mathbf{z}$ as ground truth, which fails to compensate for depth estimation and unprojection errors, particularly noticeable in thin structures and fisheye-distorted regions, e.g., hands; and 2) it leads to a considerable runtime overhead.
Inspired by \textit{learning-based optimization} methods in non-rigid human registration~\cite{chen2024sequential, xiang2023drivable}, we encode the depth point cloud $\mathbf{z}$ into the vertex of the template mesh $\mathbf{\bar{V}}$ as
%
%%%%
{\footnotesize
\begin{equation}
    \boldsymbol{\xi}_i = \begin{cases}
    \mathcal{W}^{-1} (\mathbf{z}_j - \mathbf{v}_i) & \text{if } ||\mathbf{n}_{\mathbf{v}_i} \cdot \mathbf{n}_{\mathbf{z}_j}||_2 < \epsilon_n \text{ and } ||\mathbf{z}_j - \mathbf{v}_i||_2 < \epsilon_d,\\
    \mathbf{0} & \text{otherwise}.
  \end{cases}
    \label{eq:depth-encoding}
\end{equation}
}
%%%%
% 
Here, $\mathbf{v}_i$ is the vertex of the template mesh in the posed space $\mathcal{W}(\mathbf{\bar{V}})$ and $\mathbf{z}_j$ is the nearest neighboring point.
We define the encoded feature $\boldsymbol{\xi}$ as a surface-to-point distance unwarped to canonical space using inverse skinning $\mathcal{W}^{-1}$, which is equivalent to the gradient of the L2 point-to-surface loss.
False positive correspondence with geometric disparity larger than threshold $\epsilon_d$ and normal disparity larger than $\epsilon_n$ are filtered out.
\par
Moreover, to learn the character animation model end-to-end, we propose a \textit{one}-stage network that hierarchically predicts the embedded graph parameters $\boldsymbol{\alpha}, \boldsymbol{\beta}$ and the per-vertex offset $\mathbf{o}$.
Since the embedded graph and template mesh share one UV space, our key idea is to rasterize the vertex-level input signal, i.e., depth guidance $\boldsymbol{\xi}$ with temporal position and normal map $\boldsymbol{\Theta}$, in the UV space.
This spatially continuous representation enables us to regress the control parameters $\boldsymbol{\alpha}, \boldsymbol{\beta}, \mathbf{o}$ with a convolutional UNet, as illustrated in Fig.~\ref{fig:depth-cond-avatar}.
Formally, AnimationNet $\mathcal{G}_\mathrm{Anim}$ is defined as
%
%%%%
\begin{equation}
    \{\boldsymbol{\alpha}, \boldsymbol{\beta}, \mathbf{o}\} = \mathcal{G}_\mathrm{Anim}(\boldsymbol{\Theta}, \boldsymbol{\xi}).
    \label{eq:char-anim-net}
\end{equation}
%%%%
% 
\mhe{
The final posed and deformed mesh driven by egocentric stereo views can then be obtained by plugging the regressed pose (Eq.~\ref{eq:pose_regressor}) and deformation control parameters (Eq.~\ref{eq:char-anim-net}) into our character geometry equation (Eq.~\ref{eq:dqs}). 
This can be done for multiple frames resulting in a temporal mesh sequence denoted as $\{\mathbf{V}\}^T_1$ in the following.
}
Please refer to supplemental for the training details.
%
%%%%%%%%%%%%%%%%%%%%%%%%%%%%%%%%%%%%%%%%%%%%%%%%%%%
%%%%%%%%%%%%%%%%%%%%%%%%%%%%%%%%%%%%%%%%%%%%%%%%%%%
%
\section{Relightable Avatar Appearance Modeling and Rendering} \label{sec:relighting}
%
%%%%
\begin{figure*}[t]
    \centering
    \includegraphics[width=\textwidth]{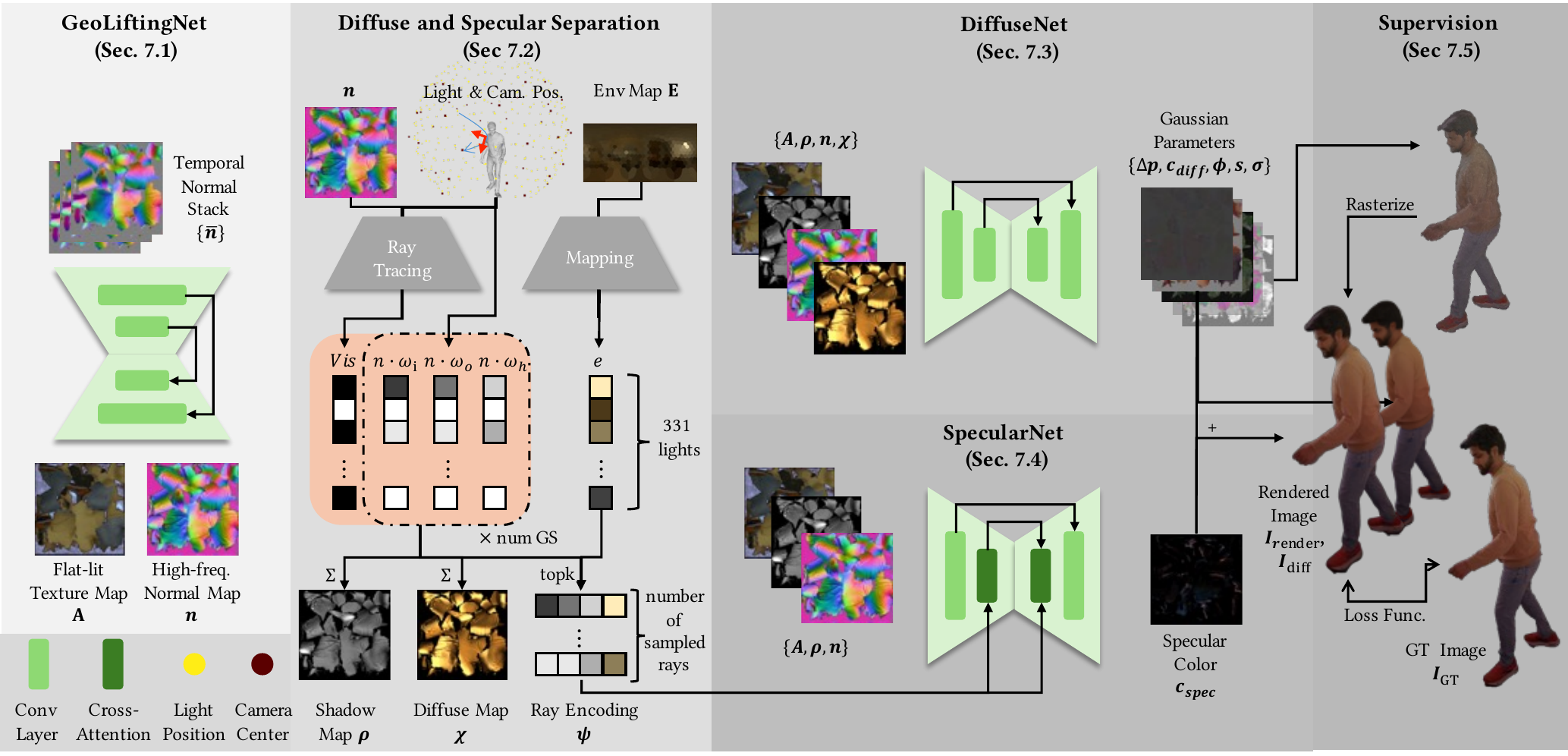}
    \caption{\textbf{Relightable Appearance Modeling Pipeline.} We create relightable photorealistic avatar in four major steps. Firstly, given a tracked mesh sequence, our GeoLiftingNet generates high-resolution UV normal and Flat-lit color texture (Sec.~\ref{sec:geo-lifting}). Then, we extract unbiased physically-informed feature for each local primitive given light and camera positions (Sec.~\ref{sec:phys-feat}). Finally, we model the diffuse and specular light transportation with Gaussian parameters separately using image-to-image translation UNet (in Sec.~\ref{sec:diffuse} and \ref{sec:specular}).}
    \label{fig:relight-pipeline}
\end{figure*}
%%%%
%
In this section, our relighting avatar model (see also Fig.~\ref{fig:relight-pipeline}) takes as input the tracked mesh sequence $\{\mathbf{V}\}^T_1$, and predicts a relightable and photorealistic avatar appearance that can be rendered to free viewpoints given a target illumination and camera view.
The recovered geometry, however, still does not match the ground truth due to limited template mesh resolution, motion-to-geometry mapping ambiguities (\eg stochastic wrinkle movement and fine-grained tracking error), and lack of observation from the input views.
Therefore, we first introduce the geometry lifting module (Sec.~\ref{sec:geo-lifting}) that generates a high-frequency normal map and a flat-lit texture map from the temporal mesh sequences as spatial clues to aid better normal reconstruction and tracking.
Then, we then introduce appearance and lighting features with a specular and diffuse separation (Sec. \ref{sec:phys-feat}).
Given those features, the diffuse light transport (Sec. \ref{sec:diffuse}) and specular light transport (Sec. \ref{sec:specular}) for 3D Gaussian splats are estimated using image-to-image translation networks.
The training target for the relightable appearance model is described in Sec. \ref{sec:relighting-loss}.
%
%%%%%%%%%%%%%%%%%%%%%%%%%%%%%%%%%%%%%%%%%%%%%%%%%%%
%
\subsection{Geometry Lifting}\label{sec:geo-lifting}
% 
%
%%%%%%%%%%%%%%%%%%%%%%%%%%%%%%%%%%%%%%%%%%%%%%
%
\begin{figure}[t]
    \centering
    \includegraphics[width=0.48\textwidth]{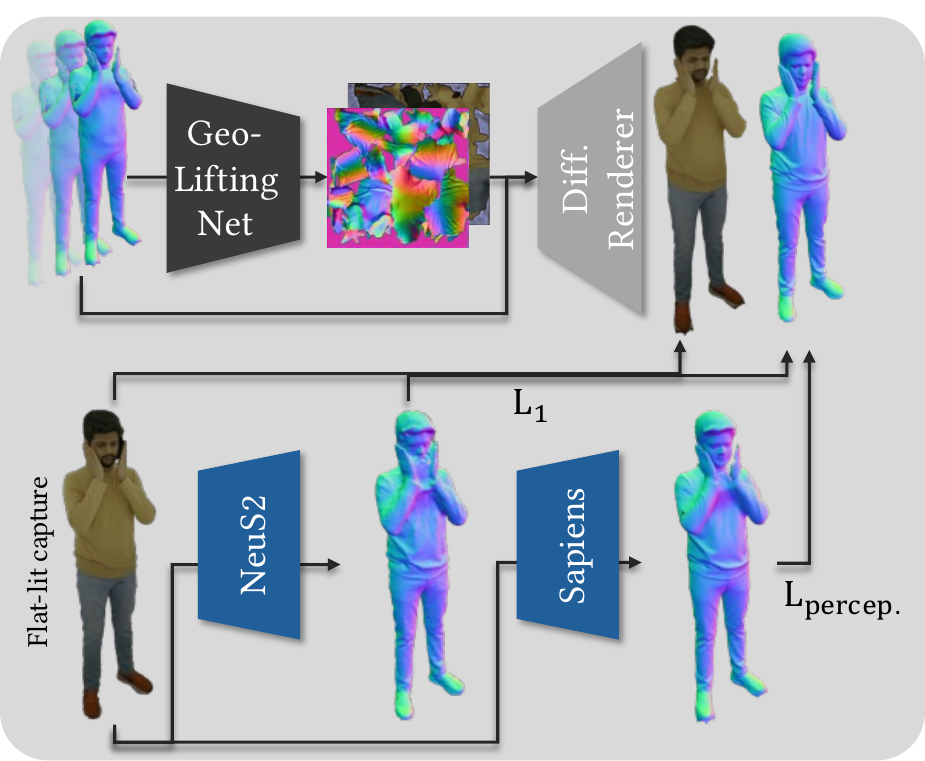}
    \caption{
    \textbf{Illustration of GeoLiftingNet.} 
    Given the temporal normal stack $\{\bar{\boldsymbol{n}}\}$ of the tracked mesh $\boldsymbol{V}$, we predict high frequency normal maps and fully-lit texture maps as an approximation of geometry and albedo in UV space. 
    We supervise the albedo prediction on the flat-lit multi-view images.
    The predicted normals are supervised against 3D consistent implicit normals~\cite{wang2023neus2} and more detailed image-space normals predicted by Sapiens~\cite{khirodkar2025sapiens}.
    }
    \label{fig:geo-lifting}
\end{figure}
%
%%%%%%%%%%%%%%%%%%%%%%%%%%%%%%%%%%%%%%%%%%%%%%
%
%  
In contrast to inverse rendering approaches for relightable avatars \cite{relightneuralactor2024eccv, chen2022relighting4d} that jointly estimate surface normals and material parameters, we suggest to model them in separate stages to eliminate geometry-material ambiguity.
The enhanced geometry is learned from interleaved flat-lit multi-view lightstage captures with sufficient brightness.
Taking a stacked normal map $\{\mathbf{\bar{n}}\}_{t-2}^t$ from the temporal mesh sequence $\{\mathbf{V}\}_{t-2}^t$ as input, our GeoLiftingNet 
% 
%%%%
\begin{equation}
    \{\mathbf{n}, \mathbf{a}\} = \mathcal{H}_\mathrm{Lift}(\mathbf{\bar{n}}_{t-2},\mathbf{\bar{n}}_{t-1},\mathbf{\bar{n}}_{t})
    \label{eq:geo-lifting}
\end{equation}
%%%%
%
predicts normal $\mathbf{n}$ and flat-lit appearance, i.e., albedo, $\mathbf{a}$ in the template mesh's texture space.
\mhe{Predicting albedo as well is motivated by the fact that the mesh sequence might suffer from on-surface drift with respect to the ground truth and, thus, albedo might sometimes be wrongly assigned.} 
\mhe{Instead, predicting it provides sufficient slack to the model in order to avoid wrong albedo assignments being wrongly compensated by other material parameters.}
\par
Unlike dynamic texture that can be directly supervised with a multi-view image loss, surface normal estimation in static illumination has been an open problem.
On the one hand, multi-view stereo  provides 3D reconstruction with 3D consistent surface normals, but oversmooth sufficient details, e.g., hand, face and frontal chest regions in Fig. \ref{fig:geo-lifting}.
On the other hand, screen-space normal estimator such as Sapiens \cite{khirodkar2025sapiens} extracts expressive fine-grained details, but lacks 3D consistency across views.
Therefore, based on a pixel-wise normal supervision with NeuS2 reconstructed meshes, we further introduce a perceptual loss to maintain similar visual quality of our normal rendering with Sapiens estimated normal, without enforcing an exact normal value.
This supervision strategy strikes a good balance between 3D consistency and detail preservation.
Concretely, we supervise our geometry lifting module $\mathcal{H}_\mathrm{Lift}(\cdot)$ with
%
%%%%
{\small
\begin{equation}
\begin{split}
    \mathcal{L}_\mathrm{Lift} 
    &= w_\mathrm{RGB}\mathcal{L}_\mathrm{L1}(\mathcal{D}(\mathbf{V}, \mathbf{a}),\mathbf{I}_\mathrm{FL}) \\
    &+ w_\mathrm{norm}\mathcal{L}_\mathrm{L1}(\mathcal{D}(\mathbf{V},\mathbf{n}),\mathbf{I}_\mathrm{Sapiens}) \\
    &+ w_\mathrm{percep}\mathcal{L}_\mathrm{percep}(\mathcal{D}(\mathbf{V},\mathbf{n}),\mathcal{D}(\mathbf{U},\mathbf{n}_{\mathbf{U}})),
    \label{eq:geolifting-loss}
\end{split}
\end{equation}
}
%%%%
%
where $\mathcal{D}$ is a differentiable renderer implemented by NVDiffrast \cite{Laine2020diffrast}, $\mathbf{U}$ is the mesh reconstruction per frame, $\mathbf{n}_{\mathbf{U}}$ is its normal map, and $\mathbf{I}_\mathrm{Sapiens}$ are the screenspace normals estimated by Sapiens.
%
%%%%%%%%%%%%%%%%%%%%%%%%%%%%%%%%%%%%%%%%%%%%%%%%%%%
%
\subsection{Diffuse and Specular Separation}\label{sec:phys-feat}
Given the refined geometry, now the main challenge of building a relightable avatar lies in effectively modeling the local light transport function, so that each local primitive generates realistic shading color, i.e., reflected radiance, according to the given input illumination.
In the upper hemisphere of an isotropic surface, two spherical functions of the incoming light direction and the outgoing view direction above the surface can be uniquely represented as a triplet of angles $\{\mathbf{n}\cdot\boldsymbol{\omega}_i,\mathbf{n}\cdot\boldsymbol{\omega}_o,\mathbf{n}\cdot\boldsymbol{\omega}_h\}$ \cite{blinn1977models, cook1982reflectance}, where the half-vector
%
%%%%
\begin{equation}
    \boldsymbol{\omega}_h = \frac{\boldsymbol{\omega}_i+\boldsymbol{\omega}_o}{||\boldsymbol{\omega}_i+\boldsymbol{\omega}_o||}
    \label{eq:half-vec}
\end{equation}
%%%%
%
is the mean of light and viewing direction.
The physics-based rendering equation \cite{kajiya1986rendering} analytically models the reflected radiance as
% 
%%%%
{\footnotesize
\begin{align}
    L_o(\mathbf{x}, \boldsymbol{\omega}_o) &= \int_{\Omega_+} L_i(\mathbf{x}, \boldsymbol{\omega}_i) \mathcal{T}(\mathbf{x}, \boldsymbol{\omega}_i, \boldsymbol{\omega}_o) (\mathbf{n}\cdot\boldsymbol{\omega}_i) d\boldsymbol{\omega}_i \label{eq:pbr} \\
    &= \int_{\Omega_+} L_i(\mathbf{x}, \boldsymbol{\omega}_i) (\frac{\mathcal{A}(\mathbf{x})}{\pi}+\mathcal{T}_\text{spec}(\mathbf{x}, \boldsymbol{\omega}_i, \boldsymbol{\omega}_o) ) (\mathbf{n}\cdot\boldsymbol{\omega}_i) d\boldsymbol{\omega}_i \label{eq:pbr-ds}
\end{align}}
%%%%
%
which we can rewrite as Eq.~\ref{eq:pbr-ds} by separating the Lambertian part from the radiance transfer function. 
Here, $L_i, L_o$ denotes the incoming and outgoing radiance on point $\mathbf{x}$, $\mathcal{T}$ denotes the radiance transfer functions (known as BRDFs) and $\mathcal{A}$ is the albedo/diffuse color.
Separating the view direction-agnostic diffuse shading from the total shading function can simplify the computation and facilitate the learning of material properties.
The goal of this subsection is to introduce appearance and lighting quantities that can fully depict and guide the learning of light transport in Eq.~\ref{eq:pbr-ds}.
\par
Given an HDR environment map $\mathbf{E}$, we first map it into light positions in the lightstage and average the light intensities $\mathbf{e}\in \mathbb{R}^{331\times 3}$.
For each texel in the UV map of the mesh, we retrieve the 3D surface point $\mathbf{p}$ using barycentric interpolation.
Then, OptiX \cite{parker2010optix} traces the light visibility
%
%%%%
\begin{equation}
    \mathbf{1}_{(\mathbf{p},\boldsymbol{\omega}_i)} = \mathtt{ray\_trace} (\mathbf{V}, \mathbf{n}, \mathbf{p}+\delta\mathbf{\bar{n}},\boldsymbol{\omega}_i)
    \label{eq:optix}
\end{equation}
%%%%
%
for each texel, i.e., $\mathbf{p}$, concerning the self-occlusion of the mesh $\mathbf{V}$, where $\delta$ is an infinitesimal that avoids self-intersection and $\mathbf{1}_{(\cdot,\cdot)}$ is a binary function indicating no intersection between the light and the mesh surface.
With sparse light sources and light visibilities, we measure the diffuse map in flat-lit frames as $\boldsymbol{\rho}$ and HDR-lit frames as $\boldsymbol{\chi}$ that are computed as
% 
%%%%
\begin{equation}
    \boldsymbol{\rho}_{\mathbf{p}} = \sum_{\boldsymbol{\omega}_i} \mathbf{1}_{(\mathbf{p},\boldsymbol{\omega}_i)} (\mathbf{n}\cdot\boldsymbol{\omega}_i);
    \quad
    \boldsymbol{\chi}_{\mathbf{p}} = \sum_{\boldsymbol{\omega}_i} \mathbf{1}_{(\mathbf{p},\boldsymbol{\omega}_i)} \boldsymbol{e}_{\boldsymbol{\omega}_i} (\mathbf{n}\cdot\boldsymbol{\omega}_i).
    \label{eq:diff-map}
\end{equation}
%%%%
%
The diffuse map $\boldsymbol{\rho}$ in flat-lit illumination acts as a shadow map, which assists in albedo recovery where we jointly predict the flat-lit texture $\mathbf{a}$.
\par 
In contrast, with an additional spherical function $\omega_i$, it is non-trivial to pre-integrate a specular map similar to the diffuse term.
Existing solutions rely on pre-integrated analytical BRDFs, e.g., \cite{burley2012physically} or PRT \cite{sloan2023precomputed}, which is a lossy compression of the full surface reflectance function, and requires hyperparameter tuning for spatially-varying materials.
On the other hand, learning the light transport function for each light~\cite{zhang2024relitlrm,singh2025relightable} is computationally expensive and redundant, whereas a majority of lights are under occlusion.
Therefore, we explore a ray sampling technique that selects the lights potentially lying in the specular lobe, and learn the specular light transport from a 6D encoding for each incoming-outgoing radiance pair.
Particularly, the ray encoding $\boldsymbol{\psi}$ is represented as a 6D vector:
%
%%%%
{\small
\begin{equation}
    \boldsymbol{\psi}_{\mathbf{p}} = \{\mathbf{n}\cdot\boldsymbol{\omega}_i,\mathbf{n}\cdot\boldsymbol{\omega}_o,\mathbf{n}\cdot\boldsymbol{\omega}_h, \mathbf{1}_{(\mathbf{p},\boldsymbol{\omega}_i)} \mathbf{e}_{\boldsymbol{\omega}_i} (\mathbf{n}\cdot\boldsymbol{\omega}_i)\} \quad \mathrm{for} \enspace (\boldsymbol{\omega}_i, \boldsymbol{\omega}_o) \in \mathcal{P},
    \label{eq:spec-map}
\end{equation}}
%%%%
%
where $\mathcal{P}$ is the set of $r$ sampled rays with top-$r$ response 
% 
%%%%
\begin{equation}
    L_\text{spec} = (\max(0,\mathbf{n}\cdot\boldsymbol{\omega}_h))^\alpha \mathbf{1}_{(\mathbf{p},\boldsymbol{\omega}_i)} ||\mathbf{e}_{\boldsymbol{\omega}_i}|| (\mathbf{n}\cdot\boldsymbol{\omega}_i)
    \label{eq:blinn-phong-spec}
\end{equation}
%%%%
%
given the Blinn-Phong model~\cite{blinn1977models} as an importance score.
As we are only interested in the order of the importance rather than the actual importance score, our model is less sensitive to the hyperparameter $\alpha$.
%
%%%%%%%%%%%%%%%%%%%%%%%%%%%%%%%%%%%%%%%%%%%%%%%%%%%
%
\subsection{Diffuse Light Transport Modeling} \label{sec:diffuse}
Diffuse shading is view-independent and, thus, only the incoming light intensity, surface geometry, and material are relevant for its computation.
Therefore, we reconstruct diffuse light transport by regressing the diffuse shading color of Gaussian primitives $\mathbf{c}_\mathrm{diff}$ given the diffuse map $\boldsymbol{\chi}$, flat-lit texture $\mathbf{a}$, shadow map $\boldsymbol{\rho}$, and high-frequency normal map $\mathbf{n}$.
To recover diffuse reflectance, we propose DiffuseNet 
%
%%%%
\begin{equation}
    \{\Delta \mathbf{p}, \mathbf{c}_\mathrm{diff}, \boldsymbol{\phi}, \mathbf{s},\boldsymbol{\sigma}\} = \mathcal{H}_\mathrm{Diff}(\mathbf{a}, \boldsymbol{\rho}, \mathbf{n}, \boldsymbol{\chi}),
    \label{eq:diff-net}
\end{equation}
%%%%
%
which is an efficient convolutional neural network (CNN) operating in texture space.
Further, as the major energy of reflectance function for the material of the human avatar (\eg garment, skin, etc.) lies in the diffuse part, we jointly predict the Gaussian parameters with diffuse shading colors. 
%
%%%%%%%%%%%%%%%%%%%%%%%%%%%%%%%%%%%%%%%%%%%%%%%%%%%
%
\subsection{Specular Light Transport Modeling} \label{sec:specular}
Similar to the diffuse light transport modeling, we aim at translating the albedo and normal features $\{\mathbf{a}, \boldsymbol{\rho}, \mathbf{n}\}$ into specular shading $\mathbf{c}_\mathrm{spec}$ given a target illumination condition $\boldsymbol{\psi}$.
However, due to the assumption that each incident light independently contributes to the shading color, we consider specular feature $\boldsymbol{\Psi}=\{\boldsymbol{\psi}_p\}_1^r$ to be permutation invariant in the axis of sampled rays.
Therefore, it is not suitable to process $\boldsymbol{\Psi}$ as a stacked feature using CNNs.
Instead, we adapt the cross-attention mechanism \cite{vaswani2017attention} to fuse them into feature maps learned from per-primitive feature stacks $\{\mathbf{a},\boldsymbol{\rho},\mathbf{n}\}$, which effectively handle different light directions in parallel.
Our proposed SpecularNet $\mathcal{H}_\mathrm{Spec}(\cdot)$ consists of CNNs similar to $\mathcal{H}_\mathrm{Diff}(\cdot)$, with interleaved cross-attention layers.
The last layer leverages a softplus activation function to produce the non-negative specular shading
%
%%%%
\begin{equation}
    \mathbf{c}_\mathrm{spec} = \mathcal{H}_\mathrm{Spec}(\mathbf{a}, \boldsymbol{\rho}, \mathbf{n}; \boldsymbol{\Psi}).
    \label{eq:color_spec}
\end{equation}
%%%%
%
\par
Particularly, in the scenario where we sample multiple incident lights for each primitive, the one-to-many attention \cite{flynn2024quark} is introduced to improve the efficiency of the cross-attention layer without sacrificing expressiveness.
The cross-attention layer operates between the per-primitive feature descriptor $\mathbf{f}'_{\mathbf{q}}$ and  ray encodings $\boldsymbol{\Psi}'_{\mathbf{q}}$ at the corresponding UV position $\mathbf{q}$, with $\mathbf{f}'_{\mathbf{q}}$ as query and $\boldsymbol{\Psi}'_{\mathbf{q}}$ as keys and values.
The specular feature is only linearly mapped once $\boldsymbol{\Psi}'_{\mathbf{q}} = \mathbf{W}^K \boldsymbol{\Psi}_{\mathbf{q}}$.
In each attention layer, compared to typical multi-head attention, the one-to-many attention folds the linear layer of keys and values into query side.
Consequently, this approach avoids the redundant computation inherent in mapping multiple keys and values to multiple heads, while retaining the benefit of multi-head attention with reduced channel dimensions in dot-product computation.
Formally, the one-to-many attention is described as
%
%%%%
\begin{equation}
    \mathrm{Attention}(\mathbf{f}'_{\mathbf{q}},\boldsymbol{\Psi}'_{\mathbf{q}}) = \mathrm{softmax}(\frac{(\mathbf{W}^Q\mathbf{f}'_{\mathbf{q}})^T\boldsymbol{\Psi}'_{\mathbf{q}}}{\sqrt{d_h}}) \boldsymbol{\Psi}'_{\mathbf{q}}
\end{equation}
%%%%
%
where $\mathbf{W}^Q$ is the weight of a linear layer, $d_h$ is the number of channel per head.
%
%%%%%%%%%%%%%%%%%%%%%%%%%%%%%%%%%%%%%%%%%%%%%%%%%%%
%
\subsection{Training Target} \label{sec:relighting-loss}
To supervise the predicted Gaussian parameters from Eq. \ref{eq:diff-net} and \ref{eq:color_spec}, we render 3D Gaussians into an image $\mathbf{I}_\mathrm{render}$ with final shading $\mathbf{c}_\mathrm{spec}+\mathbf{c}_\mathrm{diff}$ using Eq.~\ref{eq:gaussian_render}.
In addition, we render Gaussians with diffuse shading $\mathbf{c}_\mathrm{diff}$ individually as $\mathbf{I}_\mathrm{diff}$.
The DiffuseNet $\mathcal{H}_\mathrm{Diff}$ and SpecularNet $\mathcal{H}_\mathrm{Spec}$ are jointly trained from HDR-lit frames in the lightstage capture with photometric supervision
% 
%%%%
{ \small
\begin{equation}
    \begin{split}
        \mathcal{L}_\mathrm{relight} &= w_\mathrm{L1}\mathcal{L}_\mathrm{L1}(\mathbf{I}_\mathrm{render},\mathbf{I}_\mathrm{GT})+ w_\mathrm{SSIM} \mathcal{L}_\mathrm{SSIM}(\mathbf{I}_\mathrm{render},\mathbf{I}_\mathrm{GT}) \\
        &+ w_\mathrm{IDMRF} \mathcal{L}_\mathrm{IDMRF}(\mathbf{I}_\mathrm{render},\mathbf{I}_\mathrm{GT}) + w_\mathrm{Reg} \mathcal{L}_\mathrm{Reg}(\Delta \mathbf{p}, \mathbf{s})  \\
        &+ w_\mathrm{diff} \mathcal{L}_\mathrm{L1}(\mathbf{I}_\mathrm{diff},\mathbf{I}_\mathrm{GT}),
        \label{eq:relighting-loss}
    \end{split}
\end{equation}
}
%%%%
%
where $\mathbf{I}_\mathrm{GT}$ is the ground truth image and $w_{(\cdot)}$ re-weights different loss terms.
Without an explicit specular and diffuse separation, we additionally apply a supervision of the diffuse image with L1 loss, preventing specular shading to overfit view-independent color.
Despite common image losses and perceptual losses, i.e., ID-MRF loss \cite{wang2018image}), we further enforce strong regularization on the offset and scale parameter of Gaussian spheres, as we lack reliable background matting in HDR-lit frames to regularize the Gaussian shapes.
The concrete formulation of $\mathcal{L}_\mathrm{reg}$ and hyper-parameter setting are discussed in the supplemental document.
%
%%%%%%%%%%%%%%%%%%%%%%%%%%%%%%%%%%%%%%%%%%%%%%%%%%%
%%%%%%%%%%%%%%%%%%%%%%%%%%%%%%%%%%%%%%%%%%%%%%%%%%%
%
\section{Affordable Egocentric HDR Environment Map Capture} \label{sec:env-map-capture}
%
%%%%
%
%%%%%%%%%%%%%%%%%%%%%%%%%%%%%%%%%%%%%%%%%%%%%%
%
\begin{figure}[t]
    \centering
    \includegraphics[width=0.49\textwidth]{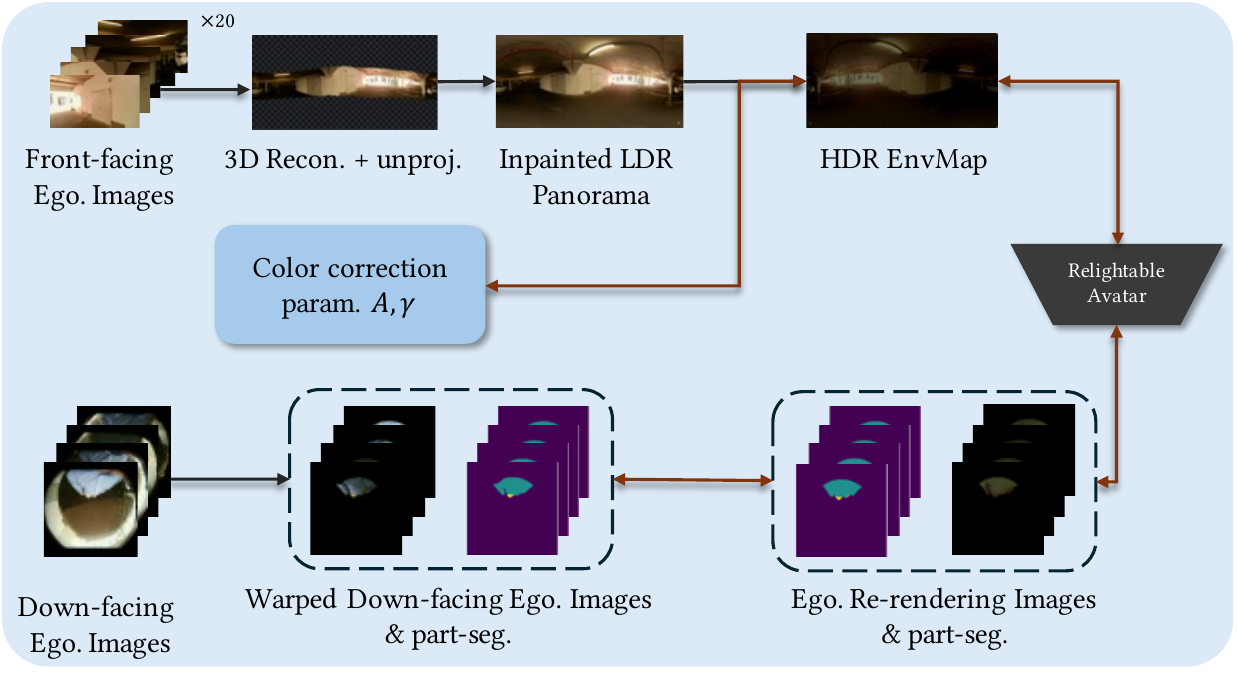}
    \caption{
    \textbf{Illustration of Egocentric HDR Environment Map Capture Pipeline.} 
    We first compose the 360 degree environment scan from front-facing egocentric camera into an LDR panorama image. 
    Then, the LDR environment map is lifted into HDR and color calibrated via inverse rendering of the pre-trained relightable avatar into down-facing egocentric images.
    }
    \label{fig:inv-render}
\end{figure}
%
%%%%%%%%%%%%%%%%%%%%%%%%%%%%%%%%%%%%%%%%%%%%%%
%
%%%%
%
With a photorealistic and relightable avatar, users can seamlessly integrate themselves into virtual environments like gaming. 
However, to enable Mixed Reality applications, the avatar must realistically interact with the physical world, which requires accurate illumination estimation.
In this section, we propose an affordable method for capturing an HDR environment map from an approximate 10-second in-place scan using the front-facing cameras of the HMD, as shown in Fig.~\ref{fig:inv-render}. 
\par 
Concretely, from a 360 degree scan of the environment from an egocentric front-facing camera with a fixed exposure setting, we uniformly sample 20 images to reconstruct a 3D scene~\cite{wang2025vggt} and unproject the images into an axis-aligned panorama environment map $\mathbf{E}_o$.
Then, we leverage a diffusion model~\cite{gemni2025gemini} to inpaint the unobserved region of the captured environment map $\mathbf{E}_o$, noted as $\mathbf{E}_l$.
However, the color space of the LDR environment maps $\mathbf{E}_l$ captured by front-facing cameras and the HDR environment maps used in the lightstage to train our appearance modules do not match.
Thus, we have to convert the LDR maps into the color space used to train our avatar model.
Therefore, we align the LDR environment map $\mathbf{E}_l$ to a plausible HDR environment map $\mathbf{E}_h$ via an inverse rendering process leveraging the relightable character and the down-facing egocentric image.
The mapping function $\mathcal{C}$ between LDR and HDR environment maps is parameterized by a group of color correction parameters $\mathbf{A}$~\cite{finlayson2015color} and gamma parameters $\boldsymbol{\gamma}$.
%
%%%%
\begin{equation}
    \mathbf{E}_h = \mathcal{C}(\mathbf{E}_l; \mathbf{A},\boldsymbol{\gamma})
    \label{eq:color-correct}
\end{equation}
% We explain the parameterization details in Appendix \jcc{x.x}.
%%%%
%
% \mhc{unclear what C is ?}
%
% \mhc{give a high-level intuition of the idea here. The idea is to leverage the human avatar model as a calibration target. and by rendering it into the ego view you can optimize for the color correction parameters.}
%

To recover the environment map directly from the HMD, we leverage the captured user as a calibration target, optimizing color correction parameters by rendering the pre-trained relightable avatar into the egocentric view.
Concretely, given a pre-trained relightable avatar $\mathcal{H}$ as a composition of models introduced in Sec.~\ref{sec:relighting} with tracked mesh surface $\mathbf{V}$ and camera pose $\mathbf{K\Pi H}$, we optimize $\mathbf{A}$ and $\boldsymbol{\gamma}$ using the objective function
%
%%%%
\begin{align}
    &\arg \min _{\mathbf{A},\boldsymbol{\gamma}} \sum_{s\in \mathcal{M}} \mathcal{L}_\mathrm{L1}(\mathbf{I}_\mathrm{render}\mathbf{M}_s, \mathcal{O}^{-1}(\mathbf{I}_\mathrm{ego})\mathbf{M}_s)+\mathcal{L}_\mathrm{Reg}(\mathbf{E}_h)
    \label{eq:inv-render} \\
    &\mathbf{I}_\mathrm{render} = \mathcal{H}(\mathbf{V}, \mathbf{K}\boldsymbol{\Pi}\mathbf{H} ; \mathcal{S}(\mathbf{E}_h)).
    \label{eq:forward-render}
\end{align}
%%%%
%
Here, $\mathcal{S}$ is the average pooling operation to sample the directional light intensity from the environment map.
$\mathcal{O}$ is the inverse warping of optical flow~\cite{teed2020raft} we estimated between the re-rendered image and the input egocentric down-facing image, which alleviates the tracking error of the mesh geometry.
$\mathbf{M}_s$ denotes the intersected segmentation masks of $\mathbf{I}_\mathrm{render}, \mathbf{I}_\mathrm{ego}$ for body parts $s$.
Since the egocentric view is greatly dominated by upper body, the introduction of part segmentation mask prevents our model from overfitting particular body tones. 
$\mathcal{L}_\mathrm{Reg}$ regularizes the pixel value in $[0,1]$.
Since our relightable avatar is learned from multiview linear-mapped HDR images, rendering into the color space of egocentric front-facing and down-facing cameras requires additional color synchronization, which is a differentiable process we omitted in Eq.~\ref{eq:forward-render}.
%
%%%%%%%%%%%%%%%%%%%%%%%%%%%%%%%%%%%%%%%%%%%%%%%%%%%
%%%%%%%%%%%%%%%%%%%%%%%%%%%%%%%%%%%%%%%%%%%%%%%%%%%
%
%
%%%%%%%%%%%%%%%%%%%%%%%%%%%%%%%%%%%%%%%%%%%%%%%%%%%
%%%%%%%%%%%%%%%%%%%%%%%%%%%%%%%%%%%%%%%%%%%%%%%%%%%
%
\section{Experiment} \label{sec:exp}
In this section, we conduct comprehensive experiments to demonstrate the capability of our egocentric driven and relightable full-body avatar.
First, we introduce the evaluation setup (Sec.~\ref{sec:exp-setup}) and show qualitative results of our method (Sec.~\ref{sec:exp-qualitative}). 
Then, we evaluate our egocentric capture performance (Sec.~\ref{sec:exp-geo}).
Next, we perform evaluations for our relightable avatar representation (Sec.~\ref{sec:exp-relightable}).
Finally, we also compare our lighting estimation with competing methods and baselines (Sec.~\ref{sec:exp-env}).
%
%%%%%%%%%%%%%%%%%%%%%%%%%%%%%%%%%%%%%%%%%%%%%%%%%%%
%%%%%%%%%%%%%%%%%%%%%%%%%%%%%%%%%%%%%%%%%%%%%%%%%%%
%
\subsection{Experimental Setup} \label{sec:exp-setup}
%
%%%%%%%%%%%%%
%
\subsubsection{Dataset.} We collect four different subjects in total with each subject following the data capture scheme described in Sec.~\ref{sec:dataset}.
Our lightstage contains 331 individually controllable RGB light sources and 40 4K HDR cameras, where we hold-out 3 cameras for testing while 37 are used for training.
Further, all qualitative and quantitative results are performed on the testing sequence (see also the last row in Tab.~\ref{tab:data-scheme}).
All videos are captured in RAW \footnote{\url{https://www.red.com/red-tech/redcode-raw}} format, which are then tone-mapped and compressed to MP4.
Paired with the external multi-view camera system, we provide a synchronized stereo down-facing egocentric camera at 720p using a fixed exposure.
Due to the limited brightness in the lightstage, we measure the head trajectory via the triangulated head-mounted ArUco markers instead of relying on the SLAM pose of headset.
Other post-processing procedures, including per-frame background matting and 3D reconstruction, follow \citet{chen2024egoavatar}.
In addition to the studio testing sequences, we record two in-the-wild testing sequences, with synchronized stereo down-facing cameras at 720p, stereo front-facing cameras at 1080p, and HMD SLAM head pose.
%
%%%%%%%%%%%%%
%
\subsubsection{Competing Methods.} 
Due to the lack of baseline that operate under the same assumptions and conditions, i.e., full-body avatar relighting from egocentric views trained with lightstage data, we choose two sets of comparisons.
Firstly, we compare with motion-driven relightable avatars~\cite{chen2022relighting4d, chen2024meshavatar}.
For fair comparison, their models take as input the ground-truth environment map and only learn the intrinsic decomposition on re-lit frames.
In addition, we combine a non-relightable Gaussian-based egocentric avatar~\cite{chen2024egoavatar} with a personalized generative image-to-image relighting module~\cite{jin2024neural} following the path of ~\citet{he2024diffrelight}.
Since these methods do not recover skeletal motion from egocentric views, we provide them the motion and geometry prediction from our method (Sec.~\ref{sec:ego-preception} and ~\ref{sec:depth-driven-avatar}), in order to isolate the relighting quality with egocentric tracking performance, which we study separately in Sec.~\ref{sec:exp-geo}.
We cannot compare to \citet{wang2025relightable} since their code is not publicly available.

%%%%%%%%%%%%%
%
\subsubsection{Metrics.} 
We follow the general practices in evaluating rendering quality using Peak Signal-to-Noise Ratio (PSNR), Structural Similarity Index (SSIM)~\cite{wang2004image}, and Learned Perceptual Image Patch Similarity (LPIPS)~\cite{zhang2018perceptual}. 
Since our evaluation is conducted on a novel test sequence without clues to reconstruct all visual details, we further introduce the generative metric FID~\cite{heusel2017gans} for distributional similarity between rendered and ground-truth images, which measures visual plausibility.
%
%%%%%%%%%%%%%%%%%%%%%%%%%%%%%%%%%%%%%%%%%%%%%%%%%%%
%%%%%%%%%%%%%%%%%%%%%%%%%%%%%%%%%%%%%%%%%%%%%%%%%%%
%
\subsection{Qualitative Results} \label{sec:exp-qualitative}
%
%
%%%%
%
%%%%%%%%%%%%%%%%%%%%%%%%%%%%%%%%%%%%%%%%%%%%%%
%
\begin{figure}[t]
    \centering
    \includegraphics[width=.49\textwidth]{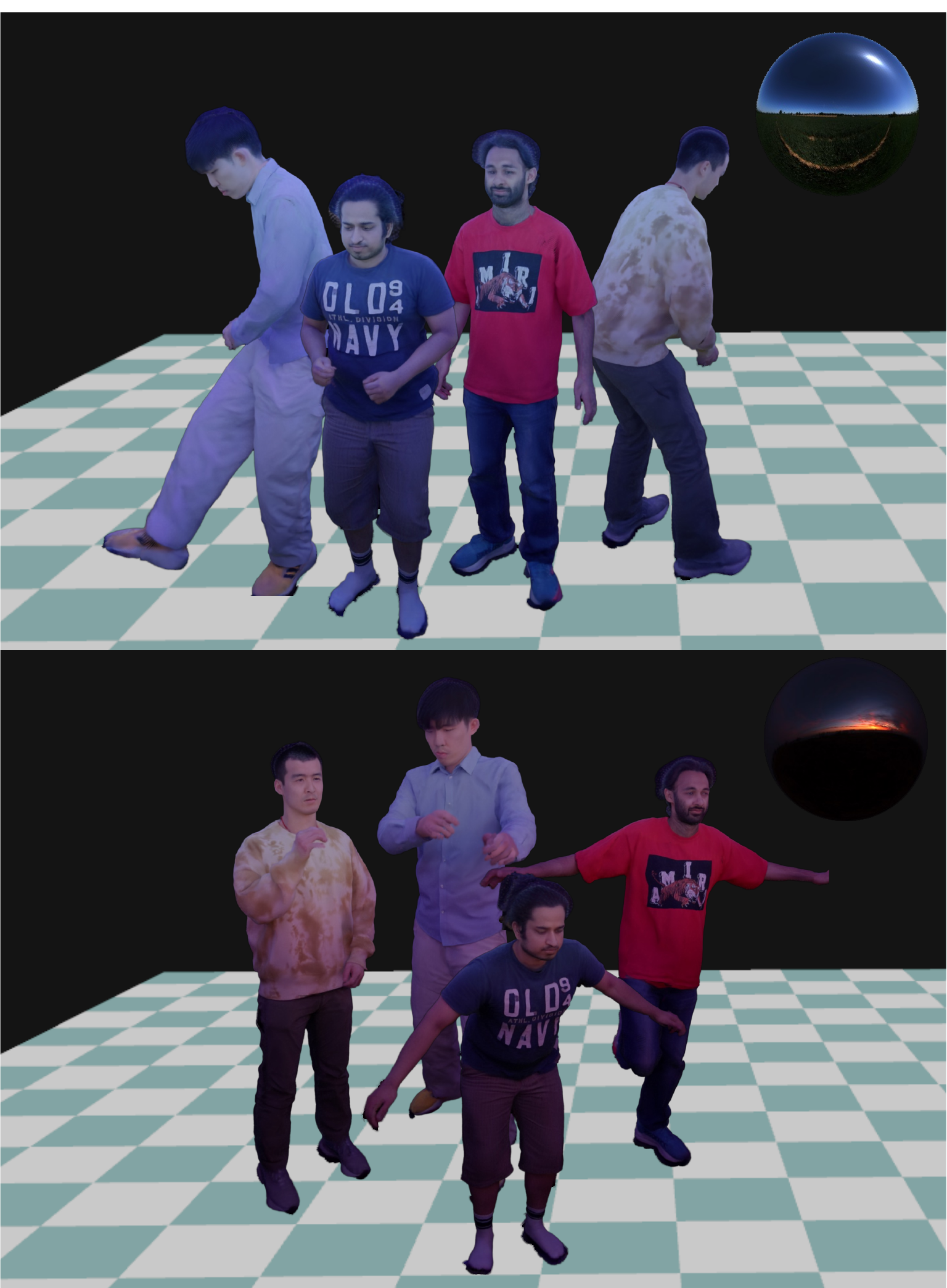}
    \caption{
    \textbf{Qualitative Results.} 
    Our approach captures human performance under novel illumination and renders photorealistic relightable appearance given arbitrary lighting conditions.
    }
    \label{fig:exp-qual}
\end{figure}
%
%%%%%%%%%%%%%%%%%%%%%%%%%%%%%%%%%%%%%%%%%%%%%%
%
%%%%
%
\mhe{In Fig.~\ref{fig:exp-qual}, we show qualitative results of our method. 
Concretely, we relight multiple egocentric driven avatars under the same target lighting illumination. 
Note that our method can preserve high-frequency appearance while accounting for light-dependent changes that are overall consistent across various subjects. 
For more qualitative results, we also refer to our supplemental video, which further demonstrates the temporal and view consistency of our method.}
%
%%%%%%%%%%%%%%%%%%%%%%%%%%%%%%%%%%%%%%%%%%%%%%%%%%%
%%%%%%%%%%%%%%%%%%%%%%%%%%%%%%%%%%%%%%%%%%%%%%%%%%%
%
\subsection{Evaluation on Egocentric Geometry Reconstruction} \label{sec:exp-geo}
%
%%%%%%%%%%%%%
%
\subsubsection{Comparison and Ablation on Egocentric Mesh Capture}
%
%%%%
\begin{table}[t!]
    \centering
    \caption{\textbf{Quantitative Comparison on Egocentric Avatar Reconstruction.} $\ddagger$: We adapt \textit{EgoDeformer} from EgoAvatar~\shortcite{chen2024egoavatar} into our stereo setup. Distances are in centimeter scale.}
    \label{tab:ego-tracking}
    \begin{tabular}{|c|c|c|c|}
        \hline
        \multirow{2}{*}{Method} & \multicolumn{2}{c|}{Point-to-surface Distance $\downarrow$} & \multirow{2}{*}{FPS $\uparrow$}\\
        \cline{2-3}
        &Full-body & \makecell{Egoview \\ Visible Surface}&\\
        \hline
        DDC~\shortcite{habermann2021real} & 1.27& 1.24& \textbf{96.09}\\
        \hline
        EgoAvatar~\shortcite{chen2024egoavatar} $\ddagger$ & \textbf{1.22}& \textbf{1.11}&0.006\\
        \hline
        Ours w/o depth cond. & 1.29& 1.24&\underline{75.71}\\
        \hline 
        \textbf{Ours} & \underline{1.23}& \underline{1.12}&46.45\\ 
        \hline
    \end{tabular}
\end{table}
%%%%
%
%%%%
\begin{figure*}[t]
    \centering
    \includegraphics[width=\textwidth]{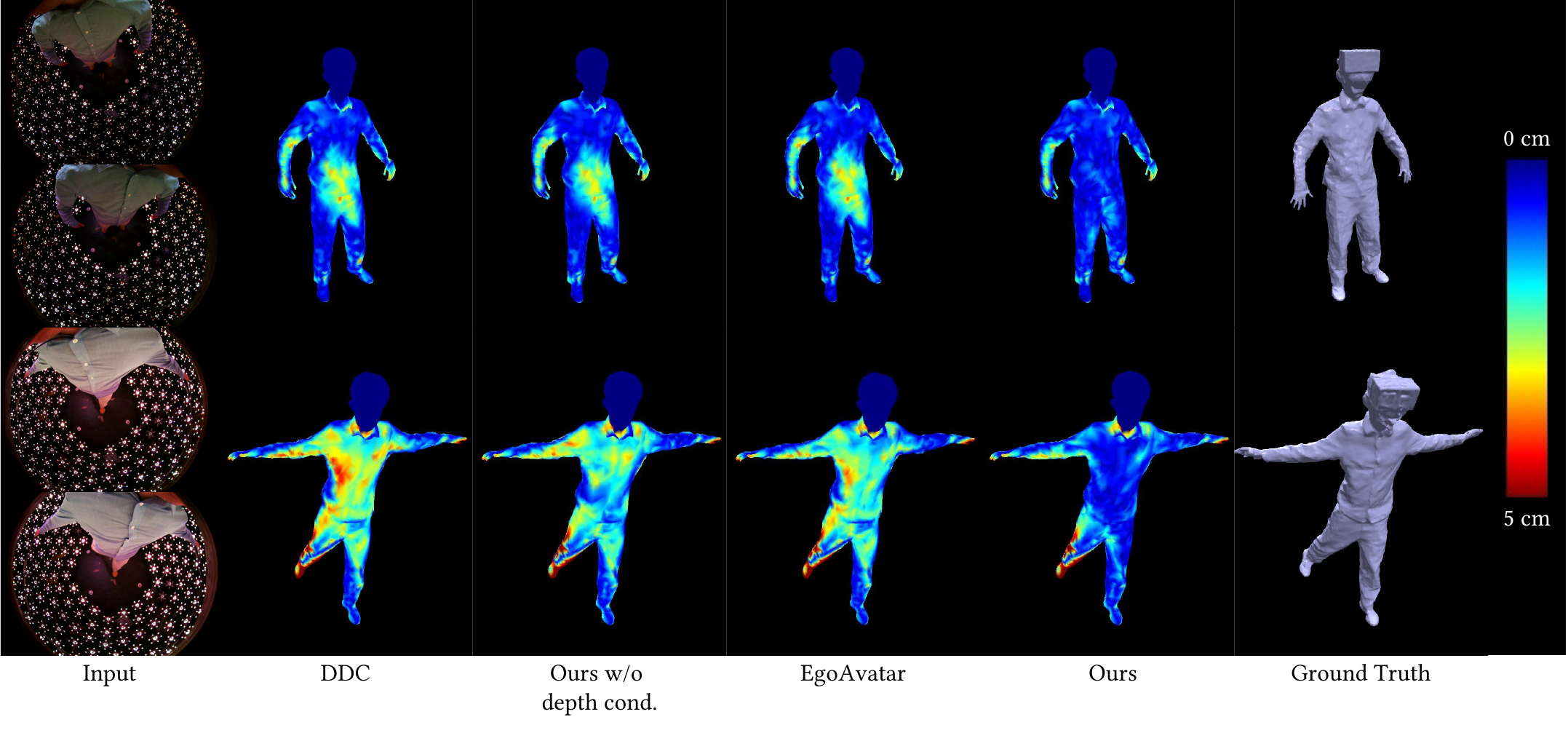}
    \caption{\textbf{Qualitative Ablation on Egocentric Mesh Capture.} Our model shows better geometry reconstruction quality under different lighting, particularly noticeable on the frontal surface, which is observable in the egocentric input views.}
    \label{fig:exp-dcaa}
\end{figure*}
%%%%
%
Tab.~\ref{tab:ego-tracking} compares the geometry reconstruction accuracy between the tracked mesh-based avatar and ground truth mesh reconstructed by NeuS2 \cite{wang2023neus2}.
Our method achieves the best reconstruction quality among feed-forward methods, approaching the performance of optimization-based methods EgoAvatar~\cite{chen2024egoavatar} while being significantly more efficient.
In contrast, solely motion-driven animatable avatars are less capable of obtaining precise surface recovery, especially for surface areas that are visible in the egocentric views.
As shown in Fig.~\ref{fig:exp-dcaa}, the most prominent improvements of our proposed depth-conditioned animatable avatar can be observed in the frontal and upper body surface, which are most likely observed in the egocentric views.
%
%%%%%%%%%%%%%
%
\subsubsection{Ablation Study on Egocentric Motion Capture}
%
%%%%
\begin{table}[t!]
    \centering
    \caption{\textbf{Quantitative Comparison and Ablation on Egocentric Pose Estimation and Inverse Kinematics.} Distances are in centimeter scale.}
    \label{tab:pose-ik}
    \begin{tabular}{|c|c|c|c|c|}
        \hline
        \multirow{2}{*}{Method} & \multicolumn{2}{c|}{MP-JPE $\downarrow$}&\multicolumn{2}{c|}{P2SDist $\downarrow$}\\
        \cline{2-5}
        &\makecell{Full\\Body} & Hands & \makecell{Full\\Body} & Hands\\
        \hline
        Ours w/o data aug. & 12.20& 14.55& 2.60& 4.68 \\
        \hline
        Ours w/o hand IK & \underline{5.38}& \underline{6.43}& \underline{1.76}& \underline{2.00}\\
        \hline
        \textbf{Ours}& \textbf{4.11}& \textbf{4.71}& \textbf{1.75}& \textbf{1.84}\\
        \hline
    \end{tabular}
\end{table}
%%%%
%
%%%%
\begin{figure}[t]
    \centering
    \includegraphics[width=.49\textwidth]{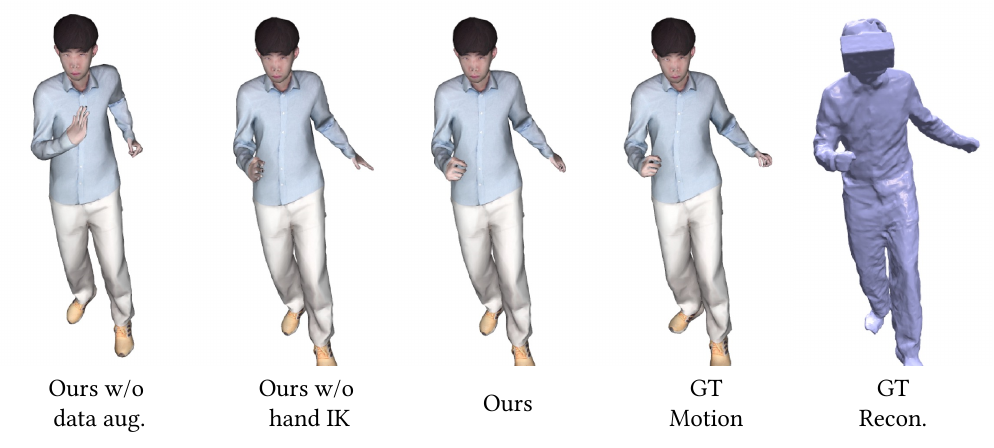}
    \caption{\textbf{Qualitative Ablation on Egocentric Motion Capture and Inverse Kinematics.} Note that our data augmentation strategy and hand inverse kinematics greatly improves the pose accuracy.}
    \label{fig:exp-ik}
\end{figure}
%%%%
%
Tab.~\ref{tab:pose-ik} proves the key finding to improve the robustness of egocentric perception under novel testing illuminations by data augmentation.
As demonstrated in Fig.~\ref{fig:exp-ik}, our IK module can recover plausible full-body with hand motions from egocentric images under novel testing environment.
In contrast, the ablated model trained only on uniform lighting fails in predicting stable 3D keypoints for inverse kinematics.
%
% \mhc{table still contains IK and PCA, but you dont discuss these results here?}
%
% \jcc{will remove}
%
%%%%%%%%%%%%%%%%%%%%%%%%%%%%%%%%%%%%%%%%%%%%%%%%%%%
%%%%%%%%%%%%%%%%%%%%%%%%%%%%%%%%%%%%%%%%%%%%%%%%%%%
%
\subsection{Relighting Experiments}\label{sec:exp-relightable}
%
%%%%%%%%%%%%%
%
\subsubsection{Quantitative and Qualitative Results}
\begin{figure*}[h]
    \centering
    \includegraphics[width=\textwidth]{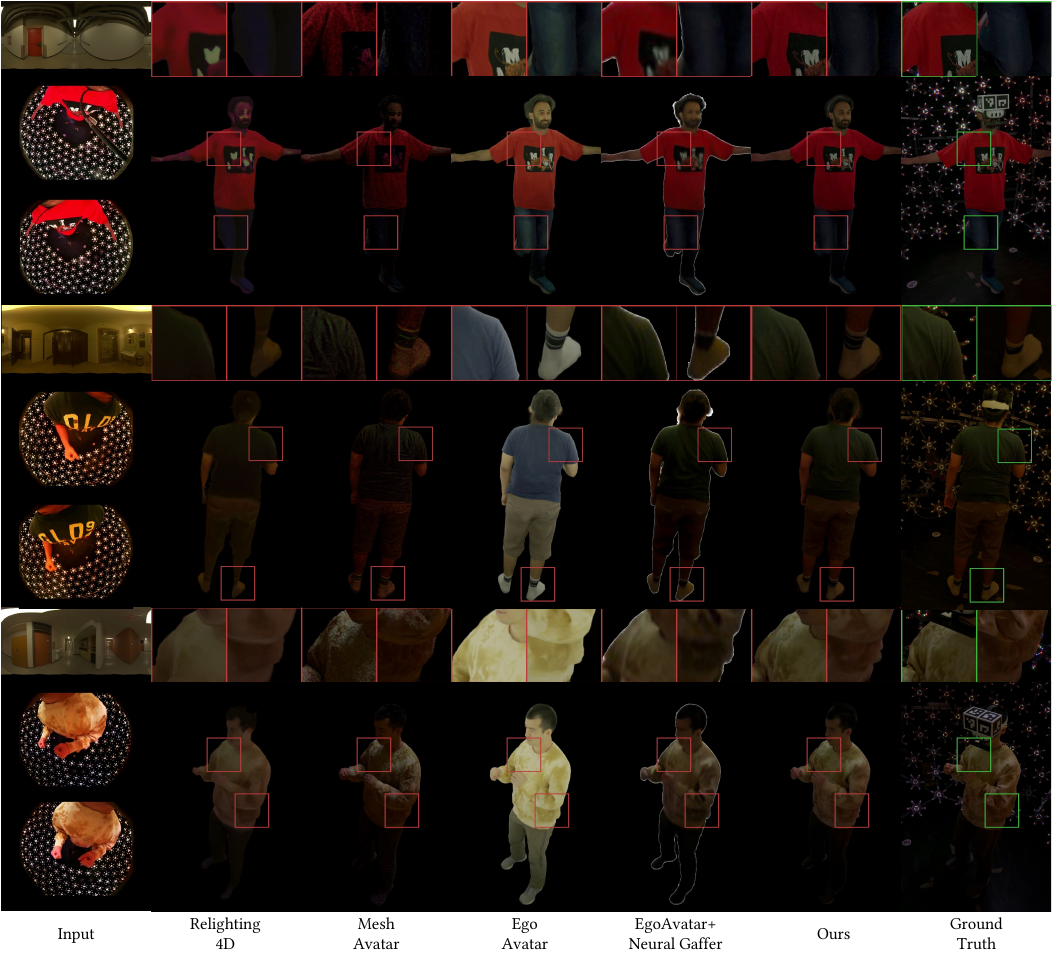}
    \caption{\textbf{Qualitative Comparison.} We show qualitative comparisons against recent physics-based relightable avatars~\cite{chen2022relighting4d, chen2024meshavatar} and image-based relighting approach~\cite{jin2024neural}. All experiments are conducted on novel testing sequences with novel view and novel illumination. To account for the disparity of head rendering with and without HMD, we exclude the head region in quantitative results.}
    \label{fig:exp-relighting-all}
\end{figure*}

In this section, we present the quantitative and qualitative comparison against the aforementioned state-of-the-art full-body relighting baselines.
As shown in Tab.~\ref{tab:relight-full}, our approach consistently outperforms prior works, achieving the top results across all subjects in PSNR, LPIPS, and FID, and yielding the highest SSIM scores in the vast majority of cases.
Concretely, the existing relightable avatar methods struggle in predicting a high-fidelity appearance while being able to match the overall color tone.
On the contrary, the non-relightable avatar~\cite{chen2024egoavatar} performs the worst in pixel-level metrics (\textit{i.e.} PSNR and SSIM) despite achieving photorealism.
Fig.~\ref{fig:exp-relighting-all} visualizes the novel view and novel light rendering of both methods in testing sequences.
In particular, Relighting4D~\cite{chen2022relighting4d} and Neural Gaffer~\cite{jin2024neural} have limited capacity in re-rendering high-resolution image while losing facial details.
MeshAvatar~\cite{chen2024meshavatar} generates sharp clothing textures and wrinkles with severe noise.
The color tone of the relit image also differs from ground truth (\textit{e.g.} the skin tone of the subject in the second row is over-reddish).
In contrast, our method can recover both photorealistic visual details and accurate shading color compared to ground truth capture.
%
%%%%
%
%%%%%%%%%%%%%%%%%%%%%%%%
%
\begin{table*}[!t]
    \centering
    \caption{\textbf{Quantitative Evaluation.} We highlight that we outperform competing methods in PSNR, LPIPS, and FID among all subjects, and achieve the highest SSIM scores in three of the four subjects.}
    % \vspace{-5pt}
    \label{tab:relight-full}
    \footnotesize 
    \setlength{\tabcolsep}{3pt} 
    \renewcommand{\arraystretch}{1.2} 
    \begin{tabular}{|c?c | c | c | c ?c |c |c|c?c|c|c|c?c|c|c|c|}
        \hline
        \multirow{2}{*}{\diagbox{Method}{Subject}} & \multicolumn{4}{c?}{\# 1} & \multicolumn{4}{c?}{\# 2} & \multicolumn{4}{c?}{\# 3} & \multicolumn{4}{c|}{\# 4}\\
        \cline{2-17}
         & PSNR & SSIM & LPIPS & FID &PSNR & SSIM & LPIPS & FID &PSNR & SSIM & LPIPS & FID &PSNR & SSIM & LPIPS & FID \\
         \hline
         Relighting4D~\shortcite{chen2022relighting4d} $\maltese$ & 32.39& \underline{92.00}& 10.02& 153.33& \underline{33.12} & \underline{82.77} & \underline{6.94} & 83.75 & \underline{34.86} & \underline{87.35} & \underline{5.90} & 96.68 & \underline{32.97} & \textbf{87.39} & \underline{7.89} & 86.03\\
         \hline
         MeshAvatar~\shortcite{chen2024meshavatar} $\maltese$ & \underline{33.45} & 89.00 & \underline{9.94} &	\underline{41.66} & 29.96 & 76.81 &	8.87 & 94.87 &32.02 & 73.46 & 7.38 & 107.15 & 32.01 & 82.38 & 8.06 & \underline{59.38}\\
         \hline
         EgoAvatar~\shortcite{chen2024egoavatar} & 20.93 &	82.48 &	16.49 &	62.64 & 26.85 &	75.20 &	8.35 & \underline{47.78} & 23.47 & 70.28 & 8.96 & \underline{80.91}& 20.41 & 72.26 & 12.18 & 107.70\\
         \hline
         EgoAvatar~\shortcite{chen2024egoavatar} + NeuralGaffer~\shortcite{jin2024neural} & 28.77 &	90.07 &	10.05 & 65.11 & 29.78 &	83.57 &	7.21 & 87.94 & 31.07 & 85.11 & 6.27 & 110.95 & 29.78 & 85.10 & 8.10 & 116.23\\
         \hline
         \textbf{Ours} & \textbf{34.81}& \textbf{92.46}& \textbf{8.56}& \textbf{31.90}& \textbf{36.02} & \textbf{89.33} & \textbf{5.90} & \textbf{23.20} & \textbf{35.47} & \textbf{88.63} & \textbf{5.23} &	\textbf{31.81} & \textbf{33.79} & \underline{87.07} & \textbf{6.89} & \textbf{42.93}\\
        \hline 
    \end{tabular}
\end{table*}
%
%%%%%%%%%%%%%%%%%%%%%%%%
%
%%%%
%
%%%%
%%%%
%
%%%%%%%%%%%%%
%
\subsubsection{Ablation Studies} \label{sec:exp-relight-ablation}
% 
%%%%
\begin{table}[t!]
\small
    \centering
    \caption{\textbf{Ablation Studies on Relightable Appearance Modeling.} $\dagger$: We leverage specular maps introduced in TotalRelighting~\cite{pandey2021total} as an alternative of cross-attention architecture to decode specular shading. SSIM and LPIPS scores are scaled by 100$\times$.}
    \label{tab:relight-ablation}
    \begin{tabular}{|c|c|c|c|c|}
        \hline
        Method & PSNR $\uparrow$ & SSIM $\uparrow$ & LPIPS $\downarrow$ & FID $\downarrow$ \\
        \hline
         w/o GeoLiftingNet $\mathcal{H}_\mathrm{Lift}$ & 34.35 & 	92.48 &	8.76 & 41.25\\
        \hline
         w/o DiffuseNet $\mathcal{H}_\mathrm{Diff}$ & 19.00& 71.34 &	13.76 & 207.42\\
        \hline
         w/o SpecularNet $\mathcal{H}_\mathrm{Spec}$ & 34.61& \textbf{92.76} &	8.79 & 76.25\\
        \hline
         w/o cross-attention $\dagger$ \shortcite{pandey2021total} & 34.53 &	92.30 &	8.67 &	\underline{31.93}\\
        \hline
         w/ DisneyBRDF \shortcite{burley2012physically} & 33.80 & 91.85 & 9.04 & 40.59\\
        \hline
         w/ num. of rays $r=2$ & 34.22 &	92.19 &	8.88 &	38.60\\
        \hline
         w/ num. of rays $r=8$ & \textbf{34.98} &	\underline{92.57} &	\underline{8.57} &	34.28\\
        \hline
         \textbf{Ours} (w/ num. of rays $r=32$)& \underline{34.81}& 92.46& \textbf{8.56}&\textbf{31.90}\\
        \hline
        
    \end{tabular}
\end{table}
%%%%
%
%%%%
\begin{figure*}[h]
    \centering
    \includegraphics[width=\textwidth]{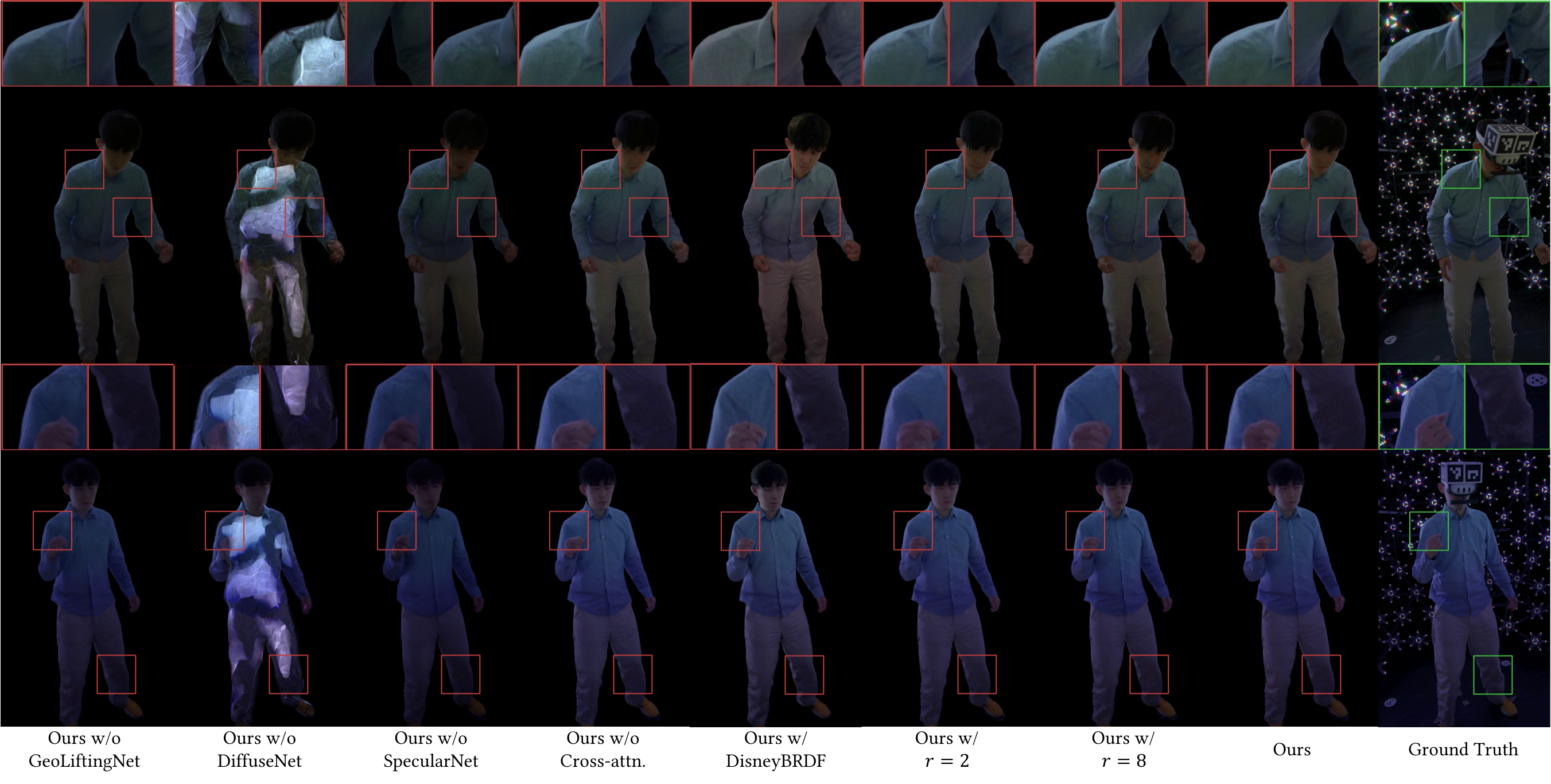}
    \vspace{-10pt}
    \caption{\textbf{Qualitative ablation study.} We demonstrate the rendering of testing sequence under novel view and novel lighting conditions. Our model shows richer and smoother wrinkle details, with more accurate highlights, particularly in the zoom-in visualizations.}
    \label{fig:exp-relighting-ablation}
\end{figure*}
%%%%
% 
Next, we ablate our proposed modules, i.e.,  GeoLiftingNet, DiffuseNet, and SpecularNet, and key design choices, i.e., the cross-attention architecture and the number of sampled rays, of our relightable avatar (Sec.~\ref{sec:relighting}).
Instead of our proposed cross-attention mechanism, we encode the specular signal following TotalRelighting~\cite{pandey2021total} as a baseline.
As confirmed in Tab.~\ref{tab:relight-ablation}, our model with 32 sampled rays performs the best in terms of LPIPS and FID scores, while showing competitive pixel accuracy in terms of PSNR.
We additionally compare against an explicit DisneyBRDF~\cite{burley2012physically} baseline that predicts per-Gaussian albedo, roughness, and metallic parameters. 
As confirmed in Tab.~\ref{tab:relight-ablation}, our model with 32 sampled rays performs the best in terms of LPIPS and FID scores, while showing competitive pixel accuracy in terms of PSNR.
This is also qualitatively confirmed in Fig.~\ref{fig:exp-relighting-ablation}, where the model without GeoLiftingNet misses wrinkle details, the model without DiffuseNet fails to converge with only 32 sampled rays, and the model without SpecularNet underfits specular highlights in the shoulder.
The model with the specular map is close to the performance of our full approach.
However, it still misses some highlights, e.g., the lower arm in the first row.
The baseline with DisneyBRDF introduces noticeable color shift relative to the ground truth, alongside unrealistic, over-metallic highlights in hair regions.
Regarding our proposed method, we observe smoother highlights in the zoom-in visualization of the shoulder part when the number of sampled rays increases.
%
%%%%%%%%%%%%%
%
\subsubsection{Robustness under Challenging Illuminations} \label{sec:exp-relight-challenging}
%
%%%%
\begin{figure*}[h]
    \centering
    \includegraphics[width=0.85\textwidth]{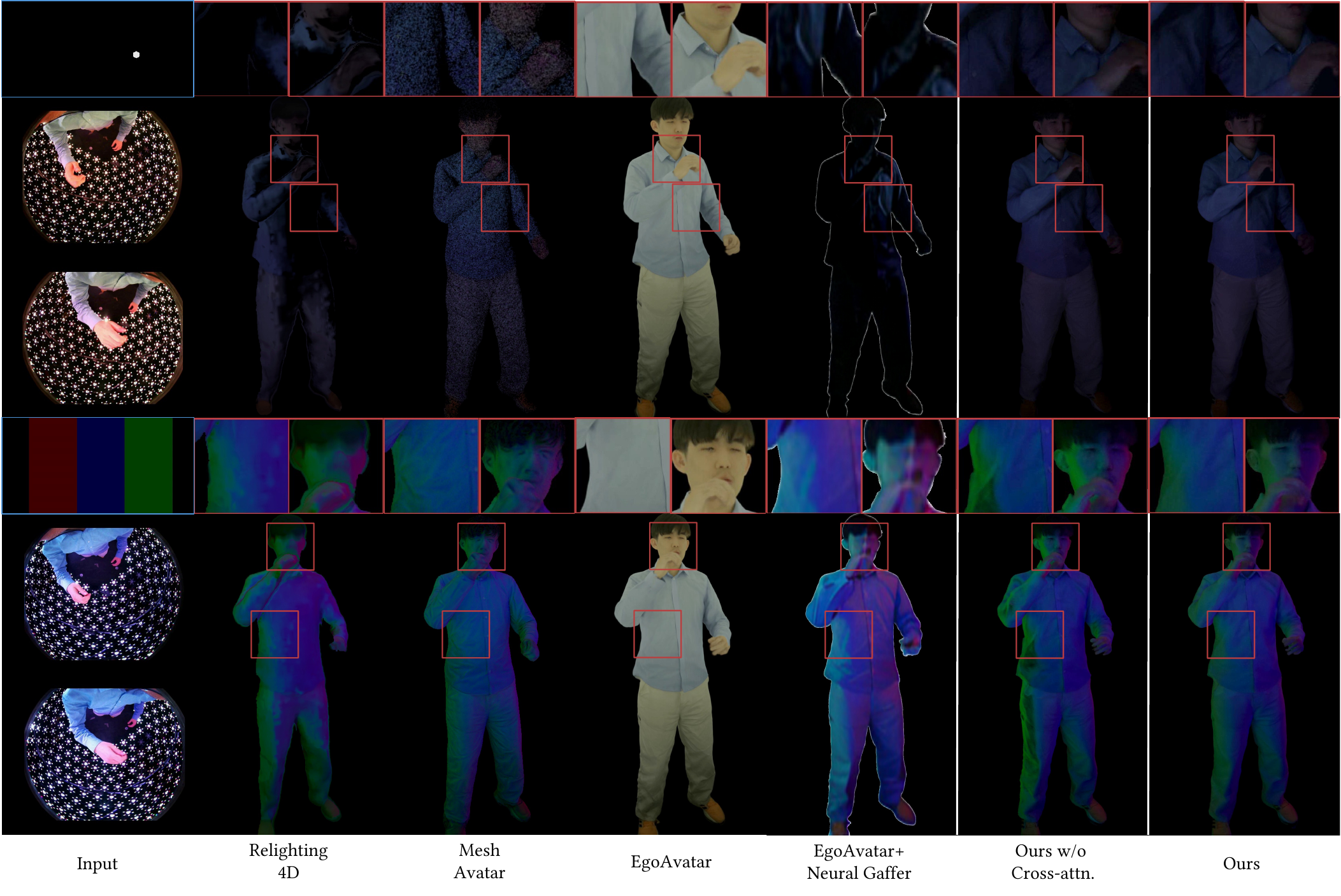}
    \vspace{-10pt}
    \caption{\textbf{Qualitative Comparison on Challenging Illuminations.} We compare the visual performance of our method against baselines in two type of unseen lighting conditions: OLAT and a rotating high-contrast environment map. Notably, our method achieves the most realistic rendering results.}
    \label{fig:exp-relighting-challenge}
\end{figure*}
%%%%
% 
Despite eight testing illuminations, we conduct further comparisons on extreme and out-of-distribution target illuminations.
Due to the difficulty of egocentric perception in extreme lighting conditions, we do not capture and quantitatively evaluate additional sequences.
Instead, we demonstrate the qualitative comparison between our full model, competing methods, and a baseline without cross-attention.
In Fig.~\ref{fig:exp-relighting-challenge}, we can see that our model generates most realistic renderings with sharp self-shadowing and highlights, whereas inverse rendering-based models~\cite{chen2022relighting4d,chen2024meshavatar} generate noisy images.
The diffusion-based model~\cite{jin2024neural} misses the correct color tone in out-of-distribution lighting patterns.
Our baseline with specular map as conditioning synthesizes reasonable OLAT lighting, while overfitting the blank environment map (in the second row) as shadow artifacts.
%
%%%%%%%%%%%%%%%%%%%%%%%%%%%%%%%%%%%%%%%%%%%%%%%%%%%
%%%%%%%%%%%%%%%%%%%%%%%%%%%%%%%%%%%%%%%%%%%%%%%%%%%
%
\subsection{Evaluation on Egocentric Illumination Estimation} \label{sec:exp-env}
%
%%%%
\begin{figure*}[h]
    \centering
    \includegraphics[width=\textwidth]{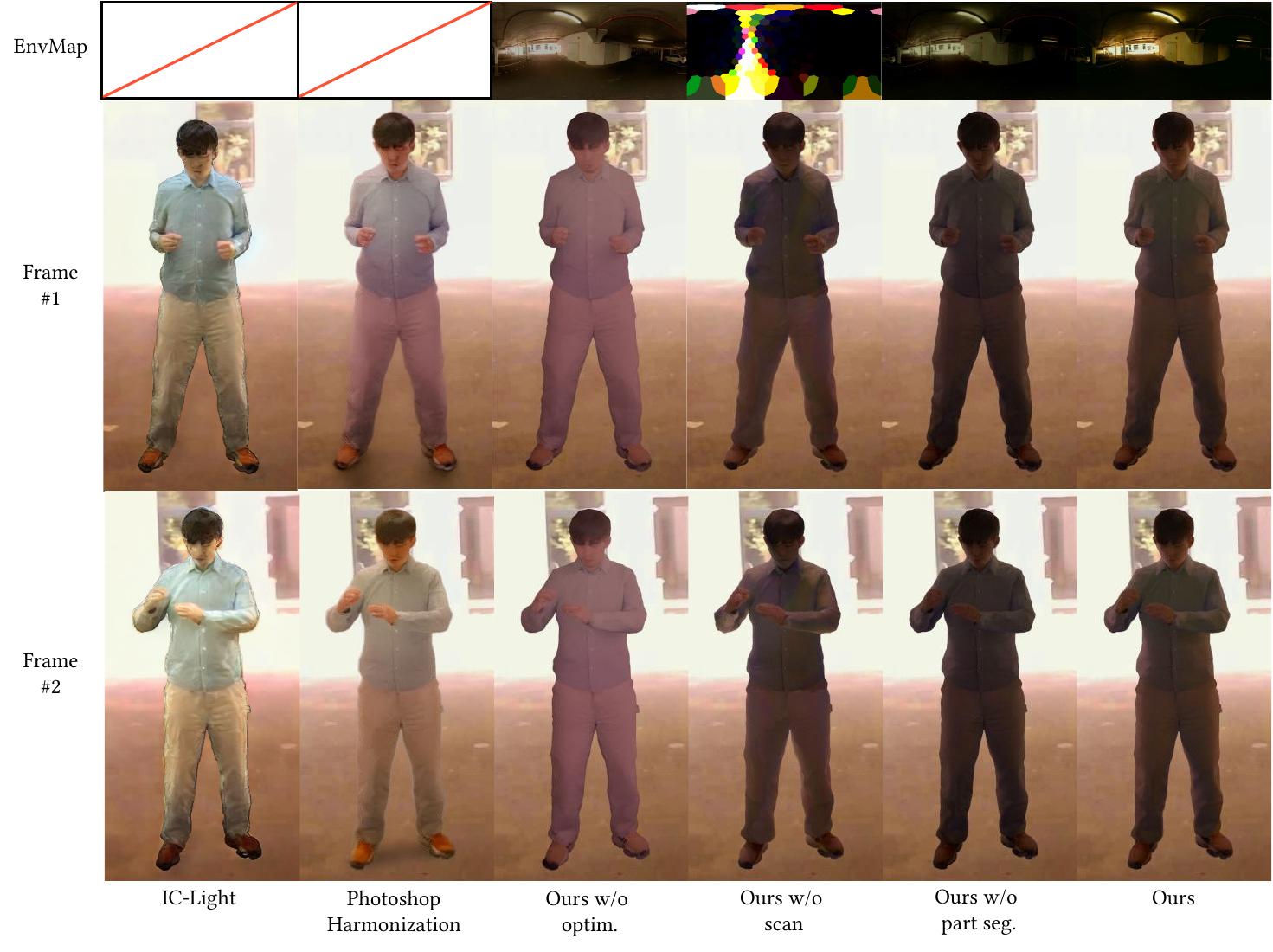}
    \caption{\textbf{Comparisons and Ablation Study on In-the-wild Environment Map Capture.} We visualize the estimated environment map and relighting renderings in an unseen testing scenario. Our model shows better plausibility compared with image-based harmonization approaches given 360 degree observation of the scene, and less artifacts compared with our ablated baselines.}
    \label{fig:exp-envmap}
\end{figure*}
%%%%
%
In this section, we compare and ablate our in-the-wild environment map acquisition (Sec.~\ref{sec:env-map-capture}).
Specifically, we compare with zero-shot generative harmonization techniques, i.e., IC-Light~\cite{zhang2025scaling} and Adobe Photoshop harmonization\footnote{\url{https://helpx.adobe.com/photoshop/using/blend-subjects-with-harmonize.html}}.
Moreover, we validate the necessity of performing our egocentric illumination capture pipeline, including the steps for inverse rendering, 360 degree environment scanning, and using part-segmentation to balance the inverse rendering weight.
In Fig.~\ref{fig:exp-envmap}, Photoshop Harmonization, as an image-based method, generates inconsistent color tone along the sequence, whereas IC-Light destructs the facial details.
Further, harmonization methods try to blend human into the background, which do not reflect the actual shading, which depends on the illumination in opposite direction.
On the other hand, directly using LDR panorama scan also leads to an incorrect color tone (over-reddish in the trousers) while removing the panorama scan as a prior results in the overfitting of egocentric images (over-greenish in the face).
Finally, we show that part segmentation prevents the overfitting of upper body part and relights human with more natural brightness.
%
%%%%%%%%%%%%%%%%%%%%%%%%%%%%%%%%%%%%%%%%%%%%%%%%%%%
%%%%%%%%%%%%%%%%%%%%%%%%%%%%%%%%%%%%%%%%%%%%%%%%%%%
%
% \input{sections/limitation}
%
%%%%%%%%%%%%%%%%%%%%%%%%%%%%%%%%%%%%%%%%%%%%%%%%%%%
%%%%%%%%%%%%%%%%%%%%%%%%%%%%%%%%%%%%%%%%%%%%%%%%%%%
%
\section{Limitation and Future Work} \label{sec:limitation_conclusion}
While our method can simultaneously achieve egocentric performance capture from any illumination and avatar relighting to any target illumination, we identify and discuss limitations in the following that open up future work in this direction.
Firstly, our model has not yet been optimized for a real-time use case (see Tab.~\ref{tab:runtime-breakdown} for a detailed runtime breakdown).
The major bottleneck so far is the iterative IK-solver, which could be replaced with more efficient second-order solvers.
\begin{table}[t!]
    \centering
    \caption{\textbf{Runtime Breakdown of Our Pipeline.} Per-frame runtime is reported in milliseconds. All evaluation are conducted in the test sequence of subject \#1 using a single Nvidia RTX 4090Ti GPU. IK is solved once as an offline sequence-level optimization over the entire clip instead of on-the-fly per-frame inference, and we report both its equivalent per-frame cost and total runtime.}
    \label{tab:runtime-breakdown}
    \begin{tabular}{|c|c|}
        \hline
        Component & Per-frame Runtime \\
        \hline
        EgoPose Estimation & 17.62 ms \\
        \hline
        Inverse Kinematics & 67.25 ms (343.04 s total) \\
        \hline
        EgoDepth Estimation & 87.71 ms \\
        \hline
        Depth-conditioned Animation & 21.50 ms \\
        \hline
        Relightable Appearance & 100.19 ms \\
        \hline
    \end{tabular}
\end{table}
Secondly, due to hardware constraints, head mounted cameras now operate with fixed exposure, which may not work under very bright/dark illumination conditions.
The LDR sensor also poses challenges for color synchronization across different group of egocentric cameras.
In the future, we seek to upgrade the egocentric camera or to leverage the multi-camera with bracketing technique to obtain HDR images.
Further, outdoor relighting under strong sunlight is currently limited by practical data capture pipeline. To align with the lightstage training setup, we clip the testing environment-map intensity (e.g. sunlight in Fig.~\ref{fig:exp-qual} reaching $\sim 1.6\times10^5$) to 1.0, which reduces contrast and yields softened shadows.
Moreover, we still observe limited facial expressiveness and occasional foot-ground contact artifacts, which can be improved by integrating additional face sensing modules \cite{bai2024universal} and physics constraints \cite{luo2024real}.
Lastly, our method greatly relies on high-quality person-specific capture from lightstage, which is not approachable for public use.
Future work can explore a universal relightable full-body avatar leveraging the power of generative priors.
%
%%%%%%%%%%%%%%%%%%%%%%%%%%%%%%%%%%%%%%%%%%%%%%%%%%%
%%%%%%%%%%%%%%%%%%%%%%%%%%%%%%%%%%%%%%%%%%%%%%%%%%%
%
\section{Conclusion}
In conclusion, this paper presented a novel approach \methodname, which integrates state-of-the-art egocentric performance capture, photorealistic relightable avatar, and test-time HDR environment map capture in a unified framework. 
\methodname{} sequentially recovers keypoints, depth maps, meshes, and Gaussian-based avatars with separate specular/diffuse shading from egocentric down-facing camera, while also capturing HDR environment maps from egocentric front-facing camera.
We demonstrate the robustness of our approach in capturing and relighting the avatar in novel outdoor scenarios, enabling a seamless virtual human insertion into the physical world.
We believe that this paper has the potential to shape the next-generation video conferencing enabling a ubiquitous and immersive mixed reality experience.
%
%%%%%%%%%%%%%%%%%%%%%%%%%%%%%%%%%%%%%%%%%%%%%%%%%%%
%%%%%%%%%%%%%%%%%%%%%%%%%%%%%%%%%%%%%%%%%%%%%%%%%%%
%
%
%%%%%%%%%%%%%%%%%%%%%%%%
% NOWLEGDEMENT
%%%%%%%%%%%%%%%%%%%%%%%%
% \section{Acknowledgements}
\begin{acks}
This work was supported by the Saarbr\"ucken Research Center for Visual Computing, Interaction, and AI.
The authors thank the anonymous reviewers for their thoughtful feedback and suggestions. 
The authors also thank Guoxing Sun, Pramod Rao, and Andrea Boscolo Camiletto for their assistance with hardware setup, data capture, and help on egocentric pose estimation.
\end{acks}

%%%%%%%%%%%%%%%%%%%%%%%%
% BIBLIOGRAPHY
%%%%%%%%%%%%%%%%%%%%%%%%
%
% \clearpage
\bibliographystyle{ACM-Reference-Format}
\bibliography{bibliography}


\begin{thebibliography}{96}

%%% the Web address.

\ifx \showCODEN    \undefined \def \showCODEN     #1{\unskip}     \fi
\ifx \showISBNx    \undefined \def \showISBNx     #1{\unskip}     \fi
\ifx \showISBNxiii \undefined \def \showISBNxiii  #1{\unskip}     \fi
\ifx \showISSN     \undefined \def \showISSN      #1{\unskip}     \fi
\ifx \showLCCN     \undefined \def \showLCCN      #1{\unskip}     \fi
\ifx \shownote     \undefined \def \shownote      #1{#1}          \fi
\ifx \showarticletitle \undefined \def \showarticletitle #1{#1}   \fi
\ifx \showURL      \undefined \def \showURL       {\relax}        \fi
% invisible to TeX
\providecommand\bibfield[2]{#2}
\providecommand\bibinfo[2]{#2}
\providecommand\natexlab[1]{#1}
\providecommand\showeprint[2][]{arXiv:#2}

\bibitem[Akada et~al\mbox{.}(2022)]%
        {akada2022unrealego}
\bibfield{author}{\bibinfo{person}{Hiroyasu Akada}, \bibinfo{person}{Jian Wang}, \bibinfo{person}{Soshi Shimada}, \bibinfo{person}{Masaki Takahashi}, \bibinfo{person}{Christian Theobalt}, {and} \bibinfo{person}{Vladislav Golyanik}.} \bibinfo{year}{2022}\natexlab{}.
\newblock \showarticletitle{Unrealego: A new dataset for robust egocentric 3d human motion capture}. In \bibinfo{booktitle}{\emph{European Conference on Computer Vision}}. Springer, \bibinfo{pages}{1--17}.
\newblock


\bibitem[Bagautdinov et~al\mbox{.}(2021)]%
        {bagautdinov2021driving}
\bibfield{author}{\bibinfo{person}{Timur Bagautdinov}, \bibinfo{person}{Chenglei Wu}, \bibinfo{person}{Tomas Simon}, \bibinfo{person}{Fabian Prada}, \bibinfo{person}{Takaaki Shiratori}, \bibinfo{person}{Shih-En Wei}, \bibinfo{person}{Weipeng Xu}, \bibinfo{person}{Yaser Sheikh}, {and} \bibinfo{person}{Jason Saragih}.} \bibinfo{year}{2021}\natexlab{}.
\newblock \showarticletitle{Driving-signal aware full-body avatars}.
\newblock \bibinfo{journal}{\emph{ACM Transactions on Graphics (TOG)}} \bibinfo{volume}{40}, \bibinfo{number}{4} (\bibinfo{year}{2021}), \bibinfo{pages}{1--17}.
\newblock


\bibitem[Bai et~al\mbox{.}(2024)]%
        {bai2024universal}
\bibfield{author}{\bibinfo{person}{Shaojie Bai}, \bibinfo{person}{Te-Li Wang}, \bibinfo{person}{Chenghui Li}, \bibinfo{person}{Akshay Venkatesh}, \bibinfo{person}{Tomas Simon}, \bibinfo{person}{Chen Cao}, \bibinfo{person}{Gabriel Schwartz}, \bibinfo{person}{Jason Saragih}, \bibinfo{person}{Yaser Sheikh}, {and} \bibinfo{person}{Shih-En Wei}.} \bibinfo{year}{2024}\natexlab{}.
\newblock \showarticletitle{Universal Facial Encoding of Codec Avatars from VR Headsets}.
\newblock \bibinfo{journal}{\emph{ACM Transactions on Graphics (TOG)}} \bibinfo{volume}{43}, \bibinfo{number}{4} (\bibinfo{year}{2024}), \bibinfo{pages}{1--22}.
\newblock


\bibitem[Bi et~al\mbox{.}(2021)]%
        {bi2021deep}
\bibfield{author}{\bibinfo{person}{Sai Bi}, \bibinfo{person}{Stephen Lombardi}, \bibinfo{person}{Shunsuke Saito}, \bibinfo{person}{Tomas Simon}, \bibinfo{person}{Shih-En Wei}, \bibinfo{person}{Kevyn Mcphail}, \bibinfo{person}{Ravi Ramamoorthi}, \bibinfo{person}{Yaser Sheikh}, {and} \bibinfo{person}{Jason Saragih}.} \bibinfo{year}{2021}\natexlab{}.
\newblock \showarticletitle{Deep relightable appearance models for animatable faces}.
\newblock \bibinfo{journal}{\emph{ACM Transactions on Graphics (ToG)}} \bibinfo{volume}{40}, \bibinfo{number}{4} (\bibinfo{year}{2021}), \bibinfo{pages}{1--15}.
\newblock


\bibitem[Blinn(1977)]%
        {blinn1977models}
\bibfield{author}{\bibinfo{person}{James~F Blinn}.} \bibinfo{year}{1977}\natexlab{}.
\newblock \showarticletitle{Models of light reflection for computer synthesized pictures}. In \bibinfo{booktitle}{\emph{Proceedings of the 4th annual conference on Computer graphics and interactive techniques}}. \bibinfo{pages}{192--198}.
\newblock


\bibitem[Burley and Studios(2012)]%
        {burley2012physically}
\bibfield{author}{\bibinfo{person}{Brent Burley} {and} \bibinfo{person}{Walt Disney~Animation Studios}.} \bibinfo{year}{2012}\natexlab{}.
\newblock \showarticletitle{Physically-based shading at disney}. In \bibinfo{booktitle}{\emph{Acm Siggraph}}, Vol.~\bibinfo{volume}{2012}. vol. 2012, \bibinfo{pages}{1--7}.
\newblock


\bibitem[Camiletto et~al\mbox{.}(2025)]%
        {camiletto2025frame}
\bibfield{author}{\bibinfo{person}{Andrea~Boscolo Camiletto}, \bibinfo{person}{Jian Wang}, \bibinfo{person}{Eduardo Alvarado}, \bibinfo{person}{Rishabh Dabral}, \bibinfo{person}{Thabo Beeler}, \bibinfo{person}{Marc Habermann}, {and} \bibinfo{person}{Christian Theobalt}.} \bibinfo{year}{2025}\natexlab{}.
\newblock \showarticletitle{FRAME: Floor-aligned Representation for Avatar Motion from Egocentric Video}. In \bibinfo{booktitle}{\emph{Proceedings of the Computer Vision and Pattern Recognition Conference}}. \bibinfo{pages}{17497--17507}.
\newblock


\bibitem[Chen et~al\mbox{.}(2024a)]%
        {chen2024sequential}
\bibfield{author}{\bibinfo{person}{Jianchun Chen}, \bibinfo{person}{Jayakorn Vongkulbhisal}, {and} \bibinfo{person}{Fernando De~la Torre~Frade}.} \bibinfo{year}{2024}\natexlab{a}.
\newblock \showarticletitle{A Sequential Learning-based Approach for Monocular Human Performance Capture}. In \bibinfo{booktitle}{\emph{Proceedings of the IEEE/CVF Winter Conference on Applications of Computer Vision}}. \bibinfo{pages}{3514--3523}.
\newblock


\bibitem[Chen et~al\mbox{.}(2024b)]%
        {chen2024egoavatar}
\bibfield{author}{\bibinfo{person}{Jianchun Chen}, \bibinfo{person}{Jian Wang}, \bibinfo{person}{Yinda Zhang}, \bibinfo{person}{Rohit Pandey}, \bibinfo{person}{Thabo Beeler}, \bibinfo{person}{Marc Habermann}, {and} \bibinfo{person}{Christian Theobalt}.} \bibinfo{year}{2024}\natexlab{b}.
\newblock \showarticletitle{EgoAvatar: Egocentric View-Driven and Photorealistic Full-body Avatars}. In \bibinfo{booktitle}{\emph{SIGGRAPH Asia 2024 Conference Papers}}. \bibinfo{pages}{1--11}.
\newblock


\bibitem[Chen et~al\mbox{.}(2024c)]%
        {chen2024meshavatar}
\bibfield{author}{\bibinfo{person}{Yushuo Chen}, \bibinfo{person}{Zerong Zheng}, \bibinfo{person}{Zhe Li}, \bibinfo{person}{Chao Xu}, {and} \bibinfo{person}{Yebin Liu}.} \bibinfo{year}{2024}\natexlab{c}.
\newblock \showarticletitle{Meshavatar: Learning high-quality triangular human avatars from multi-view videos}. In \bibinfo{booktitle}{\emph{European Conference on Computer Vision}}. Springer, \bibinfo{pages}{250--269}.
\newblock


\bibitem[Chen and Liu(2022)]%
        {chen2022relighting4d}
\bibfield{author}{\bibinfo{person}{Zhaoxi Chen} {and} \bibinfo{person}{Ziwei Liu}.} \bibinfo{year}{2022}\natexlab{}.
\newblock \showarticletitle{Relighting4d: Neural relightable human from videos}. In \bibinfo{booktitle}{\emph{European Conference on Computer Vision}}. Springer, \bibinfo{pages}{606--623}.
\newblock


\bibitem[Cook and Torrance(1982)]%
        {cook1982reflectance}
\bibfield{author}{\bibinfo{person}{Robert~L Cook} {and} \bibinfo{person}{Kenneth~E. Torrance}.} \bibinfo{year}{1982}\natexlab{}.
\newblock \showarticletitle{A reflectance model for computer graphics}.
\newblock \bibinfo{journal}{\emph{ACM Transactions on Graphics (ToG)}} \bibinfo{volume}{1}, \bibinfo{number}{1} (\bibinfo{year}{1982}), \bibinfo{pages}{7--24}.
\newblock


\bibitem[Debevec et~al\mbox{.}(2002)]%
        {debevec2002lighting}
\bibfield{author}{\bibinfo{person}{Paul Debevec}, \bibinfo{person}{Andreas Wenger}, \bibinfo{person}{Chris Tchou}, \bibinfo{person}{Andrew Gardner}, \bibinfo{person}{Jamie Waese}, {and} \bibinfo{person}{Tim Hawkins}.} \bibinfo{year}{2002}\natexlab{}.
\newblock \showarticletitle{A lighting reproduction approach to live-action compositing}.
\newblock \bibinfo{journal}{\emph{ACM Transactions on Graphics (TOG)}} \bibinfo{volume}{21}, \bibinfo{number}{3} (\bibinfo{year}{2002}), \bibinfo{pages}{547--556}.
\newblock


\bibitem[Elgharib et~al\mbox{.}(2020)]%
        {elgharib2020egocentric}
\bibfield{author}{\bibinfo{person}{Mohamed Elgharib}, \bibinfo{person}{Mohit Mendiratta}, \bibinfo{person}{Justus Thies}, \bibinfo{person}{Matthias Nie{\ss}ner}, \bibinfo{person}{Hans-Peter Seidel}, \bibinfo{person}{Ayush Tewari}, \bibinfo{person}{Vladislav Golyanik}, {and} \bibinfo{person}{Christian Theobalt}.} \bibinfo{year}{2020}\natexlab{}.
\newblock \showarticletitle{Egocentric videoconferencing}.
\newblock \bibinfo{journal}{\emph{ACM Transactions on Graphics (TOG)}} \bibinfo{volume}{39}, \bibinfo{number}{6} (\bibinfo{year}{2020}), \bibinfo{pages}{1--16}.
\newblock


\bibitem[Finlayson et~al\mbox{.}(2015)]%
        {finlayson2015color}
\bibfield{author}{\bibinfo{person}{Graham~D Finlayson}, \bibinfo{person}{Michal Mackiewicz}, {and} \bibinfo{person}{Anya Hurlbert}.} \bibinfo{year}{2015}\natexlab{}.
\newblock \showarticletitle{Color correction using root-polynomial regression}.
\newblock \bibinfo{journal}{\emph{IEEE Transactions on Image Processing}} \bibinfo{volume}{24}, \bibinfo{number}{5} (\bibinfo{year}{2015}), \bibinfo{pages}{1460--1470}.
\newblock


\bibitem[Flynn et~al\mbox{.}(2024)]%
        {flynn2024quark}
\bibfield{author}{\bibinfo{person}{John Flynn}, \bibinfo{person}{Michael Broxton}, \bibinfo{person}{Lukas Murmann}, \bibinfo{person}{Lucy Chai}, \bibinfo{person}{Matthew DuVall}, \bibinfo{person}{Cl{\'e}ment Godard}, \bibinfo{person}{Kathryn Heal}, \bibinfo{person}{Srinivas Kaza}, \bibinfo{person}{Stephen Lombardi}, \bibinfo{person}{Xuan Luo}, {et~al\mbox{.}}} \bibinfo{year}{2024}\natexlab{}.
\newblock \showarticletitle{Quark: Real-time, High-resolution, and General Neural View Synthesis}.
\newblock \bibinfo{journal}{\emph{ACM Transactions on Graphics (TOG)}} \bibinfo{volume}{43}, \bibinfo{number}{6} (\bibinfo{year}{2024}), \bibinfo{pages}{1--20}.
\newblock


\bibitem[Gardner et~al\mbox{.}(2017)]%
        {gardner2017learning}
\bibfield{author}{\bibinfo{person}{Marc-Andr{\'e} Gardner}, \bibinfo{person}{Kalyan Sunkavalli}, \bibinfo{person}{Ersin Yumer}, \bibinfo{person}{Xiaohui Shen}, \bibinfo{person}{Emiliano Gambaretto}, \bibinfo{person}{Christian Gagn{\'e}}, {and} \bibinfo{person}{Jean-Fran{\c{c}}ois Lalonde}.} \bibinfo{year}{2017}\natexlab{}.
\newblock \showarticletitle{Learning to predict indoor illumination from a single image}.
\newblock \bibinfo{journal}{\emph{ACM Transactions on Graphics (TOG)}} \bibinfo{volume}{36}, \bibinfo{number}{6} (\bibinfo{year}{2017}), \bibinfo{pages}{1--14}.
\newblock


\bibitem[Google(2025)]%
        {gemni2025gemini}
\bibfield{author}{\bibinfo{person}{Gemini~Team Google}.} \bibinfo{year}{2025}\natexlab{}.
\newblock \bibinfo{title}{Gemini Image}.
\newblock
\urldef\tempurl%
\url{https://deepmind.google/models/gemini/image/}
\showURL{%
\tempurl}


\bibitem[Grauman et~al\mbox{.}(2022)]%
        {grauman2022ego4d}
\bibfield{author}{\bibinfo{person}{Kristen Grauman}, \bibinfo{person}{Andrew Westbury}, \bibinfo{person}{Eugene Byrne}, \bibinfo{person}{Zachary Chavis}, \bibinfo{person}{Antonino Furnari}, \bibinfo{person}{Rohit Girdhar}, \bibinfo{person}{Jackson Hamburger}, \bibinfo{person}{Hao Jiang}, \bibinfo{person}{Miao Liu}, \bibinfo{person}{Xingyu Liu}, {et~al\mbox{.}}} \bibinfo{year}{2022}\natexlab{}.
\newblock \showarticletitle{Ego4d: Around the world in 3,000 hours of egocentric video}. In \bibinfo{booktitle}{\emph{Proceedings of the IEEE/CVF conference on computer vision and pattern recognition}}. \bibinfo{pages}{18995--19012}.
\newblock


\bibitem[Guo et~al\mbox{.}(2025)]%
        {guo2025pgc}
\bibfield{author}{\bibinfo{person}{Michelle Guo}, \bibinfo{person}{Matt Jen-Yuan Chiang}, \bibinfo{person}{Igor Santesteban}, \bibinfo{person}{Nikolaos Sarafianos}, \bibinfo{person}{Hsiao-yu Chen}, \bibinfo{person}{Oshri Halimi}, \bibinfo{person}{Alja{\v{z}} Bo{\v{z}}i{\v{c}}}, \bibinfo{person}{Shunsuke Saito}, \bibinfo{person}{Jiajun Wu}, \bibinfo{person}{C~Karen Liu}, {et~al\mbox{.}}} \bibinfo{year}{2025}\natexlab{}.
\newblock \showarticletitle{Pgc: Physics-based gaussian cloth from a single pose}. In \bibinfo{booktitle}{\emph{Proceedings of the Computer Vision and Pattern Recognition Conference}}. \bibinfo{pages}{21215--21225}.
\newblock


\bibitem[Habermann et~al\mbox{.}(2021)]%
        {habermann2021real}
\bibfield{author}{\bibinfo{person}{Marc Habermann}, \bibinfo{person}{Lingjie Liu}, \bibinfo{person}{Weipeng Xu}, \bibinfo{person}{Michael Zollhoefer}, \bibinfo{person}{Gerard Pons-Moll}, {and} \bibinfo{person}{Christian Theobalt}.} \bibinfo{year}{2021}\natexlab{}.
\newblock \showarticletitle{Real-time deep dynamic characters}.
\newblock \bibinfo{journal}{\emph{ACM Transactions on Graphics (ToG)}} \bibinfo{volume}{40}, \bibinfo{number}{4} (\bibinfo{year}{2021}), \bibinfo{pages}{1--16}.
\newblock


\bibitem[He et~al\mbox{.}(2024)]%
        {he2024diffrelight}
\bibfield{author}{\bibinfo{person}{Mingming He}, \bibinfo{person}{Pascal Clausen}, \bibinfo{person}{Ahmet~Levent Ta{\c{s}}el}, \bibinfo{person}{Li Ma}, \bibinfo{person}{Oliver Pilarski}, \bibinfo{person}{Wenqi Xian}, \bibinfo{person}{Laszlo Rikker}, \bibinfo{person}{Xueming Yu}, \bibinfo{person}{Ryan Burgert}, \bibinfo{person}{Ning Yu}, {et~al\mbox{.}}} \bibinfo{year}{2024}\natexlab{}.
\newblock \showarticletitle{DifFRelight: Diffusion-Based Facial Performance Relighting}. In \bibinfo{booktitle}{\emph{SIGGRAPH Asia 2024 Conference Papers}}. \bibinfo{pages}{1--12}.
\newblock


\bibitem[Heusel et~al\mbox{.}(2017)]%
        {heusel2017gans}
\bibfield{author}{\bibinfo{person}{Martin Heusel}, \bibinfo{person}{Hubert Ramsauer}, \bibinfo{person}{Thomas Unterthiner}, \bibinfo{person}{Bernhard Nessler}, {and} \bibinfo{person}{Sepp Hochreiter}.} \bibinfo{year}{2017}\natexlab{}.
\newblock \showarticletitle{Gans trained by a two time-scale update rule converge to a local nash equilibrium}.
\newblock \bibinfo{journal}{\emph{Advances in neural information processing systems}}  \bibinfo{volume}{30} (\bibinfo{year}{2017}).
\newblock


\bibitem[Hoppe et~al\mbox{.}(1992)]%
        {hoppe1992surface}
\bibfield{author}{\bibinfo{person}{Hugues Hoppe}, \bibinfo{person}{Tony DeRose}, \bibinfo{person}{Tom Duchamp}, \bibinfo{person}{John McDonald}, {and} \bibinfo{person}{Werner Stuetzle}.} \bibinfo{year}{1992}\natexlab{}.
\newblock \showarticletitle{Surface reconstruction from unorganized points}. In \bibinfo{booktitle}{\emph{Proceedings of the 19th annual conference on computer graphics and interactive techniques}}. \bibinfo{pages}{71--78}.
\newblock


\bibitem[Hu et~al\mbox{.}(2021)]%
        {hu2021egorenderer}
\bibfield{author}{\bibinfo{person}{Tao Hu}, \bibinfo{person}{Kripasindhu Sarkar}, \bibinfo{person}{Lingjie Liu}, \bibinfo{person}{Matthias Zwicker}, {and} \bibinfo{person}{Christian Theobalt}.} \bibinfo{year}{2021}\natexlab{}.
\newblock \showarticletitle{Egorenderer: Rendering human avatars from egocentric camera images}. In \bibinfo{booktitle}{\emph{Proceedings of the IEEE/CVF International Conference on Computer Vision}}. \bibinfo{pages}{14528--14538}.
\newblock


\bibitem[Iwase et~al\mbox{.}(2023)]%
        {iwase2023relightablehands}
\bibfield{author}{\bibinfo{person}{Shun Iwase}, \bibinfo{person}{Shunsuke Saito}, \bibinfo{person}{Tomas Simon}, \bibinfo{person}{Stephen Lombardi}, \bibinfo{person}{Timur Bagautdinov}, \bibinfo{person}{Rohan Joshi}, \bibinfo{person}{Fabian Prada}, \bibinfo{person}{Takaaki Shiratori}, \bibinfo{person}{Yaser Sheikh}, {and} \bibinfo{person}{Jason Saragih}.} \bibinfo{year}{2023}\natexlab{}.
\newblock \showarticletitle{Relightablehands: Efficient neural relighting of articulated hand models}. In \bibinfo{booktitle}{\emph{Proceedings of the IEEE/CVF Conference on Computer Vision and Pattern Recognition}}. \bibinfo{pages}{16663--16673}.
\newblock


\bibitem[Jin et~al\mbox{.}(2024)]%
        {jin2024neural}
\bibfield{author}{\bibinfo{person}{Haian Jin}, \bibinfo{person}{Yuan Li}, \bibinfo{person}{Fujun Luan}, \bibinfo{person}{Yuanbo Xiangli}, \bibinfo{person}{Sai Bi}, \bibinfo{person}{Kai Zhang}, \bibinfo{person}{Zexiang Xu}, \bibinfo{person}{Jin Sun}, {and} \bibinfo{person}{Noah Snavely}.} \bibinfo{year}{2024}\natexlab{}.
\newblock \showarticletitle{Neural gaffer: Relighting any object via diffusion}. In \bibinfo{booktitle}{\emph{The Thirty-eighth Annual Conference on Neural Information Processing Systems}}.
\newblock


\bibitem[Kajiya(1986)]%
        {kajiya1986rendering}
\bibfield{author}{\bibinfo{person}{James~T Kajiya}.} \bibinfo{year}{1986}\natexlab{}.
\newblock \showarticletitle{The rendering equation}. In \bibinfo{booktitle}{\emph{Proceedings of the 13th annual conference on Computer graphics and interactive techniques}}. \bibinfo{pages}{143--150}.
\newblock


\bibitem[Kavan et~al\mbox{.}(2008)]%
        {kavan2008geometric}
\bibfield{author}{\bibinfo{person}{Ladislav Kavan}, \bibinfo{person}{Steven Collins}, \bibinfo{person}{Ji{\v{r}}{\'\i} {\v{Z}}{\'a}ra}, {and} \bibinfo{person}{Carol O'Sullivan}.} \bibinfo{year}{2008}\natexlab{}.
\newblock \showarticletitle{Geometric skinning with approximate dual quaternion blending}.
\newblock \bibinfo{journal}{\emph{ACM Transactions on Graphics (TOG)}} \bibinfo{volume}{27}, \bibinfo{number}{4} (\bibinfo{year}{2008}), \bibinfo{pages}{1--23}.
\newblock


\bibitem[Kerbl et~al\mbox{.}(2023)]%
        {kerbl20233d}
\bibfield{author}{\bibinfo{person}{Bernhard Kerbl}, \bibinfo{person}{Georgios Kopanas}, \bibinfo{person}{Thomas Leimk{\"u}hler}, {and} \bibinfo{person}{George Drettakis}.} \bibinfo{year}{2023}\natexlab{}.
\newblock \showarticletitle{3d gaussian splatting for real-time radiance field rendering.}
\newblock \bibinfo{journal}{\emph{ACM Trans. Graph.}} \bibinfo{volume}{42}, \bibinfo{number}{4} (\bibinfo{year}{2023}), \bibinfo{pages}{139--1}.
\newblock


\bibitem[Khirodkar et~al\mbox{.}(2025)]%
        {khirodkar2025sapiens}
\bibfield{author}{\bibinfo{person}{Rawal Khirodkar}, \bibinfo{person}{Timur Bagautdinov}, \bibinfo{person}{Julieta Martinez}, \bibinfo{person}{Su Zhaoen}, \bibinfo{person}{Austin James}, \bibinfo{person}{Peter Selednik}, \bibinfo{person}{Stuart Anderson}, {and} \bibinfo{person}{Shunsuke Saito}.} \bibinfo{year}{2025}\natexlab{}.
\newblock \showarticletitle{Sapiens: Foundation for human vision models}. In \bibinfo{booktitle}{\emph{European Conference on Computer Vision}}. Springer, \bibinfo{pages}{206--228}.
\newblock


\bibitem[Kim et~al\mbox{.}(2024)]%
        {kim2024switchlight}
\bibfield{author}{\bibinfo{person}{Hoon Kim}, \bibinfo{person}{Minje Jang}, \bibinfo{person}{Wonjun Yoon}, \bibinfo{person}{Jisoo Lee}, \bibinfo{person}{Donghyun Na}, {and} \bibinfo{person}{Sanghyun Woo}.} \bibinfo{year}{2024}\natexlab{}.
\newblock \showarticletitle{SwitchLight: Co-design of Physics-driven Architecture and Pre-training Framework for Human Portrait Relighting}. In \bibinfo{booktitle}{\emph{Proceedings of the IEEE/CVF Conference on Computer Vision and Pattern Recognition}}. \bibinfo{pages}{25096--25106}.
\newblock


\bibitem[Laine et~al\mbox{.}(2020)]%
        {Laine2020diffrast}
\bibfield{author}{\bibinfo{person}{Samuli Laine}, \bibinfo{person}{Janne Hellsten}, \bibinfo{person}{Tero Karras}, \bibinfo{person}{Yeongho Seol}, \bibinfo{person}{Jaakko Lehtinen}, {and} \bibinfo{person}{Timo Aila}.} \bibinfo{year}{2020}\natexlab{}.
\newblock \showarticletitle{Modular primitives for high-performance differentiable rendering}.
\newblock \bibinfo{journal}{\emph{ACM Transactions on Graphics (ToG)}} \bibinfo{volume}{39}, \bibinfo{number}{6} (\bibinfo{year}{2020}), \bibinfo{pages}{1--14}.
\newblock


\bibitem[Lawrence et~al\mbox{.}(2024)]%
        {lawrence2024project}
\bibfield{author}{\bibinfo{person}{Jason Lawrence}, \bibinfo{person}{Ryan Overbeck}, \bibinfo{person}{Todd Prives}, \bibinfo{person}{Tommy Fortes}, \bibinfo{person}{Nikki Roth}, {and} \bibinfo{person}{Brett Newman}.} \bibinfo{year}{2024}\natexlab{}.
\newblock \showarticletitle{Project starline: A high-fidelity telepresence system}.
\newblock In \bibinfo{booktitle}{\emph{ACM SIGGRAPH 2024 Emerging Technologies}}. \bibinfo{pages}{1--2}.
\newblock


\bibitem[Lee et~al\mbox{.}(2025)]%
        {lee2025rewind}
\bibfield{author}{\bibinfo{person}{Jihyun Lee}, \bibinfo{person}{Weipeng Xu}, \bibinfo{person}{Alexander Richard}, \bibinfo{person}{Shih-En Wei}, \bibinfo{person}{Shunsuke Saito}, \bibinfo{person}{Shaojie Bai}, \bibinfo{person}{Te-Li Wang}, \bibinfo{person}{Minhyuk Sung}, \bibinfo{person}{Tae-Kyun Kim}, {and} \bibinfo{person}{Jason Saragih}.} \bibinfo{year}{2025}\natexlab{}.
\newblock \showarticletitle{REWIND: Real-Time Egocentric Whole-Body Motion Diffusion with Exemplar-Based Identity Conditioning}. In \bibinfo{booktitle}{\emph{Proceedings of the Computer Vision and Pattern Recognition Conference}}. \bibinfo{pages}{7095--7104}.
\newblock


\bibitem[Li et~al\mbox{.}(2024a)]%
        {li2024uravatar}
\bibfield{author}{\bibinfo{person}{Junxuan Li}, \bibinfo{person}{Chen Cao}, \bibinfo{person}{Gabriel Schwartz}, \bibinfo{person}{Rawal Khirodkar}, \bibinfo{person}{Christian Richardt}, \bibinfo{person}{Tomas Simon}, \bibinfo{person}{Yaser Sheikh}, {and} \bibinfo{person}{Shunsuke Saito}.} \bibinfo{year}{2024}\natexlab{a}.
\newblock \showarticletitle{Uravatar: Universal relightable gaussian codec avatars}. In \bibinfo{booktitle}{\emph{SIGGRAPH Asia 2024 Conference Papers}}. \bibinfo{pages}{1--11}.
\newblock


\bibitem[Li et~al\mbox{.}(2023a)]%
        {li2023ego}
\bibfield{author}{\bibinfo{person}{Jiaman Li}, \bibinfo{person}{Karen Liu}, {and} \bibinfo{person}{Jiajun Wu}.} \bibinfo{year}{2023}\natexlab{a}.
\newblock \showarticletitle{Ego-Body Pose Estimation via Ego-Head Pose Estimation}. In \bibinfo{booktitle}{\emph{Proceedings of the IEEE/CVF Conference on Computer Vision and Pattern Recognition}}. \bibinfo{pages}{17142--17151}.
\newblock


\bibitem[Li et~al\mbox{.}(2023b)]%
        {li2023megane}
\bibfield{author}{\bibinfo{person}{Junxuan Li}, \bibinfo{person}{Shunsuke Saito}, \bibinfo{person}{Tomas Simon}, \bibinfo{person}{Stephen Lombardi}, \bibinfo{person}{Hongdong Li}, {and} \bibinfo{person}{Jason Saragih}.} \bibinfo{year}{2023}\natexlab{b}.
\newblock \showarticletitle{Megane: Morphable eyeglass and avatar network}. In \bibinfo{booktitle}{\emph{Proceedings of the IEEE/CVF Conference on Computer Vision and Pattern Recognition}}. \bibinfo{pages}{12769--12779}.
\newblock


\bibitem[Li et~al\mbox{.}(2024b)]%
        {li2024animatablegaussians}
\bibfield{author}{\bibinfo{person}{Zhe Li}, \bibinfo{person}{Zerong Zheng}, \bibinfo{person}{Lizhen Wang}, {and} \bibinfo{person}{Yebin Liu}.} \bibinfo{year}{2024}\natexlab{b}.
\newblock \showarticletitle{Animatable Gaussians: Learning Pose-dependent Gaussian Maps for High-fidelity Human Avatar Modeling}. In \bibinfo{booktitle}{\emph{Proceedings of the IEEE/CVF Conference on Computer Vision and Pattern Recognition (CVPR)}}.
\newblock


\bibitem[Liu et~al\mbox{.}(2021)]%
        {liu2021neural}
\bibfield{author}{\bibinfo{person}{Lingjie Liu}, \bibinfo{person}{Marc Habermann}, \bibinfo{person}{Viktor Rudnev}, \bibinfo{person}{Kripasindhu Sarkar}, \bibinfo{person}{Jiatao Gu}, {and} \bibinfo{person}{Christian Theobalt}.} \bibinfo{year}{2021}\natexlab{}.
\newblock \showarticletitle{Neural actor: Neural free-view synthesis of human actors with pose control}.
\newblock \bibinfo{journal}{\emph{ACM transactions on graphics (TOG)}} \bibinfo{volume}{40}, \bibinfo{number}{6} (\bibinfo{year}{2021}), \bibinfo{pages}{1--16}.
\newblock


\bibitem[Loper et~al\mbox{.}(2023)]%
        {loper2023smpl}
\bibfield{author}{\bibinfo{person}{Matthew Loper}, \bibinfo{person}{Naureen Mahmood}, \bibinfo{person}{Javier Romero}, \bibinfo{person}{Gerard Pons-Moll}, {and} \bibinfo{person}{Michael~J Black}.} \bibinfo{year}{2023}\natexlab{}.
\newblock \showarticletitle{SMPL: A skinned multi-person linear model}.
\newblock In \bibinfo{booktitle}{\emph{Seminal Graphics Papers: Pushing the Boundaries, Volume 2}}. \bibinfo{pages}{851--866}.
\newblock


\bibitem[Luo et~al\mbox{.}(2024)]%
        {luo2024real}
\bibfield{author}{\bibinfo{person}{Zhengyi Luo}, \bibinfo{person}{Jinkun Cao}, \bibinfo{person}{Rawal Khirodkar}, \bibinfo{person}{Alexander Winkler}, \bibinfo{person}{Kris Kitani}, {and} \bibinfo{person}{Weipeng Xu}.} \bibinfo{year}{2024}\natexlab{}.
\newblock \showarticletitle{Real-Time Simulated Avatar from Head-Mounted Sensors}. In \bibinfo{booktitle}{\emph{Proceedings of the IEEE/CVF Conference on Computer Vision and Pattern Recognition}}. \bibinfo{pages}{571--581}.
\newblock


\bibitem[Luo et~al\mbox{.}(2021)]%
        {luo2021dynamics}
\bibfield{author}{\bibinfo{person}{Zhengyi Luo}, \bibinfo{person}{Ryo Hachiuma}, \bibinfo{person}{Ye Yuan}, {and} \bibinfo{person}{Kris Kitani}.} \bibinfo{year}{2021}\natexlab{}.
\newblock \showarticletitle{Dynamics-regulated kinematic policy for egocentric pose estimation}.
\newblock \bibinfo{journal}{\emph{Advances in Neural Information Processing Systems}}  \bibinfo{volume}{34} (\bibinfo{year}{2021}), \bibinfo{pages}{25019--25032}.
\newblock


\bibitem[Luvizon et~al\mbox{.}(2024)]%
        {relightneuralactor2024eccv}
\bibfield{author}{\bibinfo{person}{Diogo Luvizon}, \bibinfo{person}{Vladislav Golyanik}, \bibinfo{person}{Adam Kortylewski}, \bibinfo{person}{Marc Habermann}, {and} \bibinfo{person}{Christian Theobalt}.} \bibinfo{year}{2024}\natexlab{}.
\newblock \showarticletitle{Relightable Neural Actor with Intrinsic Decomposition and Pose Control}. In \bibinfo{booktitle}{\emph{European Conference on Computer Vision (ECCV)}}.
\newblock


\bibitem[Martinez et~al\mbox{.}(2017)]%
        {martinez2017simple}
\bibfield{author}{\bibinfo{person}{Julieta Martinez}, \bibinfo{person}{Rayat Hossain}, \bibinfo{person}{Javier Romero}, {and} \bibinfo{person}{James~J Little}.} \bibinfo{year}{2017}\natexlab{}.
\newblock \showarticletitle{A simple yet effective baseline for 3d human pose estimation}. In \bibinfo{booktitle}{\emph{Proceedings of the IEEE international conference on computer vision}}. \bibinfo{pages}{2640--2649}.
\newblock


\bibitem[Meka et~al\mbox{.}(2020)]%
        {meka2020deep}
\bibfield{author}{\bibinfo{person}{Abhimitra Meka}, \bibinfo{person}{Rohit Pandey}, \bibinfo{person}{Christian Haene}, \bibinfo{person}{Sergio Orts-Escolano}, \bibinfo{person}{Peter Barnum}, \bibinfo{person}{Philip David-Son}, \bibinfo{person}{Daniel Erickson}, \bibinfo{person}{Yinda Zhang}, \bibinfo{person}{Jonathan Taylor}, \bibinfo{person}{Sofien Bouaziz}, {et~al\mbox{.}}} \bibinfo{year}{2020}\natexlab{}.
\newblock \showarticletitle{Deep relightable textures: volumetric performance capture with neural rendering}.
\newblock \bibinfo{journal}{\emph{ACM Transactions on Graphics (TOG)}} \bibinfo{volume}{39}, \bibinfo{number}{6} (\bibinfo{year}{2020}), \bibinfo{pages}{1--21}.
\newblock


\bibitem[Mildenhall et~al\mbox{.}(2020)]%
        {mildenhall2020nerf}
\bibfield{author}{\bibinfo{person}{Ben Mildenhall}, \bibinfo{person}{Pratul~P Srinivasan}, \bibinfo{person}{Matthew Tancik}, \bibinfo{person}{Jonathan~T Barron}, \bibinfo{person}{Ravi Ramamoorthi}, {and} \bibinfo{person}{Ren Ng}.} \bibinfo{year}{2020}\natexlab{}.
\newblock \showarticletitle{NeRF: Representing Scenes as Neural Radiance Fields for View Synthesis}. In \bibinfo{booktitle}{\emph{European Conference on Computer Vision}}. Springer, \bibinfo{pages}{405--421}.
\newblock


\bibitem[Milgram and Kishino(1994)]%
        {milgram1994taxonomy}
\bibfield{author}{\bibinfo{person}{Paul Milgram} {and} \bibinfo{person}{Fumio Kishino}.} \bibinfo{year}{1994}\natexlab{}.
\newblock \showarticletitle{A taxonomy of mixed reality visual displays}.
\newblock \bibinfo{journal}{\emph{IEICE TRANSACTIONS on Information and Systems}} \bibinfo{volume}{77}, \bibinfo{number}{12} (\bibinfo{year}{1994}), \bibinfo{pages}{1321--1329}.
\newblock


\bibitem[Nicodemus et~al\mbox{.}(1977)]%
        {nicodemus1977geometrical}
\bibfield{author}{\bibinfo{person}{Fred~Edwin Nicodemus}, \bibinfo{person}{Joseph~C Richmond}, \bibinfo{person}{Jack~J Hsia}, \bibinfo{person}{Irving~W Ginsberg}, \bibinfo{person}{Thomas Limperis}, {et~al\mbox{.}}} \bibinfo{year}{1977}\natexlab{}.
\newblock \bibinfo{booktitle}{\emph{Geometrical considerations and nomenclature for reflectance}}. Vol.~\bibinfo{volume}{160}.
\newblock \bibinfo{publisher}{US Department of Commerce, National Bureau of Standards Washington, DC, USA}.
\newblock


\bibitem[Nimier-David et~al\mbox{.}(2019)]%
        {nimier2019mitsuba}
\bibfield{author}{\bibinfo{person}{Merlin Nimier-David}, \bibinfo{person}{Delio Vicini}, \bibinfo{person}{Tizian Zeltner}, {and} \bibinfo{person}{Wenzel Jakob}.} \bibinfo{year}{2019}\natexlab{}.
\newblock \showarticletitle{Mitsuba 2: A retargetable forward and inverse renderer}.
\newblock \bibinfo{journal}{\emph{ACM Transactions on Graphics (ToG)}} \bibinfo{volume}{38}, \bibinfo{number}{6} (\bibinfo{year}{2019}), \bibinfo{pages}{1--17}.
\newblock


\bibitem[Pandey et~al\mbox{.}(2021)]%
        {pandey2021total}
\bibfield{author}{\bibinfo{person}{Rohit Pandey}, \bibinfo{person}{Sergio Orts-Escolano}, \bibinfo{person}{Chloe Legendre}, \bibinfo{person}{Christian Haene}, \bibinfo{person}{Sofien Bouaziz}, \bibinfo{person}{Christoph Rhemann}, \bibinfo{person}{Paul~E Debevec}, {and} \bibinfo{person}{Sean~Ryan Fanello}.} \bibinfo{year}{2021}\natexlab{}.
\newblock \showarticletitle{Total relighting: learning to relight portraits for background replacement.}
\newblock \bibinfo{journal}{\emph{ACM Trans. Graph.}} \bibinfo{volume}{40}, \bibinfo{number}{4} (\bibinfo{year}{2021}), \bibinfo{pages}{43--1}.
\newblock


\bibitem[Pang et~al\mbox{.}(2024)]%
        {Pang_2024_CVPR}
\bibfield{author}{\bibinfo{person}{Haokai Pang}, \bibinfo{person}{Heming Zhu}, \bibinfo{person}{Adam Kortylewski}, \bibinfo{person}{Christian Theobalt}, {and} \bibinfo{person}{Marc Habermann}.} \bibinfo{year}{2024}\natexlab{}.
\newblock \showarticletitle{ASH: Animatable Gaussian Splats for Efficient and Photoreal Human Rendering}. In \bibinfo{booktitle}{\emph{Proceedings of the IEEE/CVF Conference on Computer Vision and Pattern Recognition (CVPR)}}. \bibinfo{pages}{1165--1175}.
\newblock


\bibitem[Parker et~al\mbox{.}(2010)]%
        {parker2010optix}
\bibfield{author}{\bibinfo{person}{Steven~G Parker}, \bibinfo{person}{James Bigler}, \bibinfo{person}{Andreas Dietrich}, \bibinfo{person}{Heiko Friedrich}, \bibinfo{person}{Jared Hoberock}, \bibinfo{person}{David Luebke}, \bibinfo{person}{David McAllister}, \bibinfo{person}{Morgan McGuire}, \bibinfo{person}{Keith Morley}, \bibinfo{person}{Austin Robison}, {et~al\mbox{.}}} \bibinfo{year}{2010}\natexlab{}.
\newblock \showarticletitle{Optix: a general purpose ray tracing engine}.
\newblock \bibinfo{journal}{\emph{Acm transactions on graphics (tog)}} \bibinfo{volume}{29}, \bibinfo{number}{4} (\bibinfo{year}{2010}), \bibinfo{pages}{1--13}.
\newblock


\bibitem[Pavlakos et~al\mbox{.}(2019)]%
        {pavlakos2019expressive}
\bibfield{author}{\bibinfo{person}{Georgios Pavlakos}, \bibinfo{person}{Vasileios Choutas}, \bibinfo{person}{Nima Ghorbani}, \bibinfo{person}{Timo Bolkart}, \bibinfo{person}{Ahmed~AA Osman}, \bibinfo{person}{Dimitrios Tzionas}, {and} \bibinfo{person}{Michael~J Black}.} \bibinfo{year}{2019}\natexlab{}.
\newblock \showarticletitle{Expressive body capture: 3d hands, face, and body from a single image}. In \bibinfo{booktitle}{\emph{Proceedings of the IEEE/CVF conference on computer vision and pattern recognition}}. \bibinfo{pages}{10975--10985}.
\newblock


\bibitem[Peng et~al\mbox{.}(2021)]%
        {peng2021neural}
\bibfield{author}{\bibinfo{person}{Sida Peng}, \bibinfo{person}{Yuanqing Zhang}, \bibinfo{person}{Yinghao Xu}, \bibinfo{person}{Qianqian Wang}, \bibinfo{person}{Qing Shuai}, \bibinfo{person}{Hujun Bao}, {and} \bibinfo{person}{Xiaowei Zhou}.} \bibinfo{year}{2021}\natexlab{}.
\newblock \showarticletitle{Neural body: Implicit neural representations with structured latent codes for novel view synthesis of dynamic humans}. In \bibinfo{booktitle}{\emph{Proceedings of the IEEE/CVF Conference on Computer Vision and Pattern Recognition}}. \bibinfo{pages}{9054--9063}.
\newblock


\bibitem[Poirier-Ginter et~al\mbox{.}(2024)]%
        {poirier2024diffusion}
\bibfield{author}{\bibinfo{person}{Yohan Poirier-Ginter}, \bibinfo{person}{Alban Gauthier}, \bibinfo{person}{Julien Phillip}, \bibinfo{person}{J-F Lalonde}, {and} \bibinfo{person}{George Drettakis}.} \bibinfo{year}{2024}\natexlab{}.
\newblock \showarticletitle{A Diffusion Approach to Radiance Field Relighting using Multi-Illumination Synthesis}. In \bibinfo{booktitle}{\emph{Computer Graphics Forum}}, Vol.~\bibinfo{volume}{43}. Wiley Online Library, \bibinfo{pages}{e15147}.
\newblock


\bibitem[Ramamoorthi and Hanrahan(2001)]%
        {ramamoorthi2001signal}
\bibfield{author}{\bibinfo{person}{Ravi Ramamoorthi} {and} \bibinfo{person}{Pat Hanrahan}.} \bibinfo{year}{2001}\natexlab{}.
\newblock \showarticletitle{A signal-processing framework for inverse rendering}. In \bibinfo{booktitle}{\emph{Proceedings of the 28th annual conference on Computer graphics and interactive techniques}}. \bibinfo{pages}{117--128}.
\newblock


\bibitem[Remelli et~al\mbox{.}(2022)]%
        {remelli2022drivable}
\bibfield{author}{\bibinfo{person}{Edoardo Remelli}, \bibinfo{person}{Timur Bagautdinov}, \bibinfo{person}{Shunsuke Saito}, \bibinfo{person}{Chenglei Wu}, \bibinfo{person}{Tomas Simon}, \bibinfo{person}{Shih-En Wei}, \bibinfo{person}{Kaiwen Guo}, \bibinfo{person}{Zhe Cao}, \bibinfo{person}{Fabian Prada}, \bibinfo{person}{Jason Saragih}, {et~al\mbox{.}}} \bibinfo{year}{2022}\natexlab{}.
\newblock \showarticletitle{Drivable volumetric avatars using texel-aligned features}. In \bibinfo{booktitle}{\emph{ACM SIGGRAPH 2022 Conference Proceedings}}. \bibinfo{pages}{1--9}.
\newblock


\bibitem[Ren et~al\mbox{.}(2024)]%
        {ren2024relightful}
\bibfield{author}{\bibinfo{person}{Mengwei Ren}, \bibinfo{person}{Wei Xiong}, \bibinfo{person}{Jae~Shin Yoon}, \bibinfo{person}{Zhixin Shu}, \bibinfo{person}{Jianming Zhang}, \bibinfo{person}{HyunJoon Jung}, \bibinfo{person}{Guido Gerig}, {and} \bibinfo{person}{He Zhang}.} \bibinfo{year}{2024}\natexlab{}.
\newblock \showarticletitle{Relightful Harmonization: Lighting-aware Portrait Background Replacement}. In \bibinfo{booktitle}{\emph{Proceedings of the IEEE/CVF Conference on Computer Vision and Pattern Recognition}}. \bibinfo{pages}{6452--6462}.
\newblock


\bibitem[Rhodin et~al\mbox{.}(2016)]%
        {rhodin2016egocap}
\bibfield{author}{\bibinfo{person}{Helge Rhodin}, \bibinfo{person}{Christian Richardt}, \bibinfo{person}{Dan Casas}, \bibinfo{person}{Eldar Insafutdinov}, \bibinfo{person}{Mohammad Shafiei}, \bibinfo{person}{Hans-Peter Seidel}, \bibinfo{person}{Bernt Schiele}, {and} \bibinfo{person}{Christian Theobalt}.} \bibinfo{year}{2016}\natexlab{}.
\newblock \showarticletitle{Egocap: egocentric marker-less motion capture with two fisheye cameras}.
\newblock \bibinfo{journal}{\emph{ACM Transactions on Graphics (TOG)}} \bibinfo{volume}{35}, \bibinfo{number}{6} (\bibinfo{year}{2016}), \bibinfo{pages}{1--11}.
\newblock


\bibitem[Saito et~al\mbox{.}(2024)]%
        {saito2024relightable}
\bibfield{author}{\bibinfo{person}{Shunsuke Saito}, \bibinfo{person}{Gabriel Schwartz}, \bibinfo{person}{Tomas Simon}, \bibinfo{person}{Junxuan Li}, {and} \bibinfo{person}{Giljoo Nam}.} \bibinfo{year}{2024}\natexlab{}.
\newblock \showarticletitle{Relightable gaussian codec avatars}. In \bibinfo{booktitle}{\emph{Proceedings of the IEEE/CVF Conference on Computer Vision and Pattern Recognition}}. \bibinfo{pages}{130--141}.
\newblock


\bibitem[Shao et~al\mbox{.}(2023)]%
        {shao2023tensor4d}
\bibfield{author}{\bibinfo{person}{Ruizhi Shao}, \bibinfo{person}{Zerong Zheng}, \bibinfo{person}{Hanzhang Tu}, \bibinfo{person}{Boning Liu}, \bibinfo{person}{Hongwen Zhang}, {and} \bibinfo{person}{Yebin Liu}.} \bibinfo{year}{2023}\natexlab{}.
\newblock \showarticletitle{Tensor4d: Efficient neural 4d decomposition for high-fidelity dynamic reconstruction and rendering}. In \bibinfo{booktitle}{\emph{Proceedings of the IEEE/CVF Conference on Computer Vision and Pattern Recognition}}. \bibinfo{pages}{16632--16642}.
\newblock


\bibitem[Shetty et~al\mbox{.}(2024)]%
        {shetty2024holoported}
\bibfield{author}{\bibinfo{person}{Ashwath Shetty}, \bibinfo{person}{Marc Habermann}, \bibinfo{person}{Guoxing Sun}, \bibinfo{person}{Diogo Luvizon}, \bibinfo{person}{Vladislav Golyanik}, {and} \bibinfo{person}{Christian Theobalt}.} \bibinfo{year}{2024}\natexlab{}.
\newblock \showarticletitle{Holoported Characters: Real-time Free-viewpoint Rendering of Humans from Sparse RGB Cameras}. In \bibinfo{booktitle}{\emph{Proceedings of the IEEE/CVF Conference on Computer Vision and Pattern Recognition}}. \bibinfo{pages}{1206--1215}.
\newblock


\bibitem[Singh et~al\mbox{.}(2025)]%
        {singh2025relightable}
\bibfield{author}{\bibinfo{person}{Kunwar~Maheep Singh}, \bibinfo{person}{Jianchun Chen}, \bibinfo{person}{Vladislav Golyanik}, \bibinfo{person}{Stephan~J Garbin}, \bibinfo{person}{Thabo Beeler}, \bibinfo{person}{Rishabh Dabral}, \bibinfo{person}{Marc Habermann}, {and} \bibinfo{person}{Christian Theobalt}.} \bibinfo{year}{2025}\natexlab{}.
\newblock \showarticletitle{Relightable Holoported Characters: Capturing and Relighting Dynamic Human Performance from Sparse Views}.
\newblock \bibinfo{journal}{\emph{arXiv preprint arXiv:2512.00255}} (\bibinfo{year}{2025}).
\newblock


\bibitem[Sloan et~al\mbox{.}(2023)]%
        {sloan2023precomputed}
\bibfield{author}{\bibinfo{person}{Peter-Pike Sloan}, \bibinfo{person}{Jan Kautz}, {and} \bibinfo{person}{John Snyder}.} \bibinfo{year}{2023}\natexlab{}.
\newblock \showarticletitle{Precomputed radiance transfer for real-time rendering in dynamic, low-frequency lighting environments}.
\newblock In \bibinfo{booktitle}{\emph{Seminal Graphics Papers: Pushing the Boundaries, Volume 2}}. \bibinfo{pages}{339--348}.
\newblock


\bibitem[Sumner et~al\mbox{.}(2007)]%
        {sumner2007embedded}
\bibfield{author}{\bibinfo{person}{Robert~W Sumner}, \bibinfo{person}{Johannes Schmid}, {and} \bibinfo{person}{Mark Pauly}.} \bibinfo{year}{2007}\natexlab{}.
\newblock \showarticletitle{Embedded deformation for shape manipulation}.
\newblock In \bibinfo{booktitle}{\emph{ACM siggraph 2007 papers}}. \bibinfo{pages}{80--es}.
\newblock


\bibitem[Teed and Deng(2020)]%
        {teed2020raft}
\bibfield{author}{\bibinfo{person}{Zachary Teed} {and} \bibinfo{person}{Jia Deng}.} \bibinfo{year}{2020}\natexlab{}.
\newblock \showarticletitle{Raft: Recurrent all-pairs field transforms for optical flow}. In \bibinfo{booktitle}{\emph{European conference on computer vision}}. Springer, \bibinfo{pages}{402--419}.
\newblock


\bibitem[TheCaptury(2020)]%
        {thecaptury2020captury}
\bibfield{author}{\bibinfo{person}{TheCaptury}.} \bibinfo{year}{2020}\natexlab{}.
\newblock \bibinfo{title}{Captury motion capture redefined: Go markerless.}
\newblock
\urldef\tempurl%
\url{https://captury.com/}
\showURL{%
\tempurl}


\bibitem[Tome et~al\mbox{.}(2019)]%
        {tome2019xr}
\bibfield{author}{\bibinfo{person}{Denis Tome}, \bibinfo{person}{Patrick Peluse}, \bibinfo{person}{Lourdes Agapito}, {and} \bibinfo{person}{Hernan Badino}.} \bibinfo{year}{2019}\natexlab{}.
\newblock \showarticletitle{xr-egopose: Egocentric 3d human pose from an hmd camera}. In \bibinfo{booktitle}{\emph{Proceedings of the IEEE/CVF International Conference on Computer Vision}}. \bibinfo{pages}{7728--7738}.
\newblock


\bibitem[Vaswani et~al\mbox{.}(2017)]%
        {vaswani2017attention}
\bibfield{author}{\bibinfo{person}{Ashish Vaswani}, \bibinfo{person}{Noam Shazeer}, \bibinfo{person}{Niki Parmar}, \bibinfo{person}{Jakob Uszkoreit}, \bibinfo{person}{Llion Jones}, \bibinfo{person}{Aidan~N Gomez}, \bibinfo{person}{{\L}ukasz Kaiser}, {and} \bibinfo{person}{Illia Polosukhin}.} \bibinfo{year}{2017}\natexlab{}.
\newblock \showarticletitle{Attention is all you need}.
\newblock \bibinfo{journal}{\emph{Advances in neural information processing systems}}  \bibinfo{volume}{30} (\bibinfo{year}{2017}).
\newblock


\bibitem[Wang et~al\mbox{.}(2025a)]%
        {wang2025vggt}
\bibfield{author}{\bibinfo{person}{Jianyuan Wang}, \bibinfo{person}{Minghao Chen}, \bibinfo{person}{Nikita Karaev}, \bibinfo{person}{Andrea Vedaldi}, \bibinfo{person}{Christian Rupprecht}, {and} \bibinfo{person}{David Novotny}.} \bibinfo{year}{2025}\natexlab{a}.
\newblock \showarticletitle{Vggt: Visual geometry grounded transformer}. In \bibinfo{booktitle}{\emph{Proceedings of the Computer Vision and Pattern Recognition Conference}}. \bibinfo{pages}{5294--5306}.
\newblock


\bibitem[Wang et~al\mbox{.}(2022)]%
        {wang2022estimating}
\bibfield{author}{\bibinfo{person}{Jian Wang}, \bibinfo{person}{Lingjie Liu}, \bibinfo{person}{Weipeng Xu}, \bibinfo{person}{Kripasindhu Sarkar}, \bibinfo{person}{Diogo Luvizon}, {and} \bibinfo{person}{Christian Theobalt}.} \bibinfo{year}{2022}\natexlab{}.
\newblock \showarticletitle{Estimating egocentric 3d human pose in the wild with external weak supervision}. In \bibinfo{booktitle}{\emph{Proceedings of the IEEE/CVF Conference on Computer Vision and Pattern Recognition}}. \bibinfo{pages}{13157--13166}.
\newblock


\bibitem[Wang et~al\mbox{.}(2021)]%
        {wang2021estimating}
\bibfield{author}{\bibinfo{person}{Jian Wang}, \bibinfo{person}{Lingjie Liu}, \bibinfo{person}{Weipeng Xu}, \bibinfo{person}{Kripasindhu Sarkar}, {and} \bibinfo{person}{Christian Theobalt}.} \bibinfo{year}{2021}\natexlab{}.
\newblock \showarticletitle{Estimating egocentric 3d human pose in global space}. In \bibinfo{booktitle}{\emph{Proceedings of the IEEE/CVF International Conference on Computer Vision}}. \bibinfo{pages}{11500--11509}.
\newblock


\bibitem[Wang et~al\mbox{.}(2023b)]%
        {wang2023scene}
\bibfield{author}{\bibinfo{person}{Jian Wang}, \bibinfo{person}{Diogo Luvizon}, \bibinfo{person}{Weipeng Xu}, \bibinfo{person}{Lingjie Liu}, \bibinfo{person}{Kripasindhu Sarkar}, {and} \bibinfo{person}{Christian Theobalt}.} \bibinfo{year}{2023}\natexlab{b}.
\newblock \showarticletitle{Scene-aware Egocentric 3D Human Pose Estimation}. In \bibinfo{booktitle}{\emph{Proceedings of the IEEE/CVF Conference on Computer Vision and Pattern Recognition}}. \bibinfo{pages}{13031--13040}.
\newblock


\bibitem[Wang et~al\mbox{.}(2024)]%
        {wang2024intrinsicavatar}
\bibfield{author}{\bibinfo{person}{Shaofei Wang}, \bibinfo{person}{Bozidar Antic}, \bibinfo{person}{Andreas Geiger}, {and} \bibinfo{person}{Siyu Tang}.} \bibinfo{year}{2024}\natexlab{}.
\newblock \showarticletitle{IntrinsicAvatar: Physically Based Inverse Rendering of Dynamic Humans from Monocular Videos via Explicit Ray Tracing}. In \bibinfo{booktitle}{\emph{Proceedings of the IEEE/CVF Conference on Computer Vision and Pattern Recognition}}. \bibinfo{pages}{1877--1888}.
\newblock


\bibitem[Wang et~al\mbox{.}(2025b)]%
        {wang2025relightable}
\bibfield{author}{\bibinfo{person}{Shaofei Wang}, \bibinfo{person}{Tomas Simon}, \bibinfo{person}{Igor Santesteban}, \bibinfo{person}{Timur Bagautdinov}, \bibinfo{person}{Junxuan Li}, \bibinfo{person}{Vasu Agrawal}, \bibinfo{person}{Fabian Prada}, \bibinfo{person}{Shoou-I Yu}, \bibinfo{person}{Pace Nalbone}, \bibinfo{person}{Matt Gramlich}, {et~al\mbox{.}}} \bibinfo{year}{2025}\natexlab{b}.
\newblock \showarticletitle{Relightable Full-Body Gaussian Codec Avatars}.
\newblock \bibinfo{journal}{\emph{arXiv preprint arXiv:2501.14726}} (\bibinfo{year}{2025}).
\newblock


\bibitem[Wang et~al\mbox{.}(2023a)]%
        {wang2023neus2}
\bibfield{author}{\bibinfo{person}{Yiming Wang}, \bibinfo{person}{Qin Han}, \bibinfo{person}{Marc Habermann}, \bibinfo{person}{Kostas Daniilidis}, \bibinfo{person}{Christian Theobalt}, {and} \bibinfo{person}{Lingjie Liu}.} \bibinfo{year}{2023}\natexlab{a}.
\newblock \showarticletitle{Neus2: Fast learning of neural implicit surfaces for multi-view reconstruction}. In \bibinfo{booktitle}{\emph{Proceedings of the IEEE/CVF International Conference on Computer Vision}}. \bibinfo{pages}{3295--3306}.
\newblock


\bibitem[Wang et~al\mbox{.}(2018)]%
        {wang2018image}
\bibfield{author}{\bibinfo{person}{Yi Wang}, \bibinfo{person}{Xin Tao}, \bibinfo{person}{Xiaojuan Qi}, \bibinfo{person}{Xiaoyong Shen}, {and} \bibinfo{person}{Jiaya Jia}.} \bibinfo{year}{2018}\natexlab{}.
\newblock \showarticletitle{Image inpainting via generative multi-column convolutional neural networks}.
\newblock \bibinfo{journal}{\emph{Advances in neural information processing systems}}  \bibinfo{volume}{31} (\bibinfo{year}{2018}).
\newblock


\bibitem[Wang(2004)]%
        {wang2004image}
\bibfield{author}{\bibinfo{person}{Zhou Wang}.} \bibinfo{year}{2004}\natexlab{}.
\newblock \showarticletitle{Image quality assessment: from error visibility to structural similarity}.
\newblock \bibinfo{journal}{\emph{IEEE transactions on image processing}} \bibinfo{volume}{13}, \bibinfo{number}{4} (\bibinfo{year}{2004}), \bibinfo{pages}{600--612}.
\newblock


\bibitem[Weng et~al\mbox{.}(2022)]%
        {weng2022humannerf}
\bibfield{author}{\bibinfo{person}{Chung-Yi Weng}, \bibinfo{person}{Brian Curless}, \bibinfo{person}{Pratul~P Srinivasan}, \bibinfo{person}{Jonathan~T Barron}, {and} \bibinfo{person}{Ira Kemelmacher-Shlizerman}.} \bibinfo{year}{2022}\natexlab{}.
\newblock \showarticletitle{Humannerf: Free-viewpoint rendering of moving people from monocular video}. In \bibinfo{booktitle}{\emph{Proceedings of the IEEE/CVF conference on computer vision and pattern Recognition}}. \bibinfo{pages}{16210--16220}.
\newblock


\bibitem[Xiang et~al\mbox{.}(2022)]%
        {xiang2022dressing}
\bibfield{author}{\bibinfo{person}{Donglai Xiang}, \bibinfo{person}{Timur Bagautdinov}, \bibinfo{person}{Tuur Stuyck}, \bibinfo{person}{Fabian Prada}, \bibinfo{person}{Javier Romero}, \bibinfo{person}{Weipeng Xu}, \bibinfo{person}{Shunsuke Saito}, \bibinfo{person}{Jingfan Guo}, \bibinfo{person}{Breannan Smith}, \bibinfo{person}{Takaaki Shiratori}, {et~al\mbox{.}}} \bibinfo{year}{2022}\natexlab{}.
\newblock \showarticletitle{Dressing avatars: Deep photorealistic appearance for physically simulated clothing}.
\newblock \bibinfo{journal}{\emph{ACM Transactions on Graphics (TOG)}} \bibinfo{volume}{41}, \bibinfo{number}{6} (\bibinfo{year}{2022}), \bibinfo{pages}{1--15}.
\newblock


\bibitem[Xiang et~al\mbox{.}(2021)]%
        {xiang2021modeling}
\bibfield{author}{\bibinfo{person}{Donglai Xiang}, \bibinfo{person}{Fabian Prada}, \bibinfo{person}{Timur Bagautdinov}, \bibinfo{person}{Weipeng Xu}, \bibinfo{person}{Yuan Dong}, \bibinfo{person}{He Wen}, \bibinfo{person}{Jessica Hodgins}, {and} \bibinfo{person}{Chenglei Wu}.} \bibinfo{year}{2021}\natexlab{}.
\newblock \showarticletitle{Modeling clothing as a separate layer for an animatable human avatar}.
\newblock \bibinfo{journal}{\emph{ACM Transactions on Graphics (TOG)}} \bibinfo{volume}{40}, \bibinfo{number}{6} (\bibinfo{year}{2021}), \bibinfo{pages}{1--15}.
\newblock


\bibitem[Xiang et~al\mbox{.}(2023)]%
        {xiang2023drivable}
\bibfield{author}{\bibinfo{person}{Donglai Xiang}, \bibinfo{person}{Fabian Prada}, \bibinfo{person}{Zhe Cao}, \bibinfo{person}{Kaiwen Guo}, \bibinfo{person}{Chenglei Wu}, \bibinfo{person}{Jessica Hodgins}, {and} \bibinfo{person}{Timur Bagautdinov}.} \bibinfo{year}{2023}\natexlab{}.
\newblock \showarticletitle{Drivable avatar clothing: Faithful full-body telepresence with dynamic clothing driven by sparse rgb-d input}. In \bibinfo{booktitle}{\emph{SIGGRAPH Asia 2023 Conference Papers}}. \bibinfo{pages}{1--11}.
\newblock


\bibitem[Xu et~al\mbox{.}(2019)]%
        {xu2019mo}
\bibfield{author}{\bibinfo{person}{Weipeng Xu}, \bibinfo{person}{Avishek Chatterjee}, \bibinfo{person}{Michael Zollhoefer}, \bibinfo{person}{Helge Rhodin}, \bibinfo{person}{Pascal Fua}, \bibinfo{person}{Hans-Peter Seidel}, {and} \bibinfo{person}{Christian Theobalt}.} \bibinfo{year}{2019}\natexlab{}.
\newblock \showarticletitle{Mo\textsuperscript{2}Cap\textsuperscript{2}: Real-time mobile 3d motion capture with a cap-mounted fisheye camera}.
\newblock \bibinfo{journal}{\emph{IEEE transactions on visualization and computer graphics}} \bibinfo{volume}{25}, \bibinfo{number}{5} (\bibinfo{year}{2019}), \bibinfo{pages}{2093--2101}.
\newblock


\bibitem[Xu et~al\mbox{.}(2024)]%
        {xu2024relightable}
\bibfield{author}{\bibinfo{person}{Zhen Xu}, \bibinfo{person}{Sida Peng}, \bibinfo{person}{Chen Geng}, \bibinfo{person}{Linzhan Mou}, \bibinfo{person}{Zihan Yan}, \bibinfo{person}{Jiaming Sun}, \bibinfo{person}{Hujun Bao}, {and} \bibinfo{person}{Xiaowei Zhou}.} \bibinfo{year}{2024}\natexlab{}.
\newblock \showarticletitle{Relightable and Animatable Neural Avatar from Sparse-View Video}. In \bibinfo{booktitle}{\emph{CVPR}}.
\newblock


\bibitem[Yang et~al\mbox{.}(2024)]%
        {yang2024depth}
\bibfield{author}{\bibinfo{person}{Lihe Yang}, \bibinfo{person}{Bingyi Kang}, \bibinfo{person}{Zilong Huang}, \bibinfo{person}{Zhen Zhao}, \bibinfo{person}{Xiaogang Xu}, \bibinfo{person}{Jiashi Feng}, {and} \bibinfo{person}{Hengshuang Zhao}.} \bibinfo{year}{2024}\natexlab{}.
\newblock \showarticletitle{Depth anything v2}.
\newblock \bibinfo{journal}{\emph{Advances in Neural Information Processing Systems}}  \bibinfo{volume}{37} (\bibinfo{year}{2024}), \bibinfo{pages}{21875--21911}.
\newblock


\bibitem[Yi et~al\mbox{.}(2025)]%
        {yi2025estimating}
\bibfield{author}{\bibinfo{person}{Brent Yi}, \bibinfo{person}{Vickie Ye}, \bibinfo{person}{Maya Zheng}, \bibinfo{person}{Yunqi Li}, \bibinfo{person}{Lea M{\"u}ller}, \bibinfo{person}{Georgios Pavlakos}, \bibinfo{person}{Yi Ma}, \bibinfo{person}{Jitendra Malik}, {and} \bibinfo{person}{Angjoo Kanazawa}.} \bibinfo{year}{2025}\natexlab{}.
\newblock \showarticletitle{Estimating body and hand motion in an ego-sensed world}. In \bibinfo{booktitle}{\emph{Proceedings of the Computer Vision and Pattern Recognition Conference}}. \bibinfo{pages}{7072--7084}.
\newblock


\bibitem[Yoon et~al\mbox{.}(2024)]%
        {yoon2024generative}
\bibfield{author}{\bibinfo{person}{Jae~Shin Yoon}, \bibinfo{person}{Zhixin Shu}, \bibinfo{person}{Mengwei Ren}, \bibinfo{person}{Cecilia Zhang}, \bibinfo{person}{Yannick Hold-Geoffroy}, \bibinfo{person}{Krishna~kumar Singh}, {and} \bibinfo{person}{He Zhang}.} \bibinfo{year}{2024}\natexlab{}.
\newblock \showarticletitle{Generative Portrait Shadow Removal}.
\newblock \bibinfo{journal}{\emph{ACM Transactions on Graphics (TOG)}} \bibinfo{volume}{43}, \bibinfo{number}{6} (\bibinfo{year}{2024}), \bibinfo{pages}{1--13}.
\newblock


\bibitem[Yuan and Kitani(2019)]%
        {yuan2019ego}
\bibfield{author}{\bibinfo{person}{Ye Yuan} {and} \bibinfo{person}{Kris Kitani}.} \bibinfo{year}{2019}\natexlab{}.
\newblock \showarticletitle{Ego-pose estimation and forecasting as real-time pd control}. In \bibinfo{booktitle}{\emph{Proceedings of the IEEE/CVF International Conference on Computer Vision}}. \bibinfo{pages}{10082--10092}.
\newblock


\bibitem[Zhan et~al\mbox{.}(2025)]%
        {zhan2025interactive}
\bibfield{author}{\bibinfo{person}{Youyi Zhan}, \bibinfo{person}{Tianjia Shao}, \bibinfo{person}{He Wang}, \bibinfo{person}{Yin Yang}, {and} \bibinfo{person}{Kun Zhou}.} \bibinfo{year}{2025}\natexlab{}.
\newblock \showarticletitle{Interactive rendering of relightable and animatable gaussian avatars}.
\newblock \bibinfo{journal}{\emph{IEEE Transactions on Visualization and Computer Graphics}} (\bibinfo{year}{2025}).
\newblock


\bibitem[Zhang et~al\mbox{.}(2025)]%
        {zhang2025scaling}
\bibfield{author}{\bibinfo{person}{Lvmin Zhang}, \bibinfo{person}{Anyi Rao}, {and} \bibinfo{person}{Maneesh Agrawala}.} \bibinfo{year}{2025}\natexlab{}.
\newblock \showarticletitle{Scaling In-the-Wild Training for Diffusion-based Illumination Harmonization and Editing by Imposing Consistent Light Transport}. In \bibinfo{booktitle}{\emph{The Thirteenth International Conference on Learning Representations}}.
\newblock
\urldef\tempurl%
\url{https://openreview.net/forum?id=u1cQYxRI1H}
\showURL{%
\tempurl}


\bibitem[Zhang et~al\mbox{.}(2018)]%
        {zhang2018perceptual}
\bibfield{author}{\bibinfo{person}{Richard Zhang}, \bibinfo{person}{Phillip Isola}, \bibinfo{person}{Alexei~A Efros}, \bibinfo{person}{Eli Shechtman}, {and} \bibinfo{person}{Oliver Wang}.} \bibinfo{year}{2018}\natexlab{}.
\newblock \showarticletitle{The Unreasonable Effectiveness of Deep Features as a Perceptual Metric}. In \bibinfo{booktitle}{\emph{CVPR}}.
\newblock


\bibitem[Zhang et~al\mbox{.}(2024)]%
        {zhang2024relitlrm}
\bibfield{author}{\bibinfo{person}{Tianyuan Zhang}, \bibinfo{person}{Zhengfei Kuang}, \bibinfo{person}{Haian Jin}, \bibinfo{person}{Zexiang Xu}, \bibinfo{person}{Sai Bi}, \bibinfo{person}{Hao Tan}, \bibinfo{person}{He Zhang}, \bibinfo{person}{Yiwei Hu}, \bibinfo{person}{Milos Hasan}, \bibinfo{person}{William~T Freeman}, {et~al\mbox{.}}} \bibinfo{year}{2024}\natexlab{}.
\newblock \showarticletitle{RelitLRM: Generative Relightable Radiance for Large Reconstruction Models}.
\newblock \bibinfo{journal}{\emph{arXiv preprint arXiv:2410.06231}} (\bibinfo{year}{2024}).
\newblock


\bibitem[Zhang et~al\mbox{.}(2021)]%
        {zhang2021nerfactor}
\bibfield{author}{\bibinfo{person}{Xiuming Zhang}, \bibinfo{person}{Pratul~P Srinivasan}, \bibinfo{person}{Boyang Deng}, \bibinfo{person}{Paul Debevec}, \bibinfo{person}{William~T Freeman}, {and} \bibinfo{person}{Jonathan~T Barron}.} \bibinfo{year}{2021}\natexlab{}.
\newblock \showarticletitle{Nerfactor: Neural factorization of shape and reflectance under an unknown illumination}.
\newblock \bibinfo{journal}{\emph{ACM Transactions on Graphics (ToG)}} \bibinfo{volume}{40}, \bibinfo{number}{6} (\bibinfo{year}{2021}), \bibinfo{pages}{1--18}.
\newblock


\bibitem[Zheng et~al\mbox{.}(2023)]%
        {zheng2023avatarrex}
\bibfield{author}{\bibinfo{person}{Zerong Zheng}, \bibinfo{person}{Xiaochen Zhao}, \bibinfo{person}{Hongwen Zhang}, \bibinfo{person}{Boning Liu}, {and} \bibinfo{person}{Yebin Liu}.} \bibinfo{year}{2023}\natexlab{}.
\newblock \showarticletitle{AvatarRex: Real-time Expressive Full-body Avatars}.
\newblock \bibinfo{journal}{\emph{ACM Transactions on Graphics (TOG)}} \bibinfo{volume}{42}, \bibinfo{number}{4} (\bibinfo{year}{2023}).
\newblock


\bibitem[Zwicker et~al\mbox{.}(2002)]%
        {zwicker2002ewa}
\bibfield{author}{\bibinfo{person}{Matthias Zwicker}, \bibinfo{person}{Hanspeter Pfister}, \bibinfo{person}{Jeroen Van~Baar}, {and} \bibinfo{person}{Markus Gross}.} \bibinfo{year}{2002}\natexlab{}.
\newblock \showarticletitle{EWA splatting}.
\newblock \bibinfo{journal}{\emph{IEEE Transactions on Visualization and Computer Graphics}} \bibinfo{volume}{8}, \bibinfo{number}{3} (\bibinfo{year}{2002}), \bibinfo{pages}{223--238}.
\newblock


\end{thebibliography}
%
%%%%%%%%%%%%%%%%%%%%%%%%
% ADDITIONAL 2 PAGE FIGURES
%%%%%%%%%%%%%%%%%%%%%%%%
%
\clearpage
%
%%%%%%%%%%%%%%%%%%%%%%%%
% APPENDIX
%%%%%%%%%%%%%%%%%%%%%%%%
%
% Appendix
% \appendix
% \section{Appendix}
%
%%%%%%%%%%%%%%%%%%%%%%%%
% END BODY
%%%%%%%%%%%%%%%%%%%%%%%%
%
%
%%%%%%%%%%%%%%%%%%%%%%%%
% END DOC
%%%%%%%%%%%%%%%%%%%%%%%%
%
\appendix
\section{Overview}

In this supplemental document, we provide details of the data collection scheme (Sec.~\ref{sup:data-collection}), the implementation of Inverse Kinematics (Sec.~\ref{sup:ik}), the Depth-conditioned Animatable Avatar (Sec.~\ref{sup:dcaa}), the Relightable Avatar Appearance Modeling (Sec.~\ref{sup:relight}), and the Color Correction Model (Sec.~\ref{sup:hdr}).

\section{Data Collection Details} 
\label{sup:data-collection}

\noindent \textbf{Dataset Overview.}
In total, we captured data from four distinct subjects, featuring diverse outfits that span a wide range of clothing materials and skin tones.
This diversity facilitates a robust evaluation of our relightable model's capability across different recorded identities and apparel.
Furthermore, to demonstrate real-world applicability, two of the subjects were selected for in-the-wild testing in distinct indoor and outdoor environments, showing that our inverse-rendering-based HDR capture works ubiquitously and is not overfitted to a specific subject or relightable model.

\noindent \textbf{Human Performance Details.}
Following the standard practice for training and evaluation of motion-driven human avatars~\cite{habermann2021real}, we record human performance sequences featuring a diverse array of motions, which are subsequently partitioned into training and testing sets.
Table~\ref{tab:motion_list} details the motion sequences captured within the lightstage for both training and evaluation.
For the training split, each motion category is recorded for 20 seconds, whereas the testing sequences are limited to 5 seconds. 
Notably, while the motion types are consistent, the exact motion sequences in the test set are distinct from those in training. 
Furthermore, we include a separate 'casual motion' session at the end of the test sequence.
This serves to strictly validate our animation and rendering models on novel motions without explicit instruction.

\begin{table}[t]
    \centering
    \caption{List of motions in mixed-motion training and testing sequences of three subjects, following~\citet{chen2024egoavatar}. Note that we collect a sequence of unseen casual motion only in the test split.}
    \begin{tabular}{c c c}
    \hline
        T-pose & Jogging & Walking \\
        Picking-up object & Waving hands & Celebrating \\ Singing & Boxing & Kicking football \\ Golfing &
        Playing archery & Bicep curls \\ Playing instrument & Opening door & Playing juggle ball \\ Hula-hooping & Mopping floor & Bowling \\ Playing judo & Giving presentation & Stretching \\ Digging & Cooking & Drinking \\ Typing keyboard & Petting animals & Using body spray\\ Sitting & Playing shuttlecock & Raising legs \\ Flying & \underline{Casual Motion} \\
    \hline
    \end{tabular}
    \label{tab:motion_list}
\end{table}

\noindent \textbf{Lightstage Capture Details.}
As introduced in Sec. 3 of the main paper, we capture synchronized egocentric and multiview videos within the lightstage.
Temporal synchronization is achieved by projecting and detecting unique light patterns.
For geometric calibration, we record a dedicated sequence to compute the hand-eye transformation between the HMD's coordinate system and the local frames of both the front-facing and down-facing egocentric cameras.

Notably, the built-in SLAM tracking of the VR headset fails to maintain stability due to the intense and flickering illumination in the lightstage.
To address this, we attach ArUco markers to the HMD. 
We then localize the 6-DoF head pose by triangulating these markers using the lightstage's external cameras and combining them with the pre-computed hand-eye transformation to precisely track the camera pose.

\noindent \textbf{In-the-wild Capture Details.}
For in-the-wild demo scenarios outside the capture dome, we rely on the VR headset's built-in tracker while recording the egocentric front- and down-facing video streams.
Synchronization across all sensors is maintained from a global clock.
To compensate for latency in the built-in tracker, we propose to run a secondary visual odometry step on the front-facing camera frames. 
We then optimize a 6-DoF offset to align the headset's tracking trajectory with this vision-based trajectory.

% \section{Details about Mesh-based Geometry Model}
% \label{sup:mesh-model}

\section{Details about Inverse Kinematics}
\label{sup:ik}
\noindent \textbf{Energy Terms.} We describe the detailed formulation of the energy terms used below.
The data term
\begin{equation}
    E_\mathrm{Data} = \sum||\mathcal{K}(\{\boldsymbol{\theta}\}_1^T) - \{\boldsymbol{\zeta}\}_1^T||_2.
\end{equation}
aligns the skeleton joints $\mathcal{K}(\{\boldsymbol{\theta}\}_1^T)$ with detected 3D joints $\{\boldsymbol{\zeta}\}_1^T$.
In addition, we leverage the DOF Limit term and regularization term to constrain the motion space, preventing undesirable motion for mesh skinning and animation.
\begin{align}
    E_\mathrm{DoFLimit}=& \sum_{d=0}^{106} || \max ( \boldsymbol{\theta}^\tau_d - \boldsymbol{\theta}_{\max,d} , -\boldsymbol{\theta}^\tau_d + \boldsymbol{\theta}_{\min,d} , 0) ||_2 \\
    E_\mathrm{Reg} =& \sum_{d=0}^{106} ||\boldsymbol{\theta}_d - \bar{\boldsymbol{\theta}}_d||_2
\end{align}
Here, $\boldsymbol{\theta}_{\min, \cdot},\boldsymbol{\theta}_{\max, \cdot}$ denote the lower and upper bounds for each joint motion, and $\bar{\boldsymbol{\theta}}_d$ denotes the mean motion in the combined training sequence of all characters.

\noindent \textbf{Optimization Details.}
We use LBFGS as our optimizer.
For each stage of the optimization, we run 100 iterations.
In the first two stages (\textit{i.e.}, global 6D pose and body motion optimization), we reweight the temporal loss by $3$, the regularization loss by $0.01$, and the DoF Limit loss by $0.1$.
As egocentric 3D hand joint detection is easily affected by self-occlusion and fast motion, we increase the temporal loss to $30$.
\section{Details about Depth-conditioned Animatable Avatar}
\label{sup:dcaa}
\noindent \textbf{Network Architecture.}
We leverage a U-Net architecture with three cascaded downsampling and upsampling layers.
The input features $\boldsymbol{\Theta}, \boldsymbol{\xi}$ are rasterized to a UV map of size $128\times 128$.
Each downsampling layer downsamples the feature map by a factor of two.
After the convolutional U-Net, we first sample the UV features for each embedded graph node and vertex, and then use two MLPs to predict the embedded graph parameters $\boldsymbol{\alpha}, \boldsymbol{\beta}$ and the per-vertex offset $\mathbf{o}$.
The embedded graph feature is sampled from the feature map at a resolution of $32\times 32$, whereas the vertex feature is sampled from the feature map at the original $128\times 128$ resolution.

\noindent \textbf{Loss Functions and Training Details.}
For training efficiency, our AnimationNet is learned using purely 3D supervision:
\begin{equation}
    \begin{split}
        \mathcal{L}_\mathrm{animation} =& w_\mathrm{EG}\mathcal{L}_\mathrm{chamf}(\boldsymbol{V}_\mathrm{EG}, \boldsymbol{V}_\mathrm{recon})+ w_\mathrm{Delta}\mathcal{L}_\mathrm{chamf}(\boldsymbol{V}_\mathrm{EG}, 
        \boldsymbol{V}_\mathrm{recon}) \\ + & w_\mathrm{ARAP}\mathcal{L}_\mathrm{ARAP}(\boldsymbol{V}_\mathrm{EG})+ w_\mathrm{spatial}\mathcal{L}_\mathrm{spatial}(\boldsymbol{V}_\mathrm{Delta}) \\ +&   w_\mathrm{ISO}\mathcal{L}_\mathrm{ISO}(\boldsymbol{V}_\mathrm{Delta})
        \label{eq:dcaa-loss}
    \end{split}
\end{equation}
where $\boldsymbol{V}_\mathrm{EG}$ refers to the mesh animated by embedded graph only with $\boldsymbol{o} = \boldsymbol{0}$ and $\boldsymbol{V}_\mathrm{Delta}$ refers to the output mesh denoted by $\boldsymbol{V}$ in the main paper, and $\boldsymbol{V}_\mathrm{recon}$ refers to the reconstructed mesh from multiview stereo~\cite{wang2023neus2}.
Crucially, we found that direct supervision with Chamfer distance leads to artifacts under self-contact, since NeuS2 only reconstructs the outer surface of the 3D human body.
Therefore, we further filter out nearest-neighbor correspondences in the Chamfer distance that have large normal disparity, similar to the correspondence filtering applied to the input depth condition $\boldsymbol{\xi}$.
We train AnimationNet for 720K steps until it fully converges.
Since our captured relit sequences lack paired egocentric depth inputs, we train a separate depth-agnostic student model. 
We fine-tune this model using the interleaved flat-lit frames. 
This allows the student model to robustly track the 3D surface in the challenging relit frames, providing the geometric consistency required to learn relightable 3D Gaussian textures within a unified UV topology.

\section{Details about RELIGHTABLE AVATAR APPEARANCE MODELING AND RENDERING}
\label{sup:relight}

\noindent \textbf{Network Architecture.}
Similar to AnimationNet $\mathcal{G}_\mathrm{Anim}$, our relightable avatar appearance model $\mathcal{H}=\{\mathcal{H}_\mathrm{Lift},\mathcal{H}_\mathrm{Diff},\mathcal{H}_\mathrm{Spec}\}$ employs a U-Net architecture, operating with an input and output spatial resolution of $512\times 512$.
To accommodate the fact that our Gaussian primitives are significantly denser than the mesh vertices $\boldsymbol{V}$, we utilize a deeper network structure with four levels of hierarchical downsampling.
Unlike the GeoLifting and Diffuse networks, the SpecularNet ($\mathcal{H}_\mathrm{Spec}$) incorporates an additional cross-attention branch to process per-primitive ray encodings.  
To mitigate computational costs, we inject these cross-attention layers specifically at the intermediate $128 \times 128$ resolution, applying them both before downsampling and after the corresponding upsampling blocks.

\noindent \textbf{Loss Functions and Training Details.}
The GeoLiftingNet is first trained with flat-lit images and NeuS2 reconstructions in 400K iterations.
Then, DiffuseNet and SpecularNet are jointly trained with multiview image losses for another 200K iterations.
For GeoLiftingNet, we balance the loss function with the perceptual loss upscaled by 80 for the Sapiens~\cite{khirodkar2025sapiens}-extracted normals.
For training the relightable appearance model, we use $w_\mathrm{SSIM}=10$ and $w_\mathrm{IDMRF}=80$, whereas all other loss weights are set to 1.

\begin{figure}[t]
    \centering
    \includegraphics[width=0.48\textwidth]{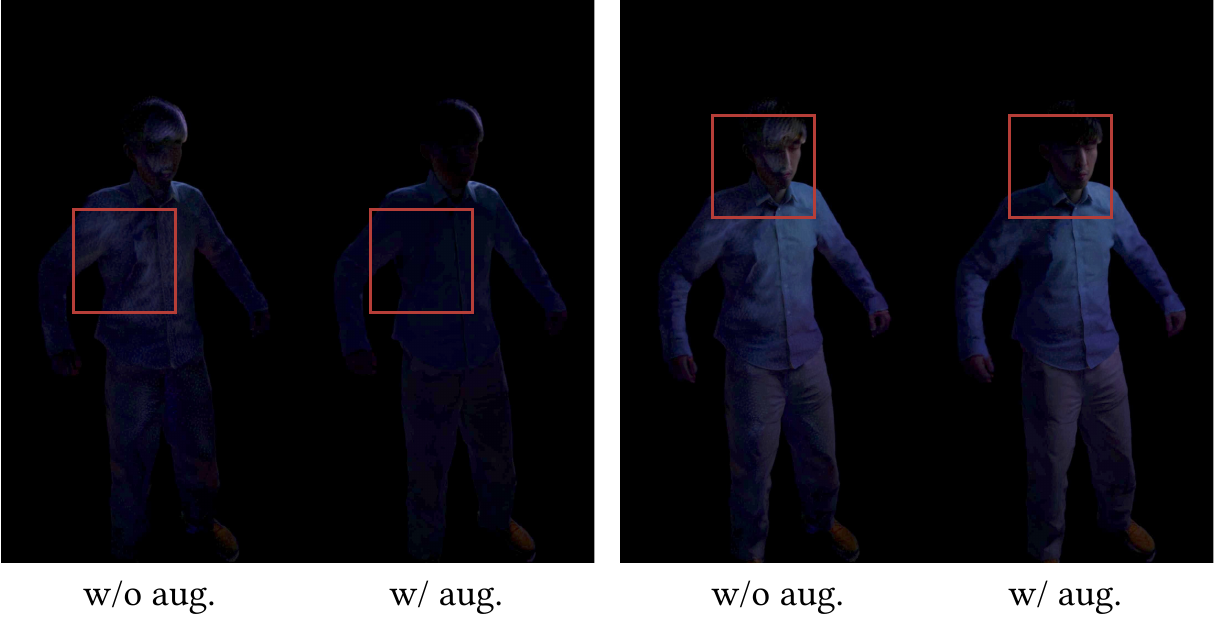}
    \caption{
    \textbf{Effect of data augmentation.}
    Qualitative comparison of our approach trained with and without our proposed augmentation strategy under novel One-Light-At-A-Time (OLAT) environment lighting.
    Without augmentation (left), the baseline suffers from severe artifacts in shadows, whereas our full pipeline (right) yields robust results.
    Note that image brightness is increased by $40\%$ for visualization purposes.
    }
    \label{fig:exp-supp-aug}
\end{figure}

To enhance the generalization of our learning-based relightable model to challenging, unseen illumination, such as OLAT and high-contrast environment maps, we incorporate a data augmentation strategy during training. 
Leveraging the linear properties of our RAW video captures, we jointly rescale both the incident light and the resulting image intensity in linear space. 
Specifically, after the first 100K iterations, we apply this augmentation with a $40\%$ probability using a scaling factor uniformly sampled from $[0, 1]$. 
As shown in Fig. \ref{fig:exp-supp-aug}, this approach effectively suppresses artifacts under extreme lighting.

%
%%%%%%%%%%%%%%%%%%%%%%%%%%%%%%%%%%%%%%%%%%%%%%
%
%
%%%%%%%%%%%%%%%%%%%%%%%%%%%%%%%%%%%%%%%%%%%%%%
%

\section{Details about EGOCENTRIC HDR ENVIRONMENT
MAP CAPTURE}
\label{sup:hdr}
\noindent \textbf{Color Calibration Model.} 
First, both egocentric images from down-facing and front-facing cameras are mapped from sRGB space to linear space.
Then, given a pixel with LDR color $(r,b,g)$ from the stitched panorama image $\boldsymbol{E}_l$, the camera model $\mathcal{C}$ transforms it to HDR space $(r',g',b')$ by
\begin{equation}
    \begin{bmatrix}
r'\\
g'\\
b'
\end{bmatrix}= \boldsymbol{A} \begin{bmatrix}
r^\gamma \\
g^\gamma \\
b^\gamma \\
\sqrt{r^\gamma g^\gamma} \\
\sqrt{g^\gamma b^\gamma} \\
\sqrt{b^\gamma r^\gamma}
\end{bmatrix}
\end{equation}
following the second-order model of \cite{finlayson2015color}, with the gamma term accounting for non-linearity.

\noindent \textbf{Optimization Details.}
In addition to the rendering loss described in the main paper, we formalize the regularization loss as 
\begin{equation}
    \mathcal{L}_\mathrm{Reg}(\mathbf{E}_h) = ||\max (\mathbf{E}_h-1, 0)||_2 + ||\max (-\mathbf{E}_h, 0)||_2 
\end{equation}
which enforces that every pixel in the environment map has a value in $[0,1]$.
For each scene, we perform a single optimization pass of 1,000 steps using the Adam optimizer until convergence.
\end{document}